\documentclass[journal]{IEEEtran}
\usepackage{setspace,amssymb,amsmath,cite,enumerate,graphicx,multirow,subfigure}
\usepackage[linesnumbered,ruled,vlined]{algorithm2e}

\begin{document}

\title{Balancing Common Treatment and Epidemic Control in Medical Procurement during COVID-19: Transform-and-Divide Evolutionary Optimization}

\author{Yu-Jun~Zheng,~\IEEEmembership{Senior Member,~IEEE,}
        Xin Chen,
        Tie-Er Gan,
        Min-Xia Zhang,
        Wei-Guo Sheng,~\IEEEmembership{Member,~IEEE,}
        and Ling Wang
\thanks{Y.J. Zheng, X. Chen, and W.G. Sheng are with the School of Information Science and Engineering, Hangzhou Normal University, Hangzhou 311121, China (e-mail: yujun.zheng@computer.org; w.sheng@ieee.org).}
\thanks{T.E. Gan is with the First Affiliated Hospital, Zhejiang Chinese Medical University, Hangzhou 310006, China (e-mail: gantieer@163.com).}
\thanks{M.X. Zhang is with the College of Computer Science \& Technology, Zhejiang University of Technology, Hangzhou 310023, China (e-mail:zmx@zjut.edu.cn).}
\thanks{L. Wang is with the Department of Automation, TSinghua University, Beijing 100084, China (e-mail:wangling@tsinghua.edu.cn).}
}


\maketitle

\begin{abstract}
Balancing common disease treatment and epidemic control is a key objective of medical supplies procurement in hospitals during a pandemic such as COVID-19. This problem can be formulated as a bi-objective optimization problem for simultaneously optimizing the effects of common disease treatment and epidemic control. However, due to the large number of supplies, difficulties in evaluating the effects, and the strict budget constraint, it is difficult for existing evolutionary multiobjective algorithms to efficiently approximate the Pareto front of the problem. In this paper, we present an approach that first transforms the original high-dimensional, constrained multiobjective optimization problem to a low-dimensional, unconstrained multiobjective optimization problem, and then evaluates each solution to the transformed problem by solving a set of simple single-objective optimization subproblems, such that the problem can be efficiently solved by existing evolutionary multiobjective algorithms. We applied the transform-and-divide evolutionary optimization approach to six hospitals in Zhejiang Province, China, during the peak of COVID-19. Results showed that the proposed approach exhibits significantly better performance than that of directly solving the original problem. Our study has also shown that transform-and-divide evolutionary optimization based on problem-specific knowledge can be an efficient solution approach to many other complex problems and, therefore, enlarge the application field of evolutionary algorithms.
\end{abstract}


\IEEEpeerreviewmaketitle

\section{Introduction}
\IEEEPARstart{T}o mount an effective response to COVID-19, hospitals must procure medical supplies for epidemic control. However, the total budget of any hospital is limited: if a hospital procures too many supplies for epidemic control, it has to reduce supplies for common disease treatment, which would damage its medical services. Consequently, it is important for a hospital to balance between common disease treatment and epidemic control in medical supplies procurement. The problem of determining the purchase quantity of each supply can be formulated as a \emph{bi-objective optimization problem} for simultaneously optimizing the effect of epidemic control and effect of common disease treatment. There are three main challenges to solving this problem. The first is to evaluate the effects of epidemic control and common disease treatment in a relatively accurate manner. The second is to meet the budget constraint, which is often strict in the pandemic. The third is to approximate the Pareto front of the problem in an efficient manner. For the first challenge, we develop a procedure to simulate the arrival and treatment of cases of infection and cases of common diseases according to a general principle of disease treatment and medical supplies usage. However, this also makes the objective functions expensive. Moreover, a major hospital often involves tens of thousands of medical supplies, which makes the dimension of the solution space too high. The combination of these reasons makes the problem very difficult to solve.

Decomposition is a general approach to solving large complex problems that are beyond the reach of standard techniques. Decomposition in optimization appears in early work on large-scale linear programming problems from the 1960s \cite{Dant60}. Many problems with separable objective functions are trivial to solve by mathematical methods. For additively decomposed functions that are not able to optimize by standard genetic algorithms, M\"{u}hlenbein and Mahnig \cite{Muhl99EA} proposed the factorized distribution algorithm that factors the distribution into conditional and marginal distributions based on function structures. Strasser et al. \cite{Strass17TEC} proposed factored evolutionary algorithms, which factors an optimization problem by creating overlapping subpopulations that optimize over subsets of variables. Many combinatorial optimization problems can be solved by using efficient methods to solve subproblems and combining the results to obtain solutions to the original problems. Zheng and Xue \cite{Zheng10Comp} utilized these characteristics to automatically derive efficient problem-solving algorithms, including evolutionary algorithms (EAs) that are mainly used for NP-hard problems. Unfortunately, the medical supplies procurement problem considered in this paper does not satisfy the basic conditions of decomposition, because it is quite common that one supply can be used in multiple diseases, and the treatment of a disease can involve many supplies.

For complex problems whose subcomponents interact and affect each other, Potter and De Jong \cite{Potter00EC} proposed a cooperative coevolution architecture that decomposes a problem into subcomponents, which are then evolved as a collection of cooperating species. Different decomposition methods have different performance in cooperative coevolution for different problems. Yang et al. \cite{Yang08InfSci} proposed a cooperative coevolution framework that uses random grouping and adaptive weighting in problem decomposition and coevolution for optimizing large nonseparable problems. Chandra \cite{Chan15TNNLS} presented a competitive cooperative coevolution method for training recurrent neural networks, where different decomposition methods compete with different features they have in terms of diversity and degree of non-separability. To capture the interdependency among variables, Hu et al. \cite{HuX17InfSci} proposed a fast interdependency identification algorithm, which avoids obtaining the full interdependency information of nonseparable variables for cooperative coevolution. Yang et al. \cite{YangQ18Access} proposed a data-driven approach, which exploits historical data to mine the evolution consistency among variables for dynamic variable grouping. Gomes et al. \cite{Gomes18TEC} extended cooperative coevolution with operators that put the number of coevolving subpopulations under evolutionary control. Wang et al. \cite{WangY18EC} proposed a formula-based grouping strategy for grouping variables directly based on the separable and nonseparable operations in the formula of an objective function. Mahdavi et al. \cite{Mahd18SOCO} proposed an incremental cooperative coevolution algorithm in which subcomponents are dynamically added to an integrated subcomponent being evolved. For multimodal optimization problems, Peng et al. \cite{Peng19TCyb} proposed a method that concurrently searches for multiple optima as informative representatives to be exchanged among subcomponents for compensation in coevolution. For dynamic optimization problems, Peng et al. \cite{Peng16KBS} used multiple populations in cooperative coevolution to compensate information in dynamic landscapes. Zhang et al. \cite{ZhangX19TEC} proposed a dynamic cooperative coevolution framework, which allocates computational resources to elitist subcomponents with superior variables. Although cooperative coevolution has shown its efficiency in solving many engineering optimization problems \cite{Zheng13APIN,Nguyen14TEC,SunL19TFS,MaX19TEC,Zheng20TEVC}, for most nonseparable problems, decomposition causes the loss of a great deal of information, and the algorithms easily gravitate towards sub-optima represented by Nash equilibria rather than global optima \cite{Wiegand03Coev}.

A multiobjective optimization problem is much more difficult than its single-objective counterpart, and decomposition is also a basic strategy in multiobjective optimization. Zhang and Li \cite{ZhangQ07TEC} proposed a multiobjective evolutionary algorithms based on decomposition (MOEA/D), which decomposes a multiobjective optimization problem into a set of single-objective optimization subproblems using decomposition approaches such as weighted sum, weighted Tchebycheff, and penalty-based boundary interaction. However, for some problems, these decomposition approaches may not be suitable for balancing the diversity and convergence. Wang et al. \cite{WangL16TEC} revolved this difficulty by imposing a constraint to an unconstrained subproblem, where the improvement region of each subproblem is determined by an adaptive control parameter. MOEA/D makes an assumption that two neighboring subproblems have similar optimal solutions, but some problems do not satisfy this assumption. Mei et al. \cite{Mei11TEC} proposed a decomposition-based memetic algorithm with neighborhood search for multiobjective capacitated arc routing problem, which combines decomposition-based and domination-based techniques for solution selection. An EA proposed by Cai et al. \cite{Cai15TEC} also combined domination-based sorting and decomposition, the former for evolving an internal population and the latter for maintaining an external archive. Jan and Zhang \cite{Jan10UKCI} introduced a penalty function to MOEA/D for multiobjective constrained optimization. Konstantinidis and Yang \cite{Kons11CompComm} adapted MOEA/D to solve a $K$-connected deployment and power assignment problem by introducing a problem-specific repair heuristic for infeasible solutions. Zhang et al. \cite{Zhang16MemComp} extended MOEA/D for big optimization problems by embedding a gradient-based local search. Chen et al. \cite{ChenJ17TEC} extended MOEA/D by assigning each subproblem with an upper bound vector based on $\epsilon$-constraint for constrained optimization. Qiao et al. equipped MOEA/D with an angle-based adaptive penalty scheme. Fang et al. \cite{FangW18TCyb} proposed coordinate transformation in the objective space to accelerate the convergence process of multiobjective EA. Jiang et al. \cite{JiangS18TEC} studied the effect of scalarizing functions and presented two new functions for improving decomposition in MOEA/D. There have been many other extensions and applications of MOEA/D in recent years \cite{Triv17TEC}. The cooperative coevolution architecture that decomposes a problem in the decision space has also been extended for multiobjective optimization \cite{Tan06TEC,WangJ16TCyb,Anton18TEC,Gong20TEC}. To our knowledge, there is only one study, by He et el. \cite{HeC19TEC}, combining decomposition in both the decision space and objective space, which is similar to our work. The key difference is that our method utilizes problem-specific knowledge to ensure the separability of the transformed problem. Unfortunately, we found that, although using decomposition-based strategies in MOEA/D and other similar algorithms \cite{Moub14EC,Dai15InfSci,Cai18TEC,ZhangX18TEC,WangY18TSMC} can reduce the complexity to a certain degree, the performance of those algorithms is still far from satisfactory in solving the medical supplies procurement problem in practice.

In this study, we present a transform-and-divide approach to efficiently solve the problem. First, we transform the original problem of determining the purchase quantity of each supply to a new problem of distributing the budget to epidemic control and all common diseases. In our case studies, the dimension of the transformed problem is only one to two percent of that of the  original problem. However, evaluating each solution to the transformed problem is itself a nontrivial optimization problem. Second, we divide the evaluation problem into a set of low-dimensional, single-objective optimization subproblems. We propose a hybrid evolutionary optimization approach, which employs a multiobjective EA to evolve a population of main solutions to the transformed problem and uses a tabu search algorithm to solve the subproblems. During the peak of COVID-19, we applied the proposed approach to six hospitals in Zhejiang Province, China. Results demonstrated that the transform-and-divide evolutionary optimization approach exhibits significantly better performance than that of directly using multiobjective EAs to solve the original problem. The main contributions of this paper are twofold:
\begin{itemize}
\item We propose a transform-and-divide evolutionary optimization approach to medical supplies procurement and demonstrate its practicability and efficiency during COVID-19.
\item We show that using problem-specific knowledge to transform and divide a complex optimization problem can lead to competitive EAs for the problem. This approach can be extended to many other problems and enlarge the application field of EAs.
\end{itemize}

The remainder of this paper is organized as follows. Section \ref{sec:prob} presents the medical supplies procurement problem. Section \ref{sec:oldalg} simply describes how to directly use basic multiobjective EAs to solve the original problem. Section \ref{sec:newalg} proposes the transform-and-divide evolutionary optimization approach. section \ref{sec:exp} presents the computational results. Section \ref{sec:conclu} concludes with a discussion.

\section{Problem Description}\label{sec:prob}
\subsection{Supplies for Epidemic Control and Common Treatment}
We consider a medical supplies procurement problem formulated as follows. In a pandemic, a hospital plans to procure medical supplies, including a set $\mathbf{S}\!=\!\{S_1,S_2,\ldots,S_n\}$ of $n$ supplies for epidemic control, and a set $\mathbf{S'}\!=\!\{S_{n\!+\!1},S_{n\!+\!2},\ldots,S_{n\!+\!n'}\}$ of $n'$ supplies for normal disease treatment. For each supply $S_k$, the current inventory is $a_k$, the unit price is $c_k$, and the unit volume is $v_k$. The problem is to determine the purchase quantity $x_k$ of each supply ($1\!\le\!k\!\le\!n\!+\!n'$), such that the effects of epidemic control and normal disease treatment are simultaneously optimized.

The supplies for epidemic control can be divided into two classes. The first class consists of supplies such as latex gloves and normal saline that must be used in the treatment of a suspected case of infection; we use $\Psi_0$ to denote the set of these supplies, and use $q_{0,k}$ to denote the quantity of each $S_k\in\Psi_0$ required to treat a case. The second class consists of supplies that are alternative in some treatment items. Table \ref{tab:ep-sup} presents six treatment items and their alternative supplies used for COVID-19 control in this study. The six sets of supplies are denoted by $\Psi_1,\Psi_2,\dots,\Psi_6$, respectively, and the quantity of each alternative $S_k\in\Psi_j$ required to treat a case is denoted by $q_{j,k}$. Different alternatives have different treatment effects. The treatment effect of using each alternative $S_k\in\Psi_j$ is estimated as $e_{j,k}$. For example, the effects of peroxide, impermeable gown, and normal gown in ``body protection'' item are estimated as 1, 0.9 and 0.7, respectively. If we choose the $S_{k_j}\in\Psi_j$ for the $j$-th treatment item ($1\!\le\!j\!\le\!6$), the corresponding epidemic control effect on the case is empirically estimated as:
\begin{equation}
E(k_1,...k_6)\!=\! (0.4e_{1,k_1}\!+\!0.6e_{2,k_2})e_{3,k_3}(0.2e_{4,k_4}\!+\!0.8e_{5,k_5})e_{6,k_6}
\label{eq:teffect}\end{equation}

\begin{table*}
\center\footnotesize\setlength\tabcolsep{3pt}
\caption{Alternative supplies for epidemic control in this study.}
\begin{tabular}{c|lllllll}\hline\
Items &Body protection &Face protection &Detection &Oxygen therapy &Antivirus &Disinfectant \\\hline
\multirow{4}{0.9cm}{Supplies}&protective clothing &face shield &nucleic acid kit &high-flow nasal cannula &$\alpha$-interferon &peroxide\\
 &impermeable gown &N95 mask+goggle &antibody kit &nasal cannula &lopinavir &chlorine-containing\\
 &normal gown &surgical mask+goggle & &oxygen mask &chloroquine phosphate &alcohols\\
 & & & & &arbidol\\\hline
\end{tabular}
\label{tab:ep-sup}\end{table*}

The hospital is capable of treating a set $\mathbf{D}\!=\!\{D_1,D_2,\ldots,D_m\}$ of $m$ diseases. Similarly, for each disease $D_i$, the set of supplies that must be used is denoted by $\Psi_{i,0}$, and the set of supplies that are alternative in $J_i$ treatment items are denoted by $\Psi_{i,1},\Psi_{i,2},\ldots,\Psi_{i,J_i}$, respectively. The quantity of each $S_k\in\Psi_{i,0}$ required to treat a case is $q_{i,0,k}$, and the quantity of each alternative $S_k\in\Psi_{i,j}$ required to treat a case is $q_{i,j,k}$. Different alternatives have different treatment effects. The treatment effect of using $S_k\in\Psi_{i,j}$ is estimated as $e_{i,j,k}$. If we choose the $S_{k_j}\in\Psi_{i,j}$ for the $j$-th treatment item ($1\!\le\!j\!\le\!J_i$), the corresponding treatment effect on the case is empirically estimated by a therapeutic effect function $E_i(k_1,k_2,...,k_{J_i})$. Like Eq. (\ref{eq:teffect}), the typical expression of $E_i$ is a weighted sum or product of $e_{i,j,k}$ \cite{Song19TII}.

\subsection{Number of Cases}
Let $T$ be the procurement decision cycle. In our case studies, the hospital procures medical supplies every 15 days. The supply quantities are determined based on the estimation of the number of hospital visits in the next decision cycle. For the number of cases of each disease $D_i$, we estimate three values: the expected value $r_i$, lower limit (optimistic value) $\underline{r_i}$, and upper limit (pessimistic value) $\overline{r_i}$. The values can be obtained based on historical morbidity data and environmental influence factors \cite{Dmit13BioEng,Land02EHP,Song17Neucom,Song19IJERPH}.

The number of suspected cases of epidemic infection is estimated based on the number of hospital visits of different diseases. For each disease $D_i$, we estimate a probability $p_i$ that a patient of $D_i$ is a suspected case of COVID-19. In general, a disease having more similar symptoms with the epidemic has a higher $p_i$. For example, an acute respiratory infectious disease has a high $p_i$. For a disease (such as fracture) that is unrelated to the epidemic, we set $p_i$ to the current incidence $p_e$ of infection (including suspected infection) in the local region. We also estimate an average number $r'_i$ of accompanying persons of a patient of $D_i$; in general, a critical disease has a large $r'_i$. The probability that an accompanying person of a patient of $D_i$ is a suspected case of COVID-19 is $p'_i$, which is set to $0.5p_i$ if $D_i$ has similar symptoms with the epidemic and $p_e$ otherwise. 
The total number of suspected cases of infection in the next decision cycle is estimated as follows (we use $\overline{r_i}$ as we take a serious or pessimistic view of epidemic control):
\begin{equation}
r_0\!=\! \sum_{i=1}^m (p_i+p'_ir'_i)\overline{r_i}
\label{eq:n-infec}\end{equation}

\subsection{Objective Function Evaluation}
A solution to the medical supplies procurement problem can be represented by a $(n\!+\!n')$-dimensional vector $\mathbf{x}=\{x_1,\dots,x_n,x_{n\!+\!1},\dots,x_{n\!+\!n'}\}$. The fitness of $\mathbf{x}$ is evaluated by two objective functions: (1) the epidemic control effect $\Upsilon(\mathbf{x})$, which is the sum of treatment effects of all suspected cases of infection; (2) the common disease treatment effect $\Upsilon'(\mathbf{x})$, which is the weighted sum of treatment effects of all common cases, where the weight of each $D_i$ is $w_i$. It is assumed that the arrival time of cases follows a uniform distribution. That is, for each disease $D_i$, as the expected number of cases in a decision cycle of 15 days is $r_i$, then there is a case arriving every $(15\!\times h_w)/r_i$ hours, while $h_w$ is the daily working hours (24 for emergency diseases and 8 for non-emergency diseases in our study). Moreover, there is one suspected case of infection in every $1/(p_i+p'_ir'_i)$ cases of $D_i$.

A general principle of disease treatment is ``focusing on the current patient'', i.e., whenever a new case arrives, the physician always chooses the most effective supply from the available alternatives, as he does not know how many cases would come later. Based on this first-come-first-served discipline, we sort supplies in $\Psi_j$ ($1\!\le\!j\!\le\!6$) or $\Psi_{i,j}$ ($1\!\le\!i\!\le\!m;1\!\le\!j\!\le\!J_i$) in nonincreasing order of $e_{j,k}$ or $e_{i,j,k}$, and simulate the arrival and treatment of all cases according to the procedure shown in Algorithm \ref{alg:queue} to calculate the values of $\Upsilon(\mathbf{x})$ and $\Upsilon'(\mathbf{x})$. In Algorithm \ref{alg:queue}, the boolean variable $\textit{tr}$ denotes whether the remaining supplies are capable of treating a suspected case of epidemic infection, and $\textit{tr}_i$ denotes whether the remaining supplies are capable of treating a case of $D_i$.

\begin{algorithm}\small
Initialize $\Upsilon=0$, $\textit{tr}=\textit{true}$\;
\lFor{$i=1$ to $m$}{ initialize $\Upsilon_i=0$, $\textit{tr}_i=\textit{true}$\;}
\lFor{$k=1$ to $n\!+\!n'$}{ $a_k\!\leftarrow\!a_k\!+\!x_k$\; }
Start timing simulation\;
\While{the decision cycle is not complete}{
    \If{a new case of $D_i$ arrives \textnormal{and} $\textit{tr}_i=\textit{true}$}{
        \ForEach{$S_k\in\Psi_{i,0}$}{
            $a_k\leftarrow a_k-q_{i,0,k}$\;
            \lIf{$a_k<q_{i,0,k}$}{$\textit{tr}_i\leftarrow\textit{false}$\;}
        }
        \For{$j=1$ to $J_i$}{
            let $S_k$ be the first supply in $\Psi_{i,j}$\;
            \While{$a_k<q_{i,j,k}$}{
                Remove $S_k$ from $\Psi_{i,j}$\;
                \lIf{$\Psi_{i,j}=\emptyset$}{$\textit{tr}_i\leftarrow\textit{false}$\;}
            }
            let $k_j=k$\;
            $a_k\leftarrow a_k-q_{i,j,k}$\;
        }
        $\Upsilon_i\leftarrow \Upsilon_i+E_i(k_1,\ldots,k_{J_i})$\;
    }
    \If{the case is a suspected infected case \textnormal{and} $\textit{tr}=\textit{true}$}{
        \ForEach{$S_k\in\Psi_0$}{
            $a_k\leftarrow a_k-q_{0,k}$\;
            \lIf{$a_k<q_{0,k}$}{$\textit{tr}\leftarrow\textit{false}$\;}
        }
        \For{$j=1$ to $6$}{
            let $S_k$ be the first supply in $\Psi_j$\;
            \While{$a_k<q_{j,k}$}{
                Remove $S_k$ from $\Psi_j$\;
                \lIf{$\Psi_j=\emptyset$}{$\textit{tr}\leftarrow\textit{false}$\;}
            }
            $a_k\leftarrow a_k-q_{j,k}$\;
            let $k_j=k$\;
        }
        $\Upsilon\leftarrow \Upsilon+E(k_1,\dots,k_6)$\;
    }
}
\Return{$\Upsilon(\mathbf{x})=\Upsilon$ \textnormal{and} $\Upsilon'(\mathbf{x})=\sum_{i=1}^m w_i\Upsilon_i$.}
\caption{Procedure for evaluating the effects of epidemic control and disease treatment for the original problem.}
\label{alg:queue}\end{algorithm}

\subsection{Constraints}
A procurement solution $\mathbf{x}$ must satisfy problem constraints. First, the total procurement cost cannot exceed the budget $C$:

\begin{equation}
\sum_{k=1}^{n\!+\!n'}c_kx_k\le C
\label{eq:constrB}\end{equation}

The hospital should perform its normal functions. In this study, it is required that the hospital is able to treat $\underline{r_i}$ (the lower number) cases of each disease $D_i$. These constraints can be tested by simulating the arrival and treatment of the lower numbers of cases in Algorithm \ref{alg:queue}: if any new case cannot be treated, i.e., whenever the condition $\textit{tr}_i\!=\!\textit{false}$ (Line 6 of Algorithm \ref{alg:queue}) is triggered while the number of simulated cases of $D_i$ is less than $\underline{r_i}$, the constraint is violated.

It is also required that the hospital is able to treat $r_0$ suspected cases of infection. Whenever the condition $\textit{tr}\!=\!\textit{false}$ (Line 18) in Algorithm \ref{alg:queue} is triggered, the constraint is violated.

\section{Basic Evolutionary Optimization Methods}\label{sec:oldalg}
For the above $(n\!+\!n')$-dimensional, constrained bi-objective optimization problem, we can use evolutionary constrained  multiobjective algorithms to search for the Pareto optimal solutions. The search range of each dimension $k$ is $[\underline{x_k},\overline{x_k}]$. The lower limit $\underline{x_k}$ is set to the total quantity of $S_k$ required in non-alternative treatment items for all cases:
\begin{equation}
\underline{x_k}=\left\{\begin{array}{l@{\quad\quad}l}
        \max(0,rq_{0,k}-a_k), & 1\!\le\!k\!\le\!n\\
        \max(0,\sum_{i=1}^m r_iq_{i,0,k}-a_k), & n\!+\!1\!\le\!k\!\le\!n\!+\!n'
    \end{array}\right.
\label{eq:range-l}\end{equation}

The upper limit $\overline{x_k}$ can be set to the total required quantity of $S_k$ under the assumption that $S_k$ is always chosen whenever $S_k$ is an alternative. That is, if $1\!\le\!k\!\le\!n$, we set
\begin{equation}
\overline{x_k}= \big(\sum\nolimits_{j'\in\{j|0\!\le\!j\!\le\!6\wedge S_k\in\Phi_j\}}rq_{j',k}\big)-a_k
\label{eq:range-u}\end{equation}

Otherwise, we set
\begin{equation}
\overline{x_k}=\big(\sum\nolimits_{(i',j')\in\{(i,j)|1\!\le\!i\!\le\!m\wedge 0\!\le\!j\!\le\!J_i\wedge S_k\in\Phi_{i,j}\}}r_{i'}q_{i',j',k}\big)-a_k
\label{eq:range-u'}\end{equation}

We adopt the following five well-known evolutionary constrained multiobjective algorithms to solve the problem:
\begin{itemize}
\item The nondominated sorting genetic algorithm II (NSGA-II) with the constrained-domination principle \cite{Deb02TEVC}.
\item The constrained multiobjective evolutionary algorithm (CMOEA) based on an adaptive penalty function and a distance measure \cite{Wold09TEC}.
\item The multiobjective evolutionary algorithm based on decomposition (MOEA/D) \cite{ZhangQ07TEC} with a penalty function for constrain handling \cite{Jan10UKCI}.
\item The differential evolution with self-adaptation and local search for constrained multiobjective optimization (DECMOSA) \cite{Zamu09CEC}, which combines constrained-domination and penalty function for constrain handling.
\item The multi-objective particle swarm optimizer based on dominance with decomposition (D$^2$MOPSO) \cite{Moub14EC}.
\end{itemize}

CMOEA, MOEA/D, and DECMOSA employ penalty functions for constrain handling. Violation of constraint (\ref{eq:constrB}) is calculated as $\max(0,\sum_{k=1}^{n\!+\!n'}c_kx_k-C)$. For constraints that all suspected cases of infection and the lower number of cases of each common disease must be treated, we set the violation of each constraint equal to the budget $C$, i.e., the violation is $C$ times the number of false $\textit{tr}_i$ and $\textit{tr}$ in Algorithm \ref{alg:queue}.

Nevertheless, the performance of all the above algorithms is not satisfying, mainly because the dimension $(n\!+\!n')$ is very high (approximately 10,000$\sim$40,000 in a major hospital in our case studies) and the evaluation of a solution using Algorithm \ref{alg:queue} is computationally expensive.

\section{A New Transform-and-Divide Evolutionary Optimization Method}\label{sec:newalg}
In this section, we propose a new transform-and-divide approach to efficiently solve the problem. First, we transform the original high-dimensional, constrained bi-objective optimization problem to a low-dimensional, unconstrained bi-objective optimization problem, which can be solved using evolutionary (unconstrained) multiobjective algorithms. The evaluation of each solution to the transformed problem can be divided into a set of low-dimensional, single-objective optimization subproblems, which can be solved using a tabu search algorithm.



\subsection{Problem Transformation}\label{sec:trans}
We transform the original problem of determining the purchase quantity of each supply to a problem of determining the purchase budget for epidemic control and the purchase budget for each disease. First of all, we calculate the cost for purchasing the supplies that must be used in the non-alternative treatment items and, therefore, obtain the remaining budget as:
\begin{equation}
C' = C-\sum_{k=1}^{n\!+\!n'} c_k\underline{x_k}
\label{eq:cost0}\end{equation}

Consequently, the transformed problem is to distribute $C'$ to $m\!+\!1$ components, denoted by $\{y_0,y_1,\ldots,y_m\}$, where $y_0$ is the budget for purchasing alternative supplies for epidemic control, and $y_i$ is the budget for purchasing alternative supplies for treating disease $D_i$ ($1\!\le\!i\!\le\!m$). The dimension of the transformed problem is $m$ (approximately 200$\sim$600 in a major hospital in our case studies), which is significantly smaller than the dimension $n\!+\!n'$ of the original problem.

Moreover, the search range of each dimension of the transformed problem is also much smaller than that of the original problem. For epidemic control, the lower limit $\underline{y_0}$ of budget $y_0$ can be obtained using the following steps:
\begin{enumerate}
\item Use supplies in storage to treat as many suspected cases of infection as possible;
\item If there is no remaining case, set $\underline{y_0}\!=\!0$;
\item Else, for each remaining case, always purchase the cheapest supply among the alternatives, and set $\underline{y_0}$ to the total purchase cost.
\end{enumerate}

And the upper limit $\overline{y_0}$ of budget $y_0$ can be obtained using the following steps:
\begin{enumerate}
\item Treat each suspected case in the most effective way, i.e, always select the supply with the maximum treatment effect $e_{j,k}$ among the alternatives, and calculate the total required quantity of each supply;
\item Calculate the purchase quantity of each supply, and set $\overline{y_0}$ to the total purchase cost.
\end{enumerate}

Therefore, the search range of $y_0$ is limited to $[\underline{y_0},\overline{y_0}]$. We can obtain the search range $[\underline{y_i},\overline{y_i}]$ of each $y_i$ in a similar manner. In our case studies, the average value of $(\overline{y_i}\!-\!\underline{y_i})$ is approximately 95 (in unit of 100 RMB), while that of $(\overline{x_k}\!-\!\underline{x_k})$ is approximately 1100 (in minimum order quantity).

\subsection{Problem Division}\label{sec:div}
The solution space of the transformed problem is significantly smaller than that of the original problem. But how to evaluate a solution $\mathbf{y}=\{y_0,y_1,\ldots,y_m\}$ to the transformed problem? The task can be divided into $m\!+\!1$ optimization subproblems. The first subproblem is to determine the purchase quantities under the budget $y_0$ so as to maximize the epidemic control effect. Each of the remaining $m$ subproblems is to determine the purchase quantities under the budget $y_i$ so as to maximize the treatment effect of disease $D_i$ ($1\!\le\!i\!\le\!m)$.

However, the division leads to difficulty in allocating supplies in storage to different diseases. We overcome this difficulty by employing a procedure similar to Algorithm \ref{alg:queue} to simulate the arrival and treatment of all cases. But the procedure has two differences from Algorithm \ref{alg:queue}:
\begin{itemize}
\item Initially, we only consider supplies in storage, i.e., Line 3 of Algorithm \ref{alg:queue} is not executed.
\item If there is no supply in storage that can be used for a treatment item (i.e., the condition in Line 14 or Line 26 is satisfied), we temporarily purchase ``in advance'' the cheapest alternative supply for the item.
\end{itemize}

The procedure also produces the ``cheapest'' solution to each subproblem, which can be evolved to an optimal or near-optimal solution, as described in the next subsection.

\subsection{Hybrid Evolutionary Optimization}\label{sec:hea}
The proposed method employs an evolutionary multiobjective algorithm to evolve a population of main solutions to the transformed problem, and employs a tabu search algorithm to solve the subproblems for evaluating each main solution.

For the first subproblem, each solution $\mathbf{z}$ can be represented by six vectors as follows (the vector lengths do not need to be the same):
\begin{eqnarray}
&\{z_{1,1},z_{1,2},\dots,z_{1,|\Psi_1|}\} \nonumber\\
&\{z_{2,1},z_{2,2},\dots,z_{2,|\Psi_2|}\} \nonumber\\
&\vdots \nonumber\\
&\{z_{6,1},z_{6,2},\dots,z_{6,|\Psi_6|}\} \nonumber
\end{eqnarray}
where $z_{j,k}$ denotes the number of cases that use the $k$-th alternative supply for the $j$-th treatment item, and each vector satisfies $\big(\sum_{k=1}^{|\Psi_j|}z_{j,k}\big)=r_0$.

The procedure described in Sec. \ref{sec:div} produces the cheapest solution to the subproblem, denoted by $\mathbf{z}^\dag$. First, we continually use the following steps to improve $\mathbf{z}^\dag$ by replacing an alternative supply to a more effective alternative for a randomly selected case until $\mathbf{z}^\dag$ cannot be further improved:
\begin{enumerate}
\item Randomly selecting two components $z_{j,k}$ and $z_{j,k'}$ in a vector satisfying $z_{j,k'}\!>\!0$;
\item Set $z_{j,k'}\!=\!z_{j,k'}\!-\!1$ and $z_{j,k}\!=\!z_{j,k}\!+\!1$ if doing so would not violate the budget constraint.
\end{enumerate}

Starting from the improved $\mathbf{z}^\dag$, the tabu search algorithm continually uses the following steps to search around and improve $\mathbf{z}^\dag$ until the stopping condition is satisfied:
\begin{enumerate}
\item Generate $k_N$ neighboring solutions of the current $\mathbf{z}^\dag$, each being obtained by randomly selecting two components $z_{j,k}$ and $z_{j',k'}$ satisfying $k\!<\!|\Psi_j|$, $k'\!<\!|\Psi_{j'}|$, $z_{j,k}\!>\!0$, and $z_{j',k'+1}\!>\!0$, and setting $z_{j,k}\!=\!z_{j,k}\!-\!1$, $z_{j,k+1}\!=\!z_{j,k+1}\!+\!1$, $z_{j',k'}\!=\!z_{j',k'}\!+\!1$, and $z_{j',k'+1}\!=\!z_{j',k'+1}\!-\!1$, if doing so would not violate the budget constraint;
\item Select the best neighbor that is not tabued or is better than the current $\mathbf{z}^\dag$, make $\mathbf{z}^\dag$ move to this neighbor, and add this move to the tabu list.
\end{enumerate}

The remaining $m$ subproblems can be solved by tabu search in a similar way. As demonstrated by the experiments, the tabu search algorithm can quickly obtain optimal solutions for most subproblem instances, given that the dimensions of the subproblems are relatively small. For example, as we can observe from Table \ref{tab:ep-sup}, the dimension of the first subproblem is 18 (note that the last dimension of each vector can be determined by other dimensions of the vector, and the actual dimension in the solution space is only 12). Therefore, the tabu search algorithm is very suitable for the subproblems, as it will be invoked many times to evaluate main solutions.

For the main transformed problem, we adopt the following evolutionary multiobjective algorithms to evolve main solutions and invoke the tabu search algorithm:
\begin{itemize}
\item NSGA-II \cite{Deb02TEVC}.
\item MOEA/D \cite{ZhangQ07TEC}.
\item A differential evolution for multiobjective optimization with self-adaptation (DEMOwSA) \cite{Zamu07CEC}.
\item A multiobjective particle swarm optimization (MOPSO) algorithm \cite{Zheng14TEVC} which extends comprehensive learning \cite{Liang06TEC} for multiobjective optimization.
\end{itemize}

\subsection{Complexity Analysis}
In this subsection, we theoretically compare the complexities of the original problem and the transformed problem. For notational simplicity, we regard suspected infection as a disease with subscript $i\!=\!0$. The number of all possible solutions to the original problem is
\begin{equation}
N= \prod_{k=1}^{n\!+\!n'}(\overline{x_k}-\underline{x_k})
\label{eq:ns}\end{equation}

And the algorithm \ref{alg:queue} for evaluating each solution to the original problem has a time complexity
\begin{equation}
O(f)= \sum_{i=0}^m\sum_{j=0}^{J_i}r_i|\Phi_{i,j}|
\label{eq:cf}\end{equation}

After transformation, the number of all possible solutions to the new problem is
\begin{equation}
N'= \prod_{i=0}^m(\overline{y_i}-\underline{y_i})
\label{eq:ns'}\end{equation}

To evaluate each solution to the transformed problem, we need to solve $m\!+\!1$ subproblems. The number of possible solutions to the $i$-th subproblem is
\begin{equation}
N_i= \prod_{j=0}^{J_i}\prod_{k=1}^{|\Phi_{i,j}|}(\overline{z_{j,k}}-\underline{z_{j,k}})
\label{eq:nss}\end{equation}
where $\overline{z_{j,k}}$ and $\underline{z_{j,k}}$ denote the upper and lower limits of decision variable $z_{j,k}$.

And the time complexity of evaluating each solution to the $i$-th subproblem is
\begin{equation}
O(f_i)= \sum_{j=0}^{J_i}r_i|\Phi_{i,j}|
\label{eq:cfs}\end{equation}

Consequently, the total time complexity of the transformed problem is $N'\big(\sum_{i=0}^mN_iO(f_i)\big)$, while that of the original problem is $N\!\cdot\!O(f)$, and the complexity reduction ratio of transformation is
\begin{equation}
R_c= \frac{\log\big(N\!\cdot\!O(f)\big)}{\log\big(N'(\sum_{i=0}^m N_iO(f_i))\big)}
\end{equation}

As the expressions of $N$, $N'$, $N_i$, $O(f)$ and $O(f_i)$ are complicated, in practice, we can use the average range $\widehat{x}$ of all $(\overline{x_k}-\underline{x_k})$ in Eq. (\ref{eq:ns}), the average $\widehat{|\Phi|}$ of all $\big(\sum_{j=0}^{J_i}|\Phi_{i,j}|\big)$ in Eqs. (\ref{eq:cf}) and (\ref{eq:cfs}), the average $\widehat{y}$ of all $(\overline{y_i}-\underline{x_i})$ in Eq. (\ref{eq:ns'}), and the average $\widehat{z}$ of all $(\overline{z_{j,k}}-\underline{z_{j,k}})$ in Eq. (\ref{eq:nss}). Let $r\!=\!\sum\limits_{i=0}^m r_i$ be the number of all cases, we have
\begin{eqnarray}
R_c&\approx& \frac{\log\big(\widehat{x}^{n\!+\!n'}r\widehat{|\Phi|}\big)} {\log\big(\widehat{y}^m \widehat{z}^{\widehat{|\Phi|}}r\widehat{|\Phi|}\big)} \nonumber\\
&=& \frac{(n\!+\!n')\log\widehat{x}+\log(r\widehat{|\Phi|})} {m\log\widehat{y}+\widehat{|\Phi|}\log\widehat{z}+\log(r\widehat{|\Phi|})}
\end{eqnarray}

In our case studies, the average values are $\widehat{x}\!\approx\!1100$, $\widehat{y}\!\approx\!95$, $\widehat{z}\!\approx\!66$, $\widehat{|\Phi|}\!\approx\!37$, $n\!+\!n'\!\approx\!25000$, and $m\!\approx\!450$. Therefore, the average complexity reduction ratio on the instances is approximately 58.

\section{Computational Results}\label{sec:exp}
\subsection{Problem Instances}
We use the proposed method for medical supplies procurement in Zhejiang Hospital of Traditional Chinese Medicine (ZJHTCM) from 15 Feb to 15 Apr, 2020, the peak of COVID-19 in Zhejiang Province, China. Since 15 Mar, we also extend the application to other five hospitals (denoted by H1--H5).  Therefore, there are 14 real-world instances of the medical supplies procurement problem. Table \ref{tab:ins} summarizes the main characteristics of the instances, where $\sum_ir_i$ denotes the total expected number of cases of all common diseases, $\overline{J}$ denotes the average treatment items per disease, $\overline{|\Phi|}$ denotes the average number of alternatives per treatment item, and the budget $C$ is in RMB. The instances are solved on a workstation with an i7-6500 2.5GH CPU, 8GB DDR4 RAM, and an NVIDIA Quadro M500M card.

\begin{table*}
\center\footnotesize
\caption{Summary of the real-world instances of the medical supplies procurement problem.}
\begin{tabular}{cc|rrrrrrrrr}\hline\
Hospital &Period &$m$ &$n\!+\!n'$ &$\sum_ir_i$ &$r_0$ &\multicolumn{1}{c}{$\overline{J}$} &$\overline{|\Phi|}$ &\multicolumn{1}{c}{$C$} \\\hline
\multirow{4}{0.9cm}{ZJHTCM}& 2$^\textnormal{nd}$ half Feb &476 &32,535 &71,196 &64 &5.84 &7.27 &3,516,000\\
 &1$^\textnormal{st}$ half Mar& 476 & 32,416 &76,580 & 38 & 5.84 &7.27 & 3,378,000\\
 &2$^\textnormal{nd}$ half Mar& 479 & 32,628 &78,331 & 34 & 5.86 &7.29 & 3,022,000\\
 &1$^\textnormal{st}$ half Apr& 479 & 32,628 &90,459 & 36 & 5.86 &7.32 & 3,698,000\\\hline
\multirow{2}{0.9cm}{H1}& 2$^\textnormal{nd}$ half Mar& 162 &17,522 &8,208 & 4 &7.46 &5.41 &521,000\\
 &1$^\textnormal{st}$ half Apr& 162 &17,510 &13,640 & 3 & 7.46 & 5.41 & 830,000\\\hline
\multirow{2}{0.9cm}{H2}& 2$^\textnormal{nd}$ half Mar& 193 &15,666 &17,353 &24 &8.06 &5.25 &785,000\\
 &1$^\textnormal{st}$ half Apr& 193 &15,681 &19,309 & 14 & 8.06 & 5.25 & 902,500\\\hline
\multirow{2}{0.9cm}{H3}& 2$^\textnormal{nd}$ half Mar& 328 &24,469 &32,052 &14 &7.84 &5.87 &1,682,000\\
 &1$^\textnormal{st}$ half Apr& 328 &24,469 &42,667 & 17 & 7.84 & 5.97 & 2,127,000\\\hline
\multirow{2}{0.9cm}{H4}& 2$^\textnormal{nd}$ half Mar& 393 &27,600 &35,733 &50 &6.90 &6.13 &2,415,500\\
 &1$^\textnormal{st}$ half Apr& 399 &27,215 &38,452 & 28 & 6.87 & 6.14 & 2,607,200\\\hline
\multirow{2}{0.9cm}{H5}& 2$^\textnormal{nd}$ half Mar& 573 &35,906 &60,900 &27 &6.66 &5.36 &3,920,000\\
 &1$^\textnormal{st}$ half Apr& 573 &34,902 &75,393 & 30 & 6.66 & 5.48 & 4,818,000 &\\\hline
\end{tabular}
\label{tab:ins}\end{table*}

\subsection{Performance for Solving the Subproblem}
Before testing the algorithms for solving the main problem, we first test the performance of the tabu search algorithm for subproblems. From the above real-world main problem instances, we select 16 subproblem instances, the dimensions $D$ of which range from 12 to 72. For the algorithm, we set the neighborhood size $k_N$ to $2D$, tabu length to 12, and the maximum number of iterations to $50D$.  On each instance, we run the algorithm 50 times to test whether and how long it can obtain the exact optimal solution (validated by an exact branch-and-bound algorithm \cite{Derp06EJOR}).

\begin{figure*}[!t]
\centering
\subfigure[$D\!=\!12$]{\includegraphics[scale=0.26]{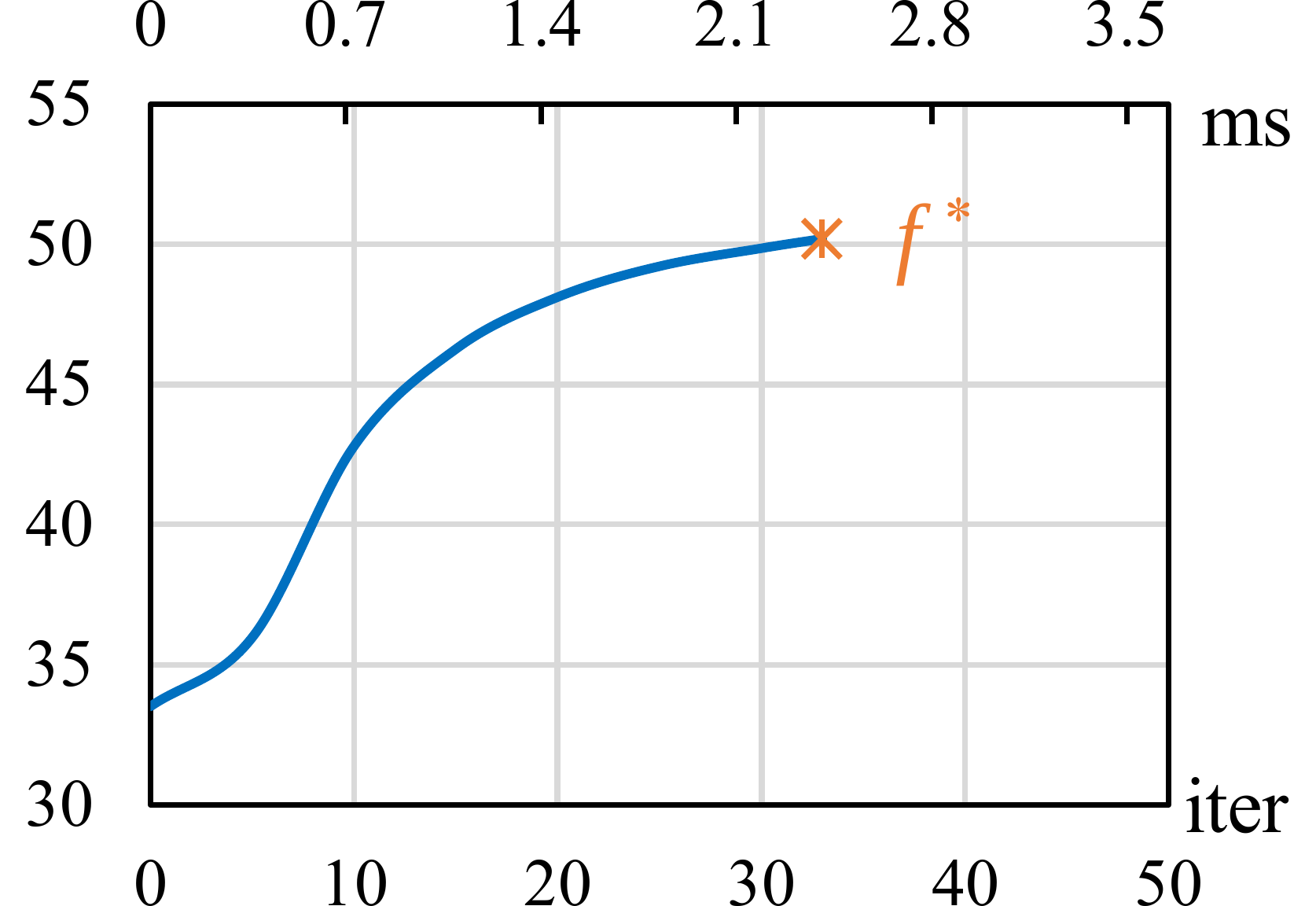}}
\subfigure[$D\!=\!15$]{\includegraphics[scale=0.26]{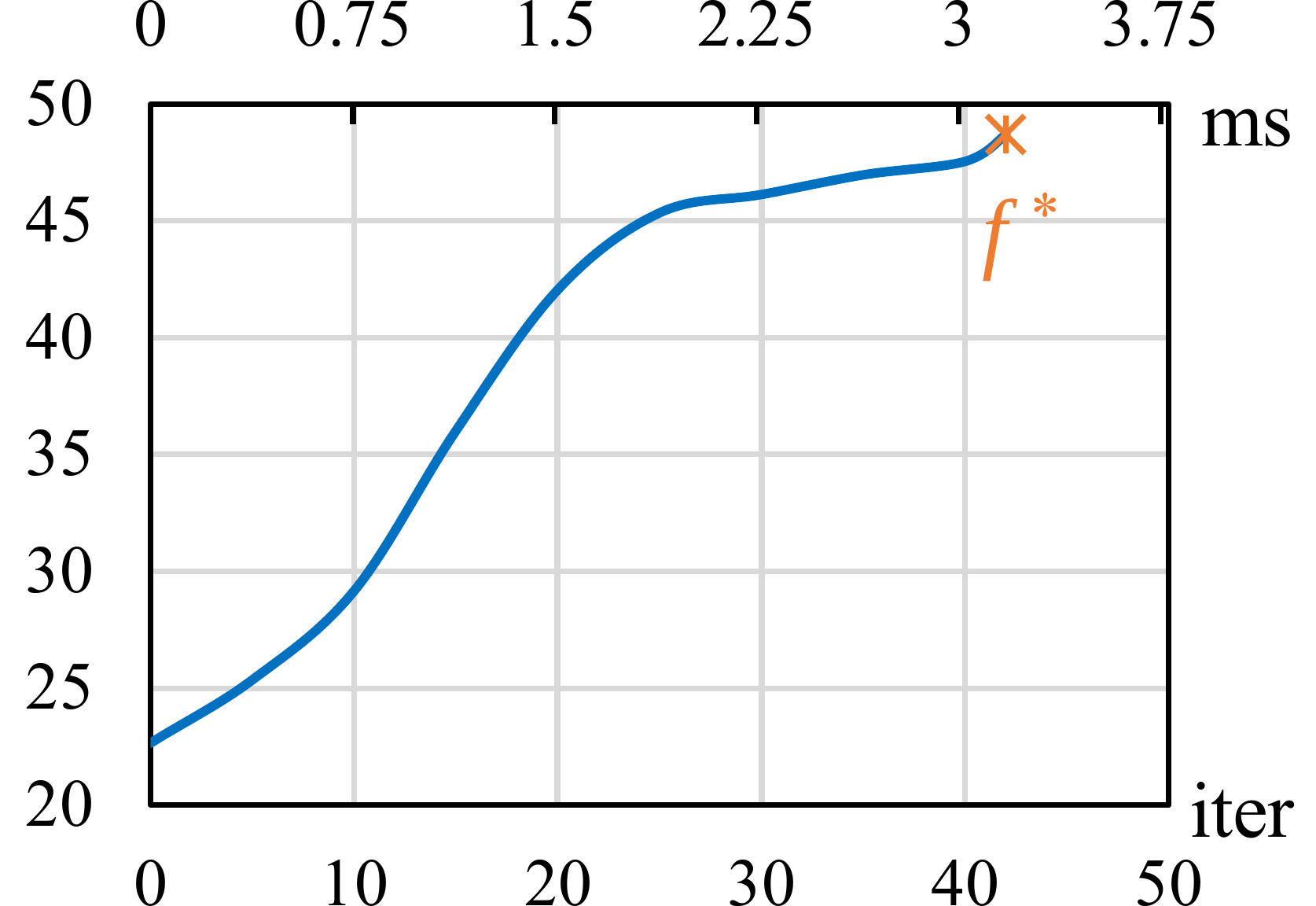}}
\subfigure[$D\!=\!18$]{\includegraphics[scale=0.26]{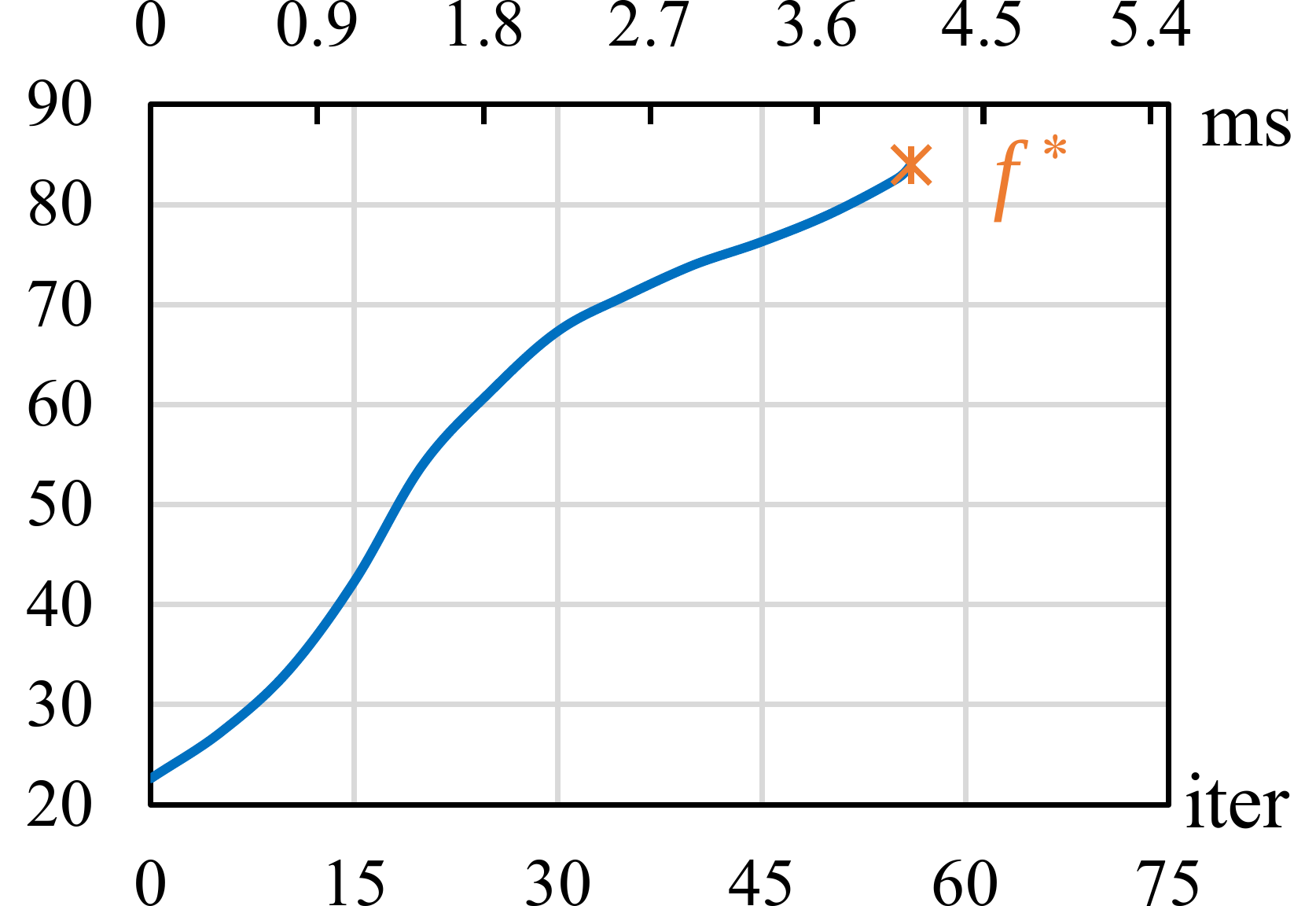}}
\subfigure[$D\!=\!21$]{\includegraphics[scale=0.26]{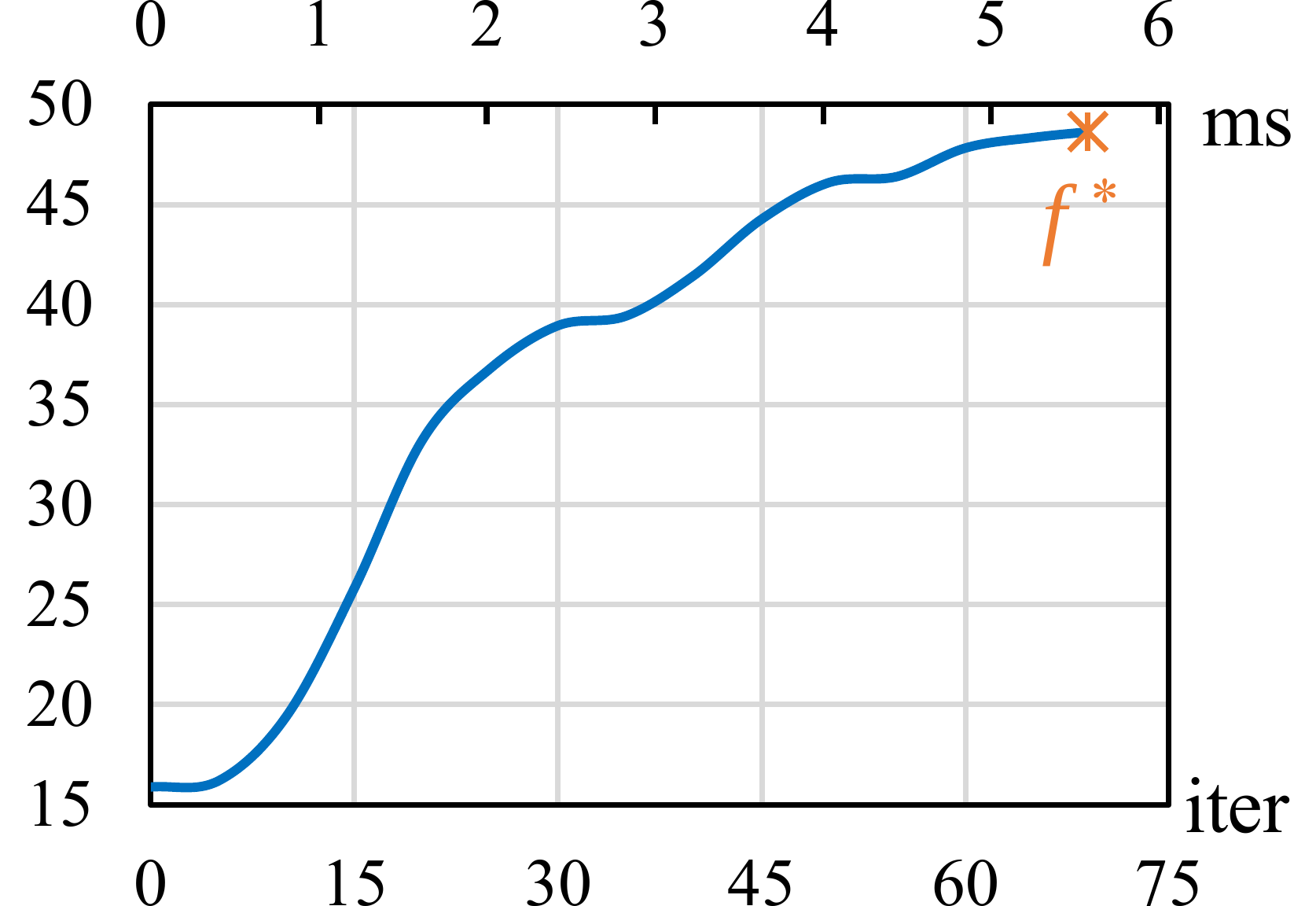}}
\hfil
\subfigure[$D\!=\!24$]{\includegraphics[scale=0.26]{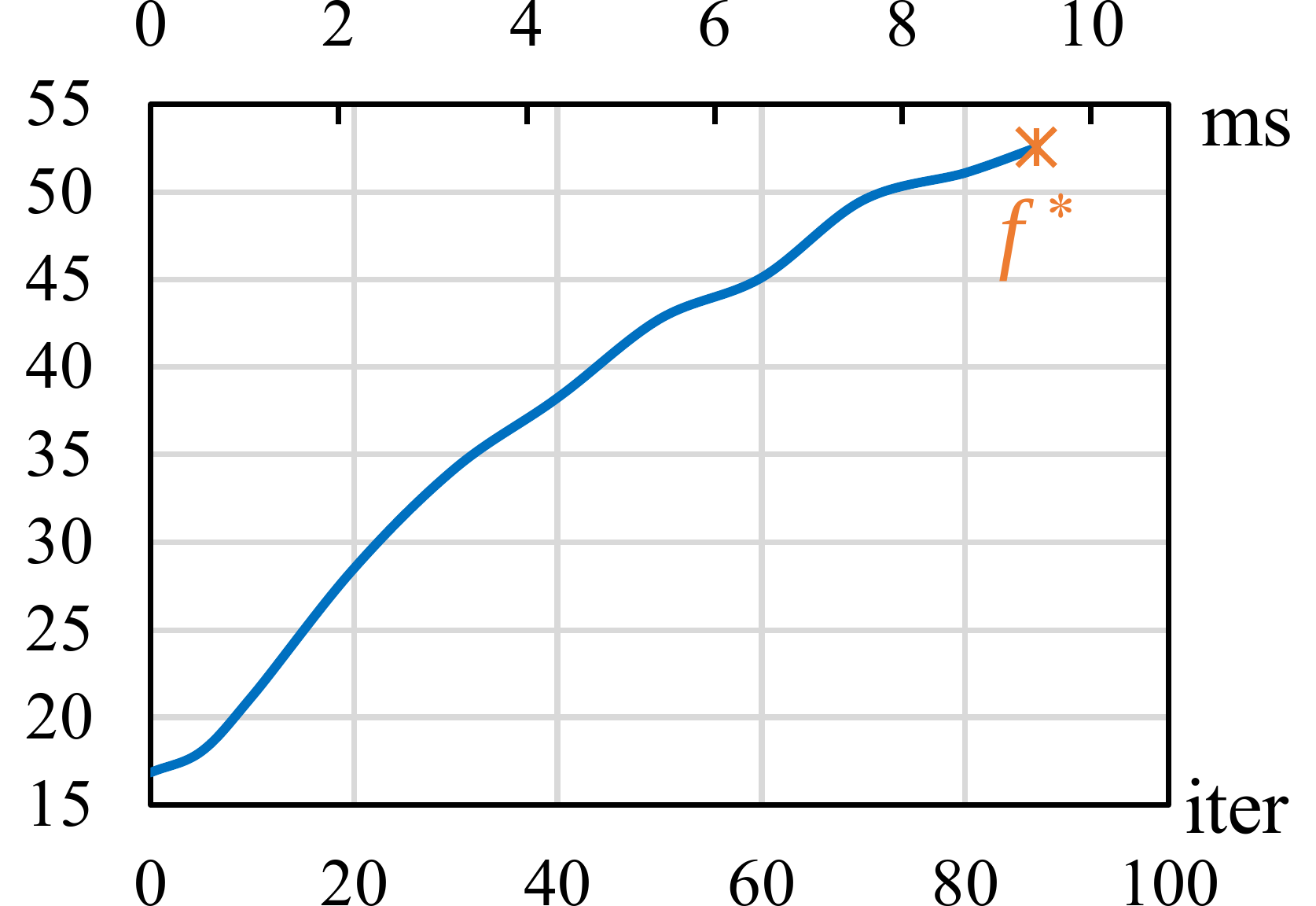}}
\subfigure[$D\!=\!27$]{\includegraphics[scale=0.26]{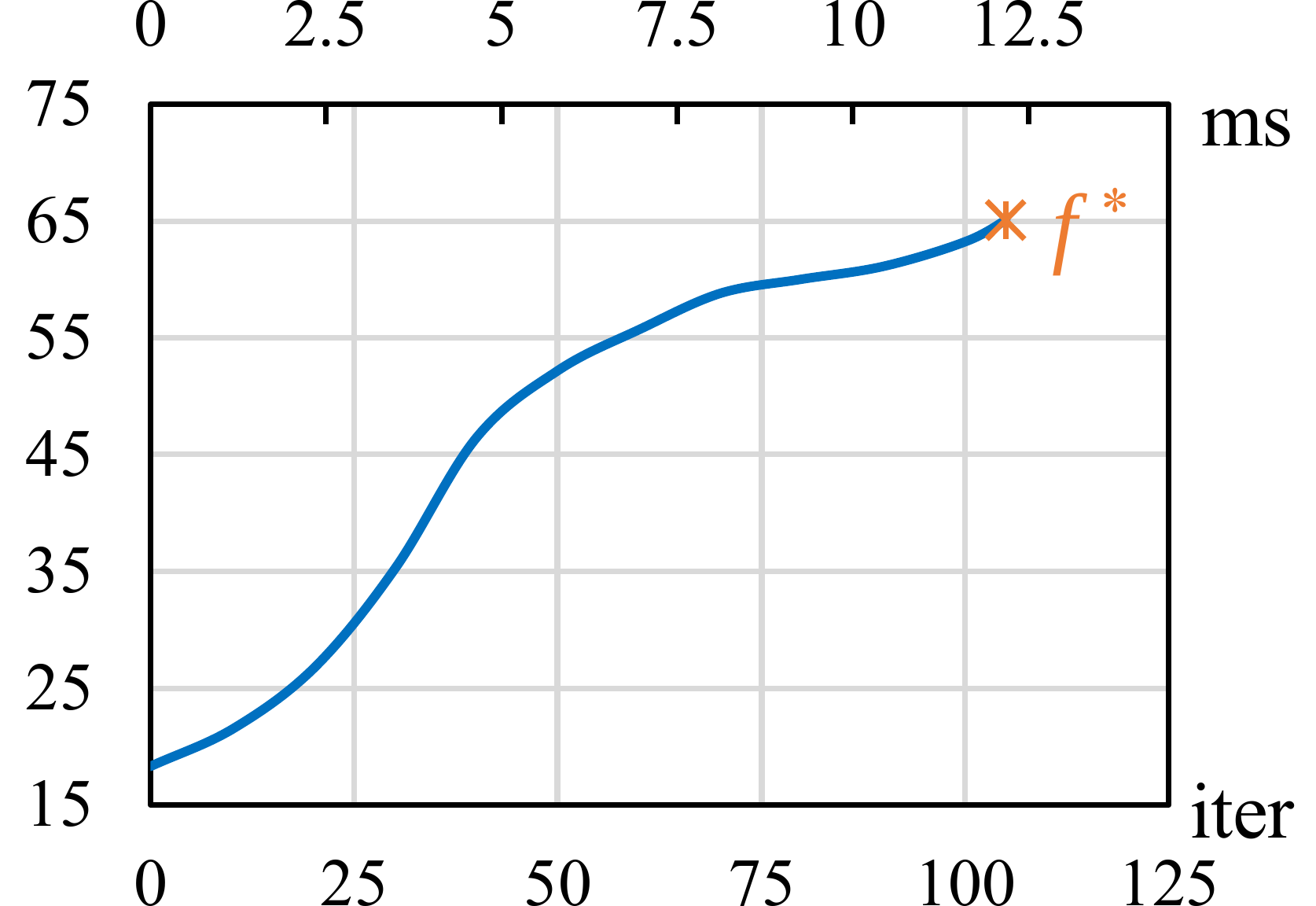}}
\subfigure[$D\!=\!30$]{\includegraphics[scale=0.26]{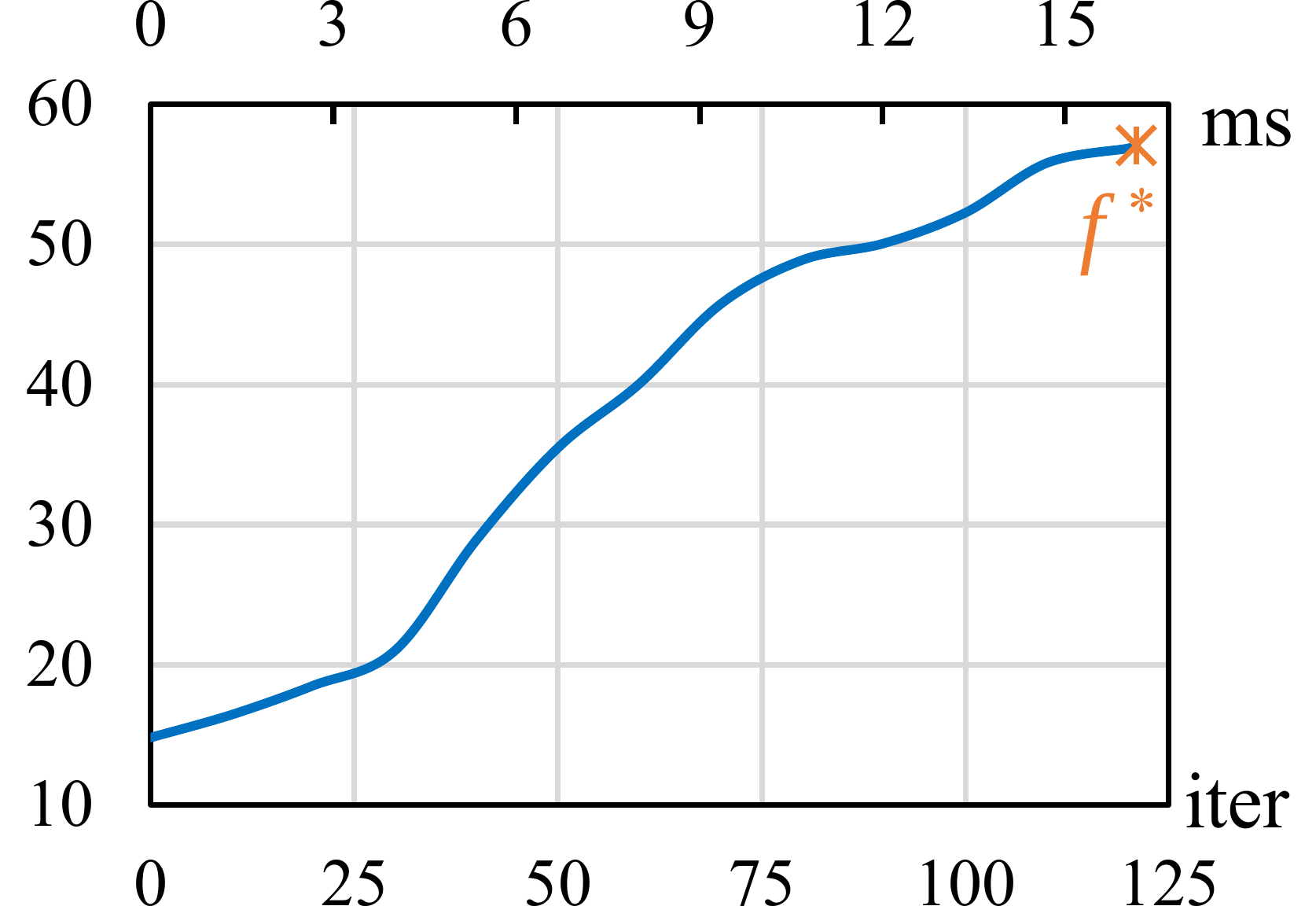}}
\subfigure[$D\!=\!34$]{\includegraphics[scale=0.26]{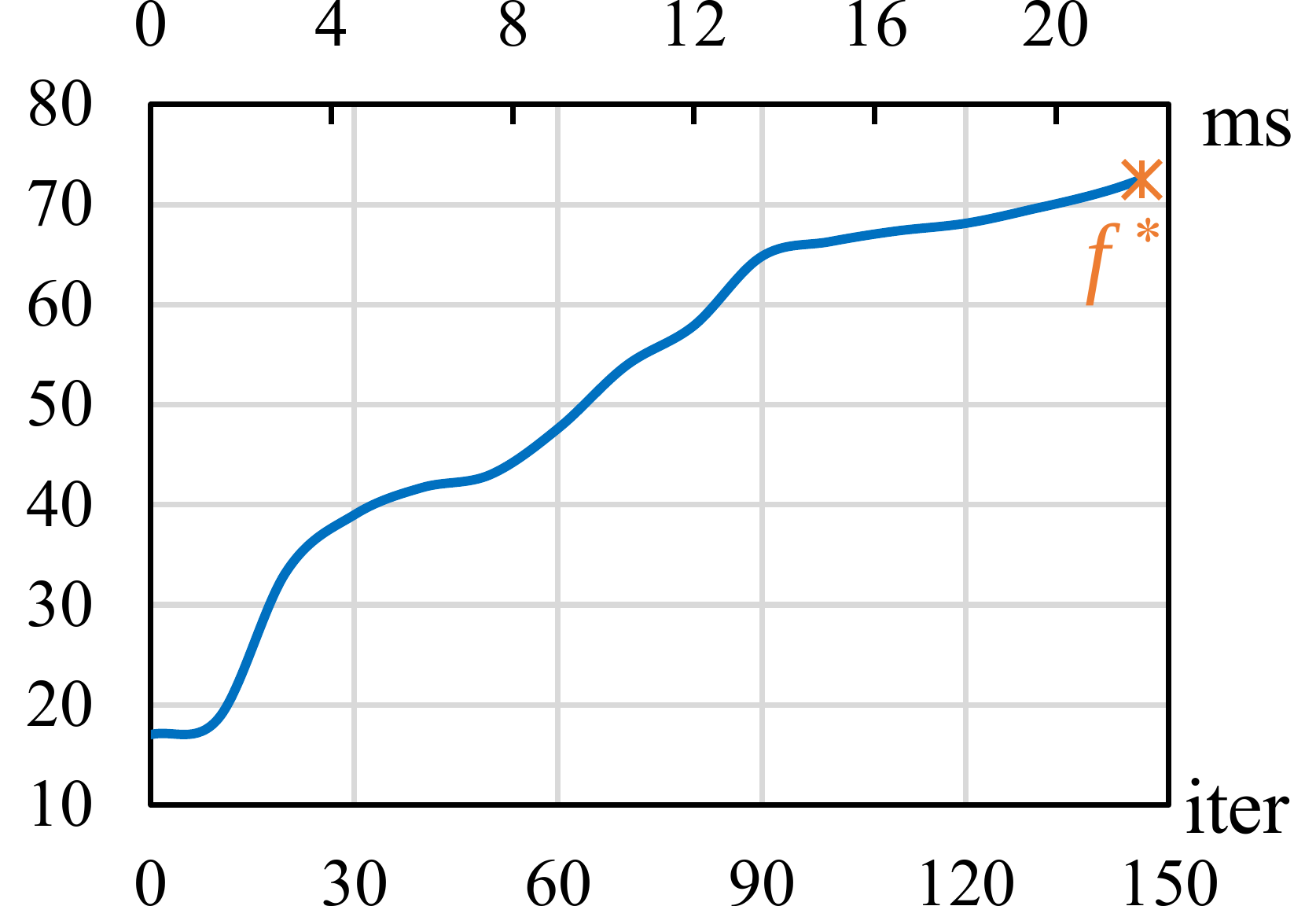}}
\hfil
\subfigure[$D\!=\!38$]{\includegraphics[scale=0.26]{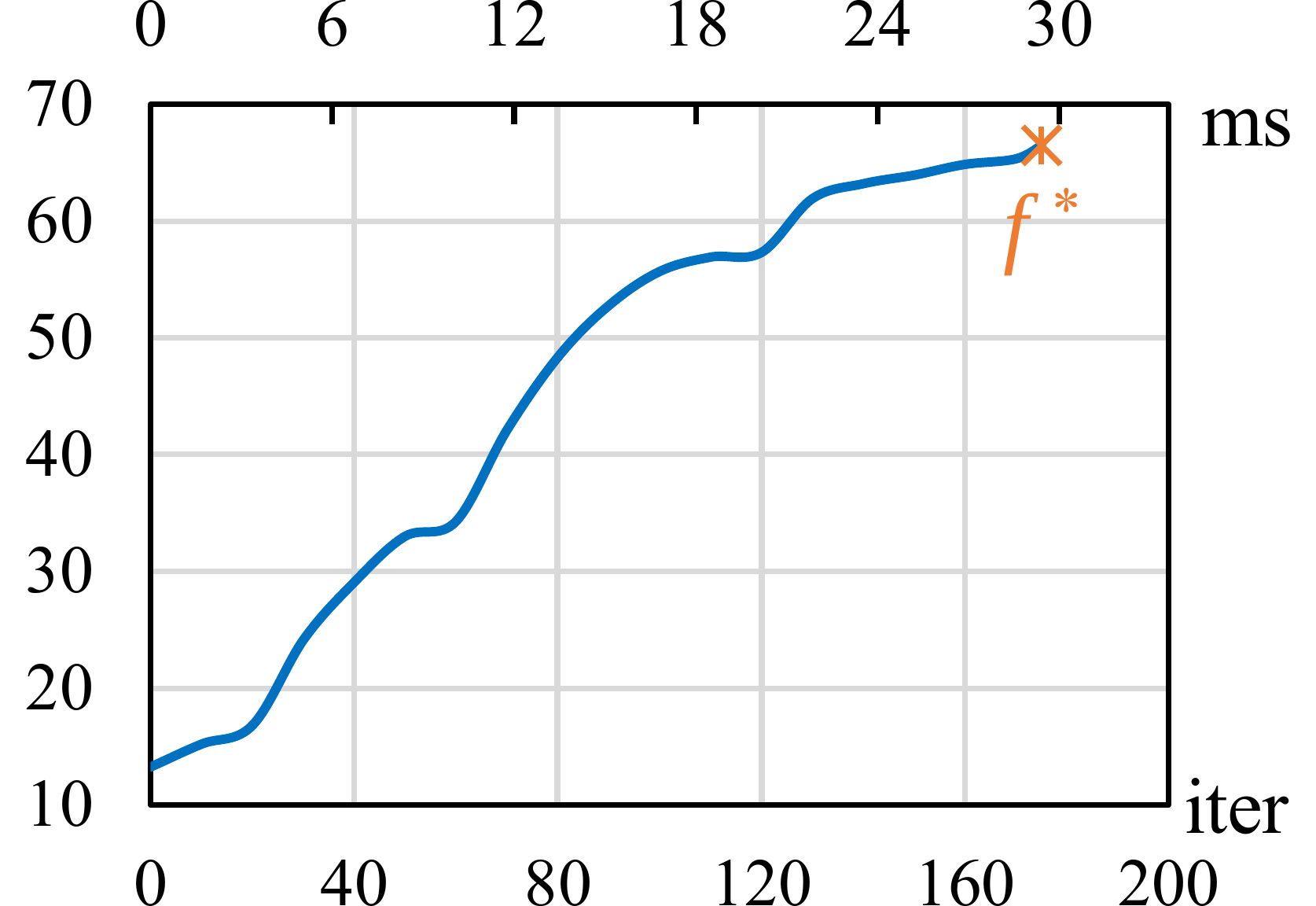}}
\subfigure[$D\!=\!42$]{\includegraphics[scale=0.26]{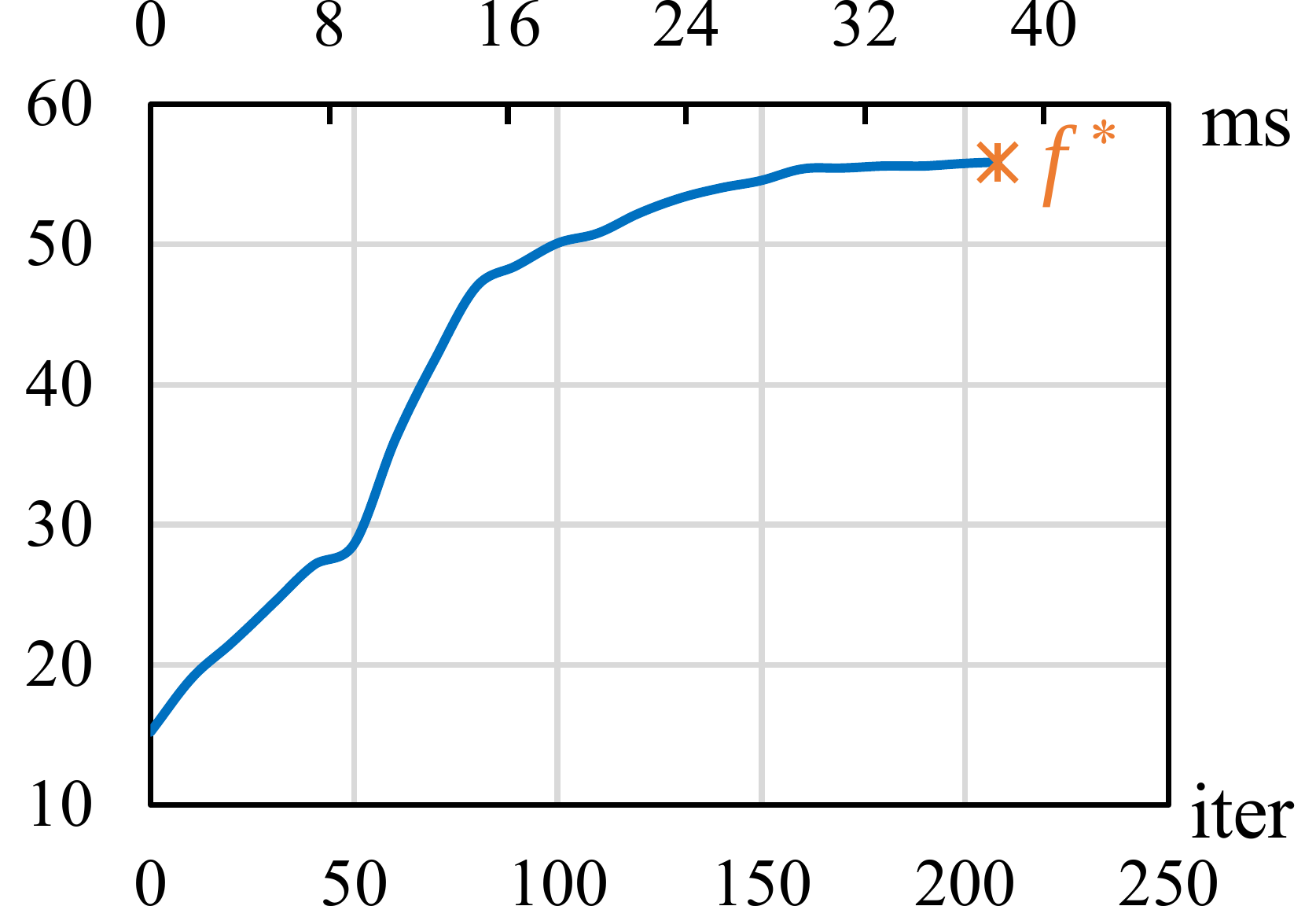}}
\subfigure[$D\!=\!46$]{\includegraphics[scale=0.26]{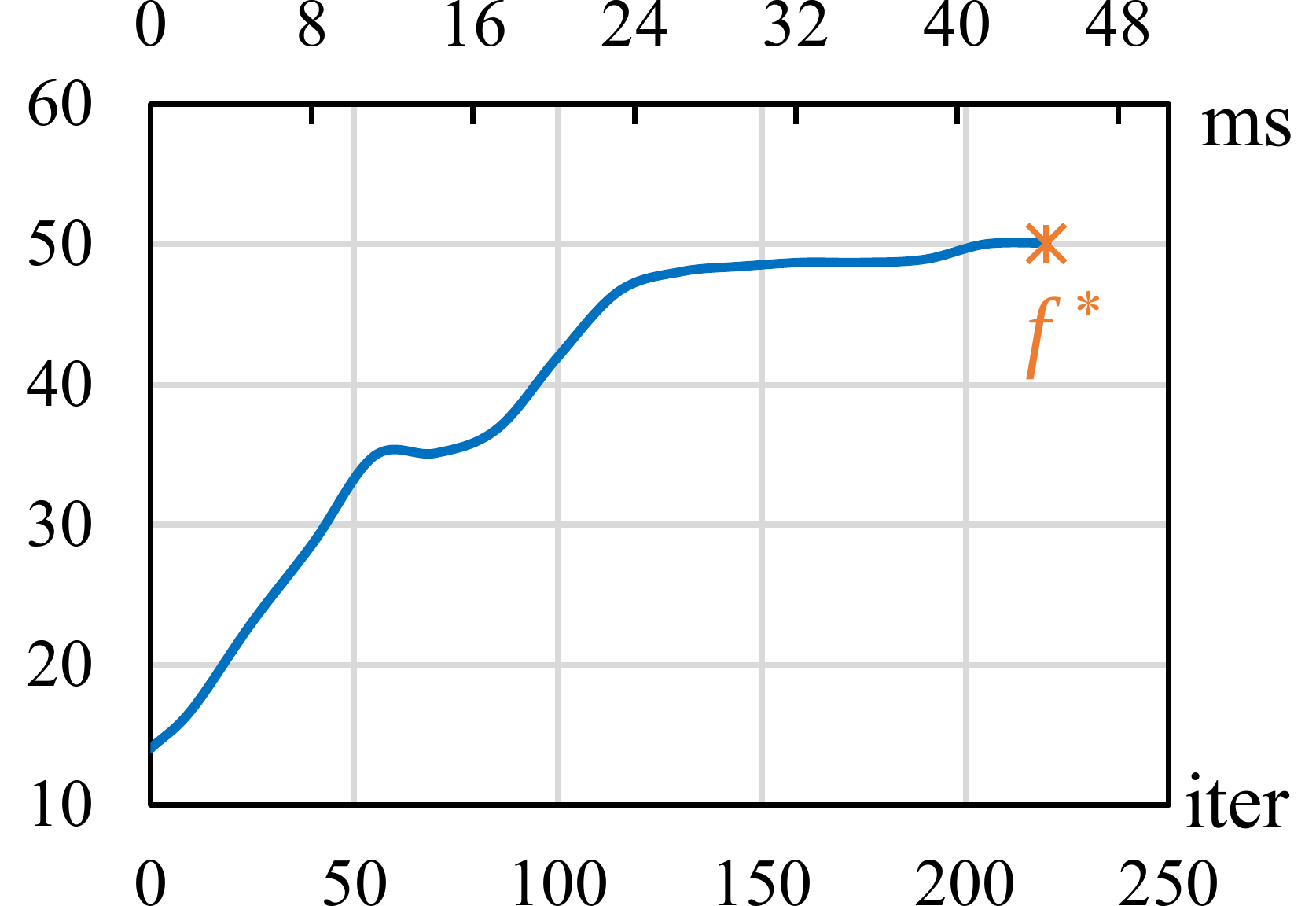}}
\subfigure[$D\!=\!50$]{\includegraphics[scale=0.26]{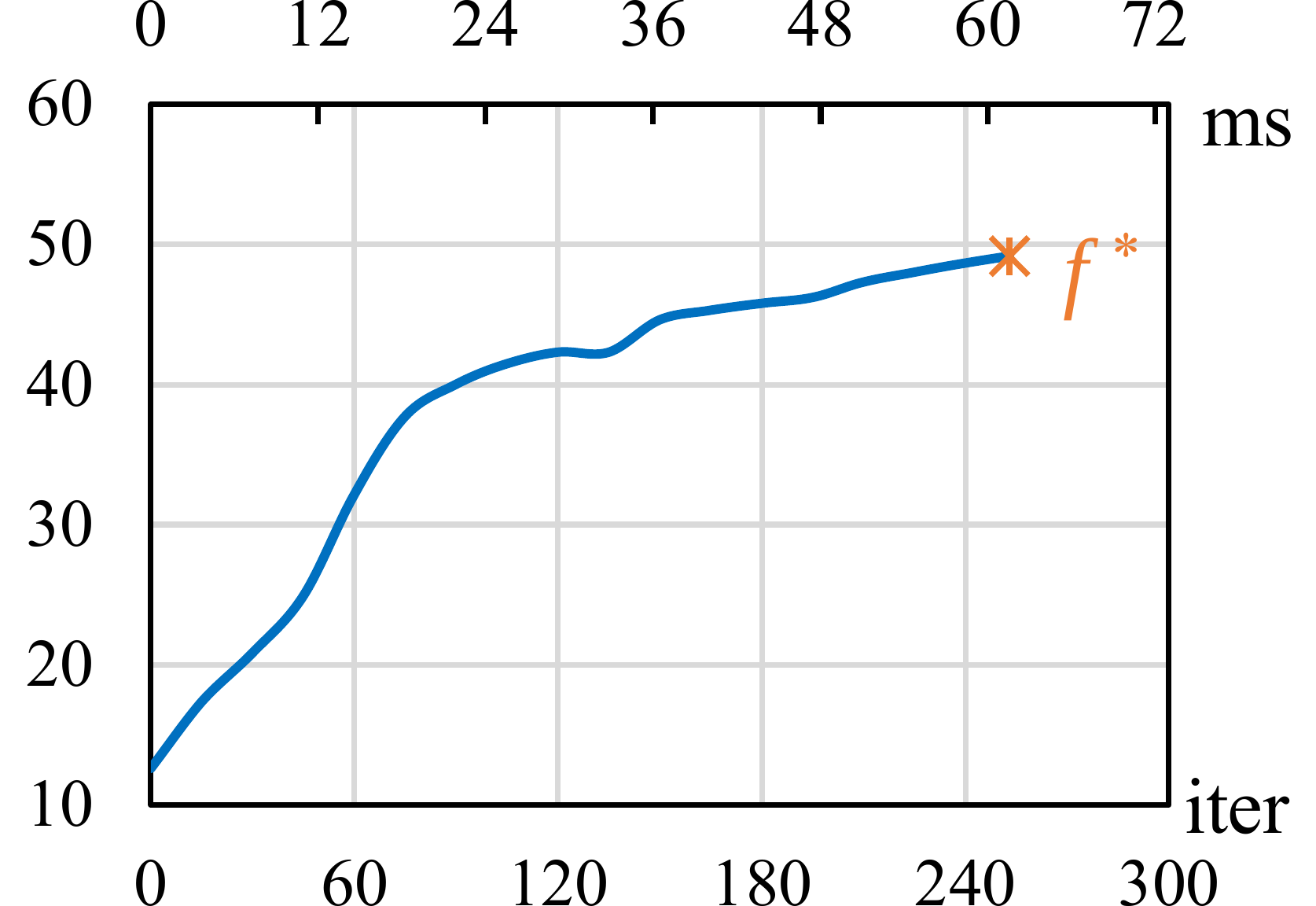}}
\hfil
\subfigure[$D\!=\!55$]{\includegraphics[scale=0.26]{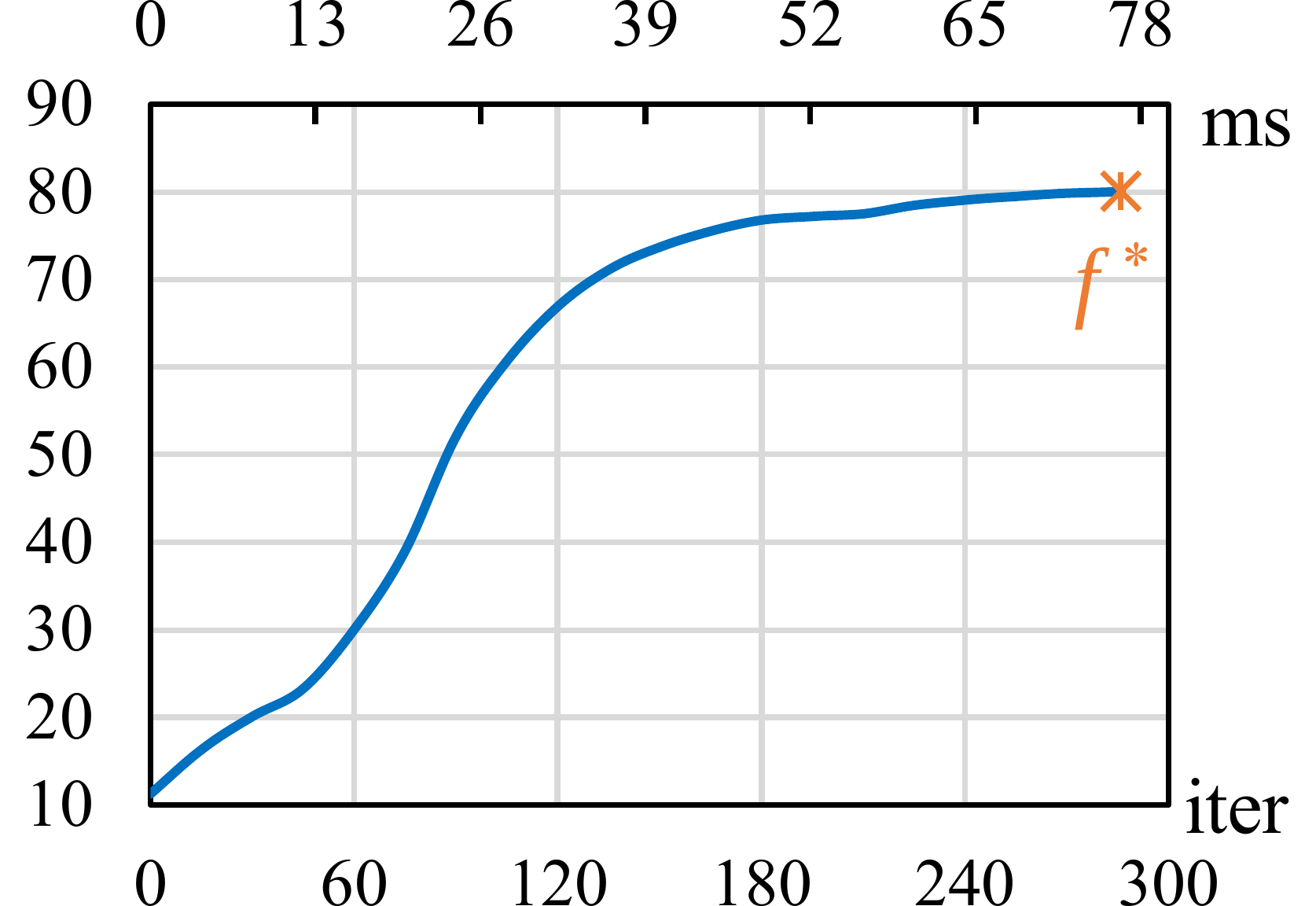}}
\subfigure[$D\!=\!60$]{\includegraphics[scale=0.26]{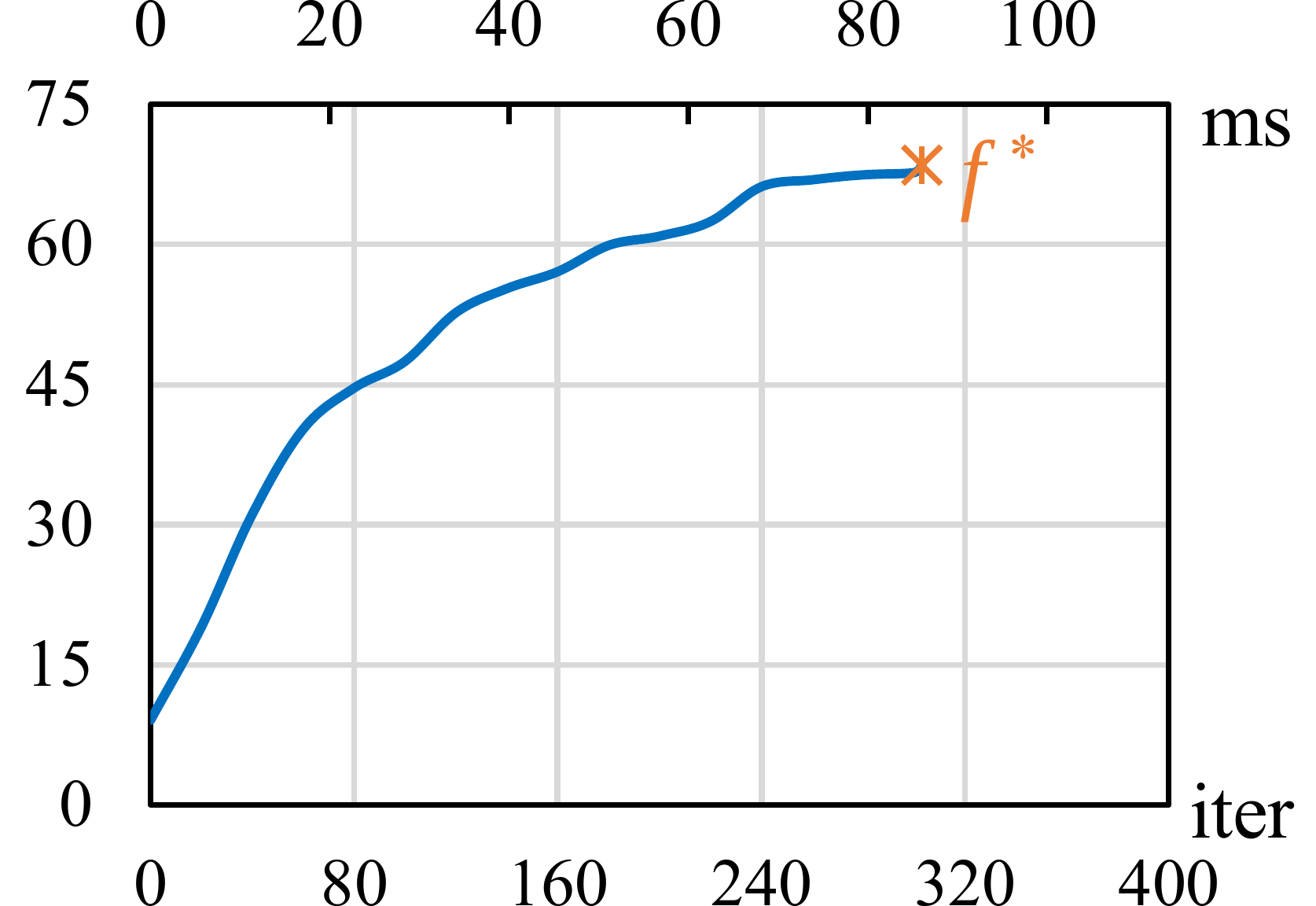}}
\subfigure[$D\!=\!66$]{\includegraphics[scale=0.26]{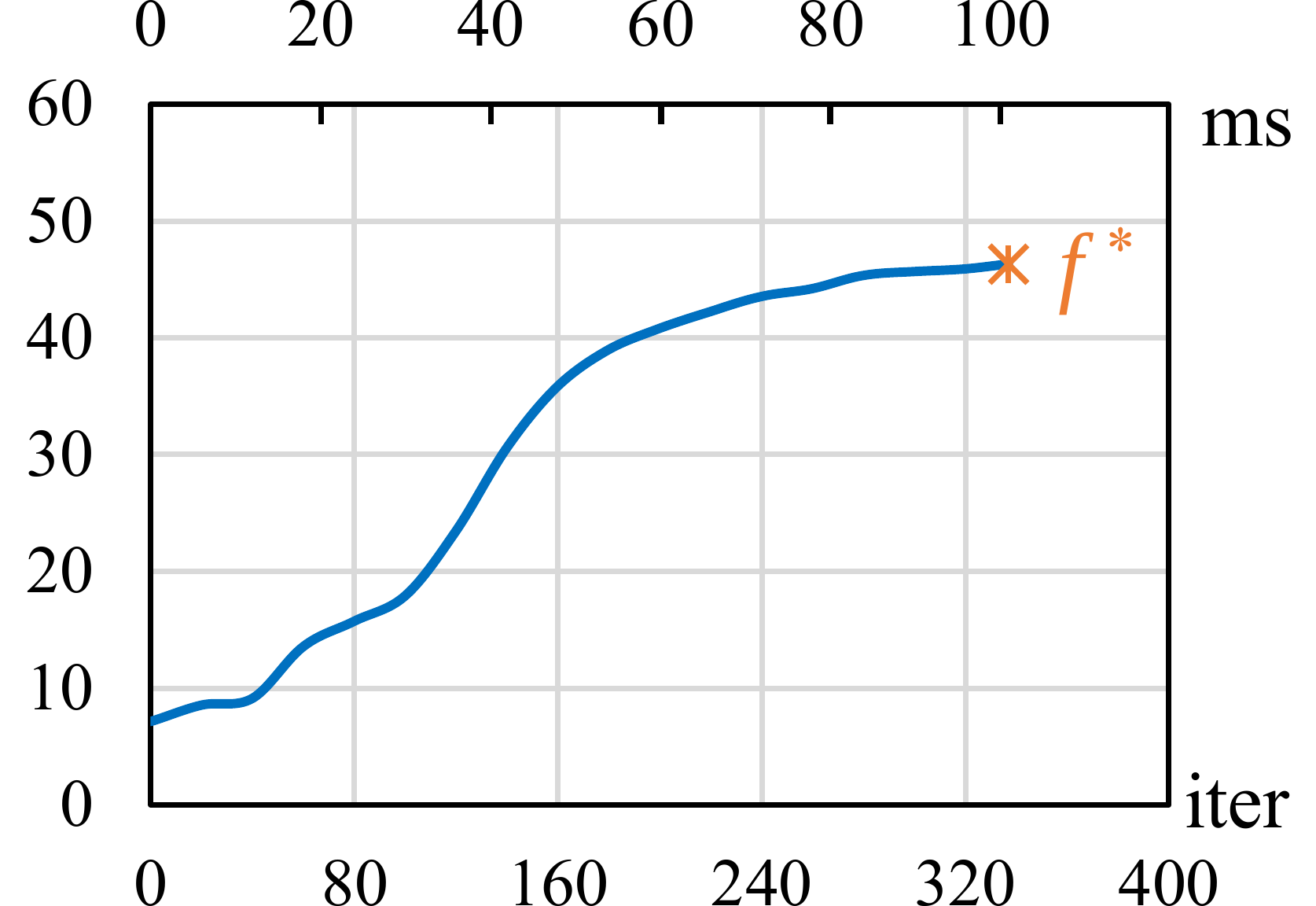}}
\subfigure[$D\!=\!72$]{\includegraphics[scale=0.26]{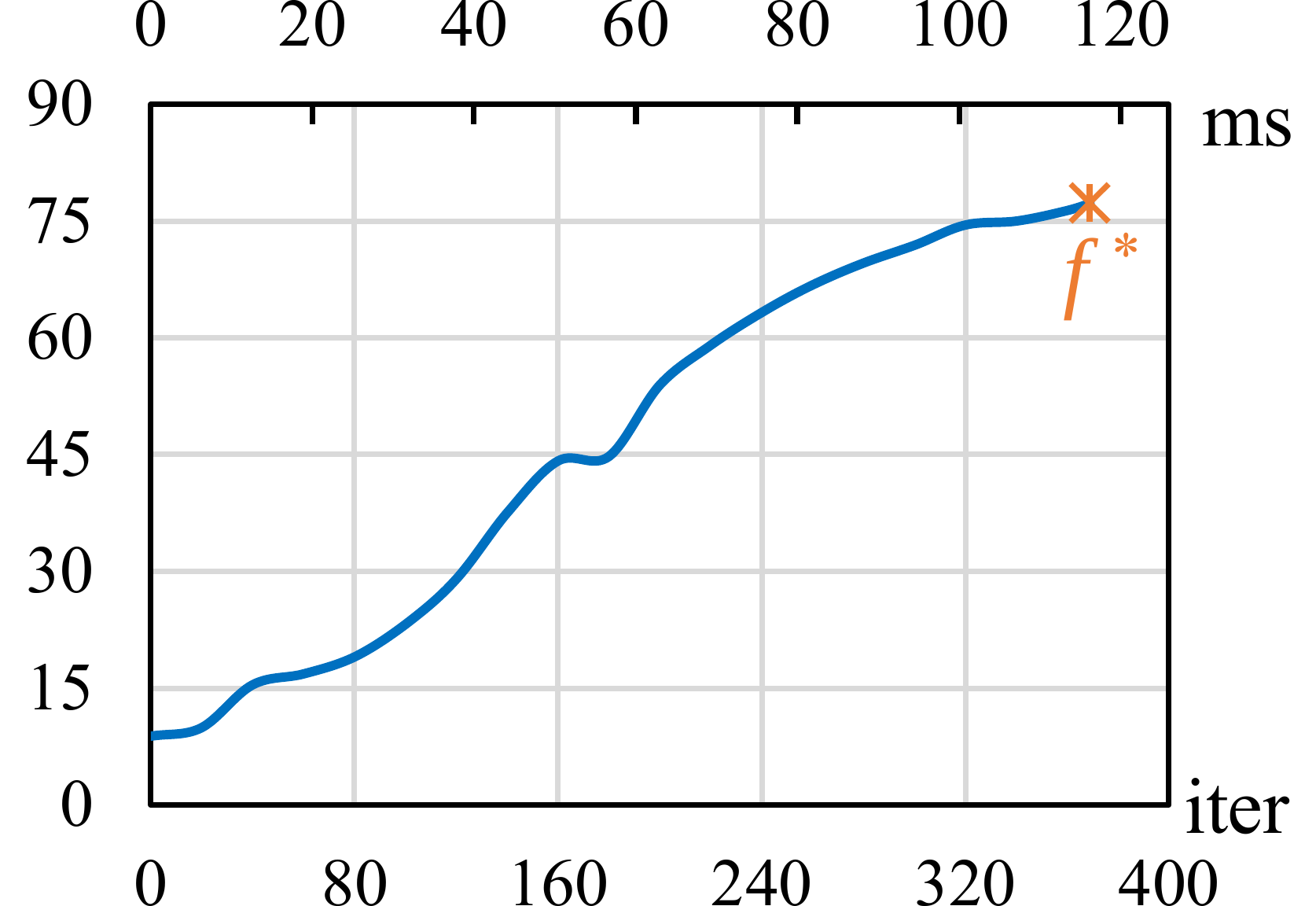}}
\caption{Convergency curves of the tabu search algorithm on subproblem instances. The bottom horizontal axis is the number of iterations, the top horizontal axis is the CPU time (in milliseconds), the vertical axis is the objective function value, and $f^*$ is the exact optimal objective function value.}
\label{fig:ts}\end{figure*}

Fig. \ref{fig:ts} presents the convergency curves (averaged over the 50 runs) of the tabu search algorithm on the subproblem instances. The algorithm reaches the optima within 100 iterations (10 ms in our computing environment) when the problem dimension is smaller than 24, within 200 iterations (30 ms) when the dimension is smaller than 40, and within 400 iterations (120 ms) on all instances. In our case studies, the average dimension of the instances is approximately 37, which can be solved using approximately 160 iterations (25 ms); the dimension of the largest instance is 72, which can be solved using 369 iterations (116 ms). Using multithreading and GPU acceleration, the average CPU time for evaluating a main solution to a problem of 400 diseases is approximately 600 ms.

\subsection{Performance for Solving the Original and Transformed Problems}
For each main problem instance, we use five evolutionary constrained multiobjective algorithms, including NSGA-II with constraint handling (denoted by NSGA-II-C) \cite{Deb02TEVC}, CMOEA \cite{Wold09TEC}, MOEA/D with constraint handling (denoted by MOEA/D-C) \cite{Jan10UKCI}, DECMOSA \cite{Zamu09CEC}, and D$^2$MOPSO \cite{Moub14EC}, to solve the original problem; we also use four evolutionary multiobjective algorithms, including NSGA-II \cite{Deb02TEVC}, MOEA/D \cite{ZhangQ07TEC}, DEMOwSA \cite{Zamu07CEC}, and MOPSO \cite{Zheng14TEVC}, all combined with tabu search, to solve the transformed problem. The control parameters of all algorithms are tuned on the whole set of instances. For a fair comparison, all the algorithms use the same stopping criterion that the CPU time does not exceed 90 minutes, which is also applied in our practice. On each instance, each algorithm is run 30 times.

\begin{figure*}[!t]
\centering
\subfigure[ZJHTCM, 1$^\textnormal{st}$ half Mar]{\includegraphics[scale=0.281]{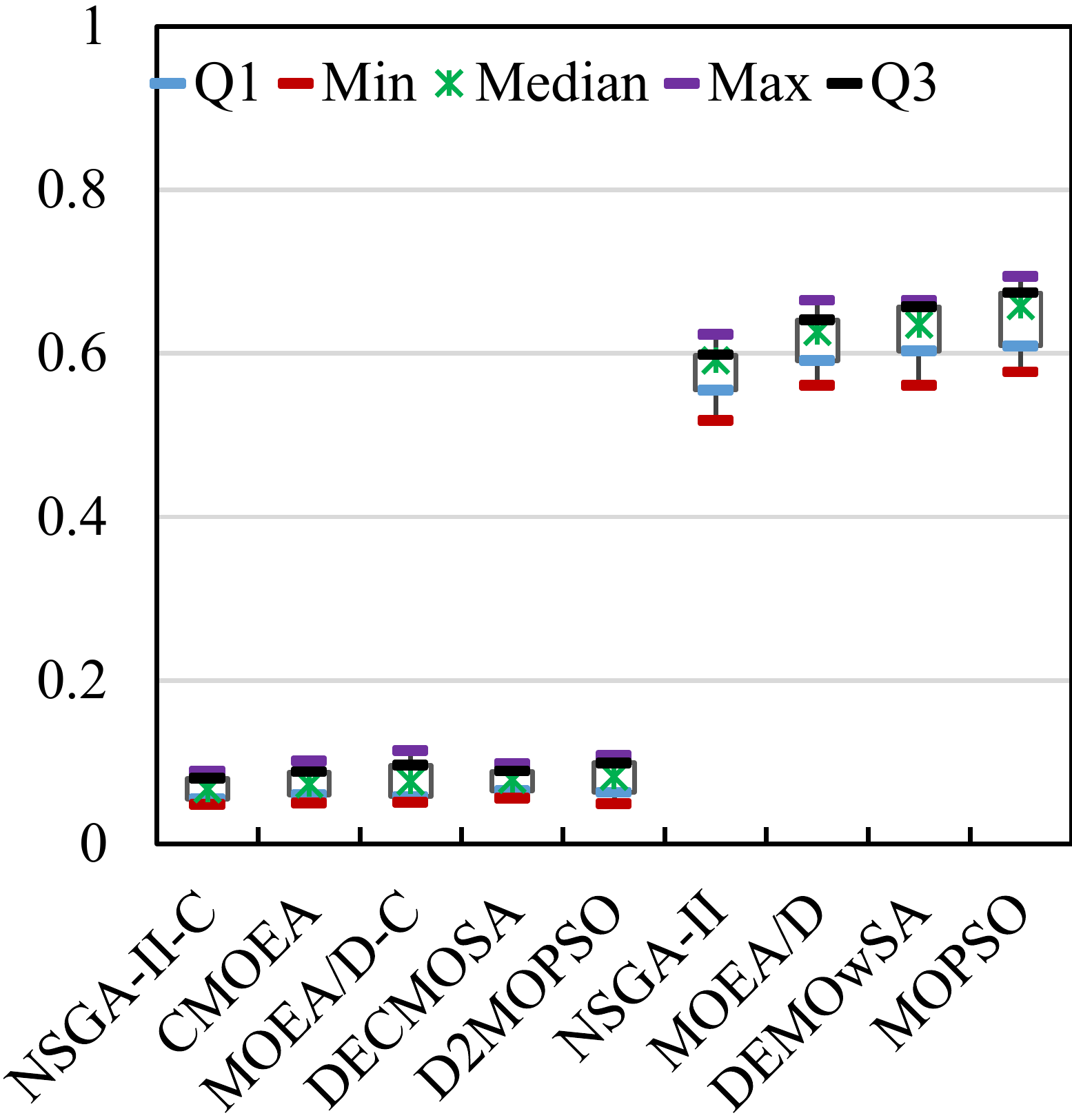}}
\subfigure[ZJHTCM, 2$^\textnormal{nd}$ half Mar]{\includegraphics[scale=0.281]{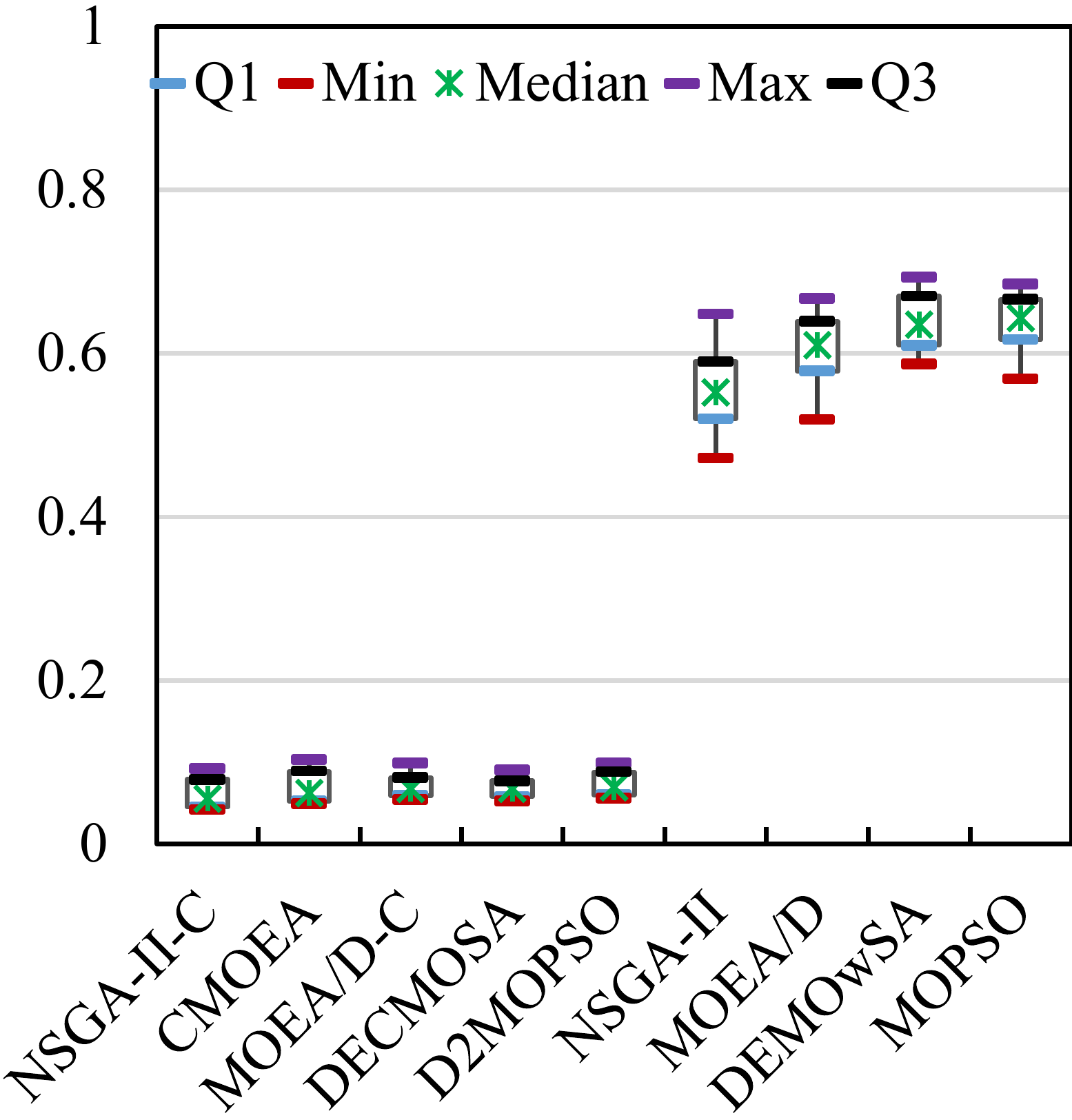}}
\subfigure[ZJHTCM, 1$^\textnormal{st}$ half Apr]{\includegraphics[scale=0.281]{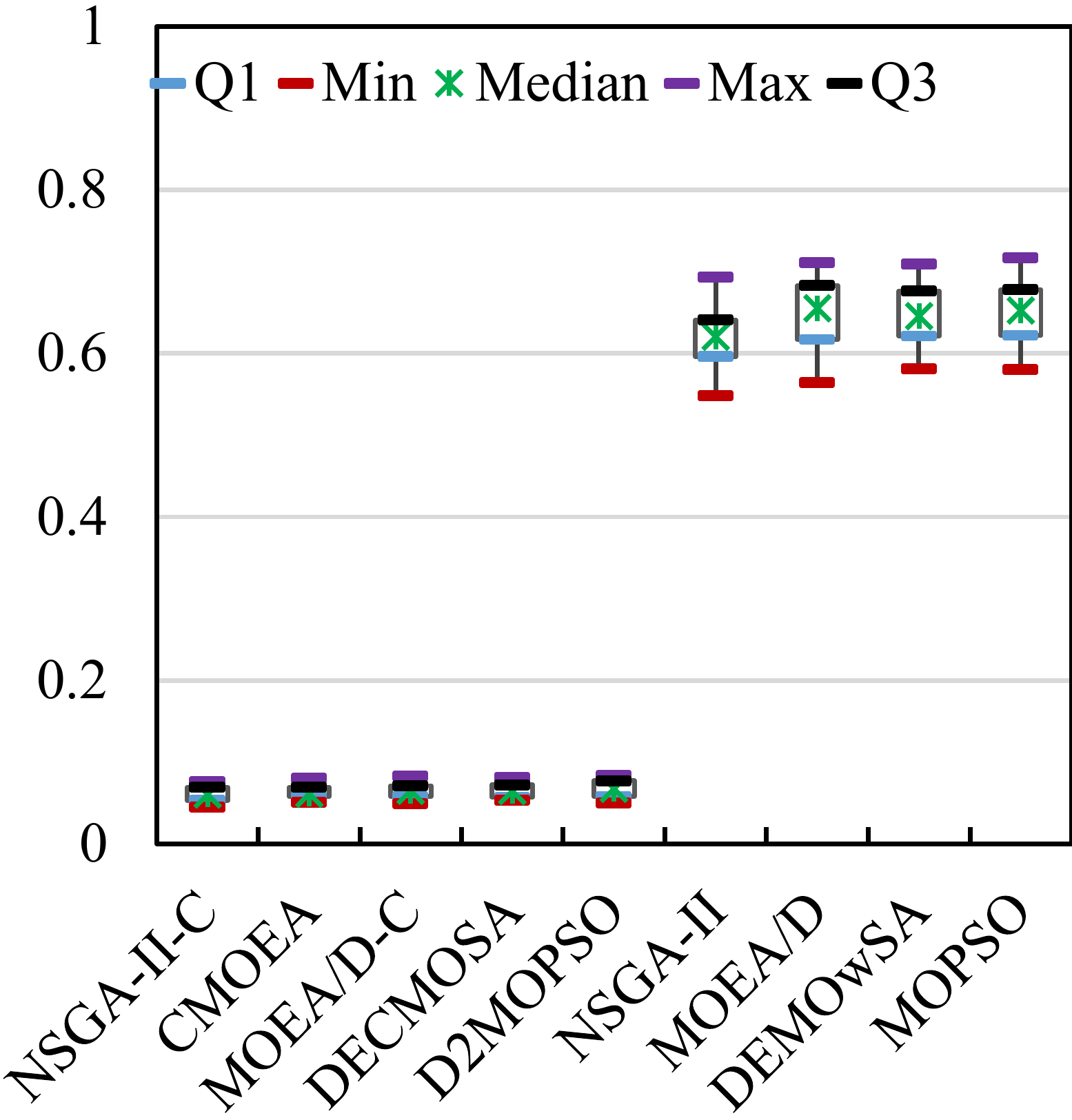}}
\subfigure[ZJHTCM, 1$^\textnormal{st}$ half Apr]{\includegraphics[scale=0.281]{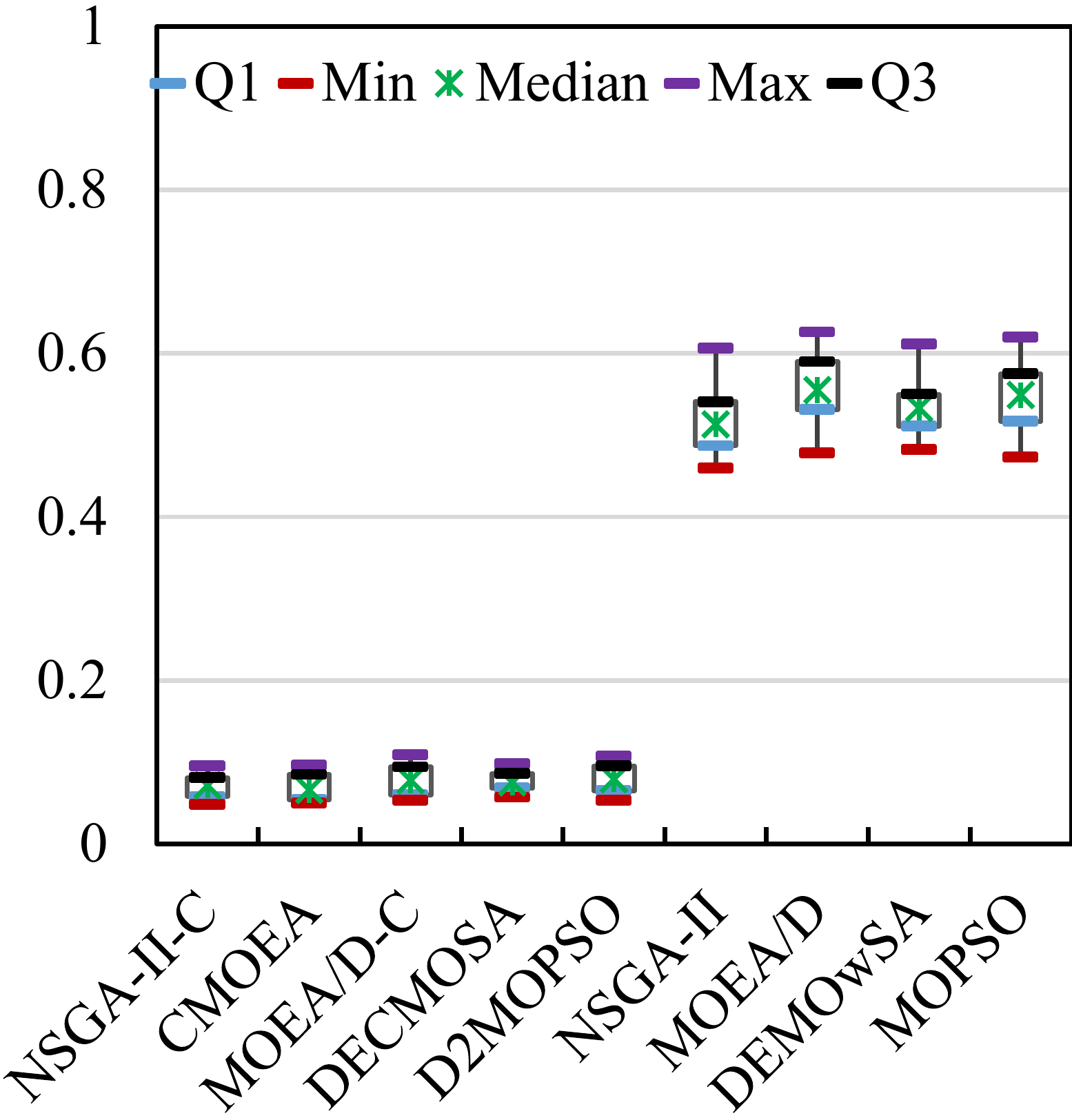}}
\hfil
\subfigure[H1, 2$^\textnormal{nd}$ half Mar]{\includegraphics[scale=0.281]{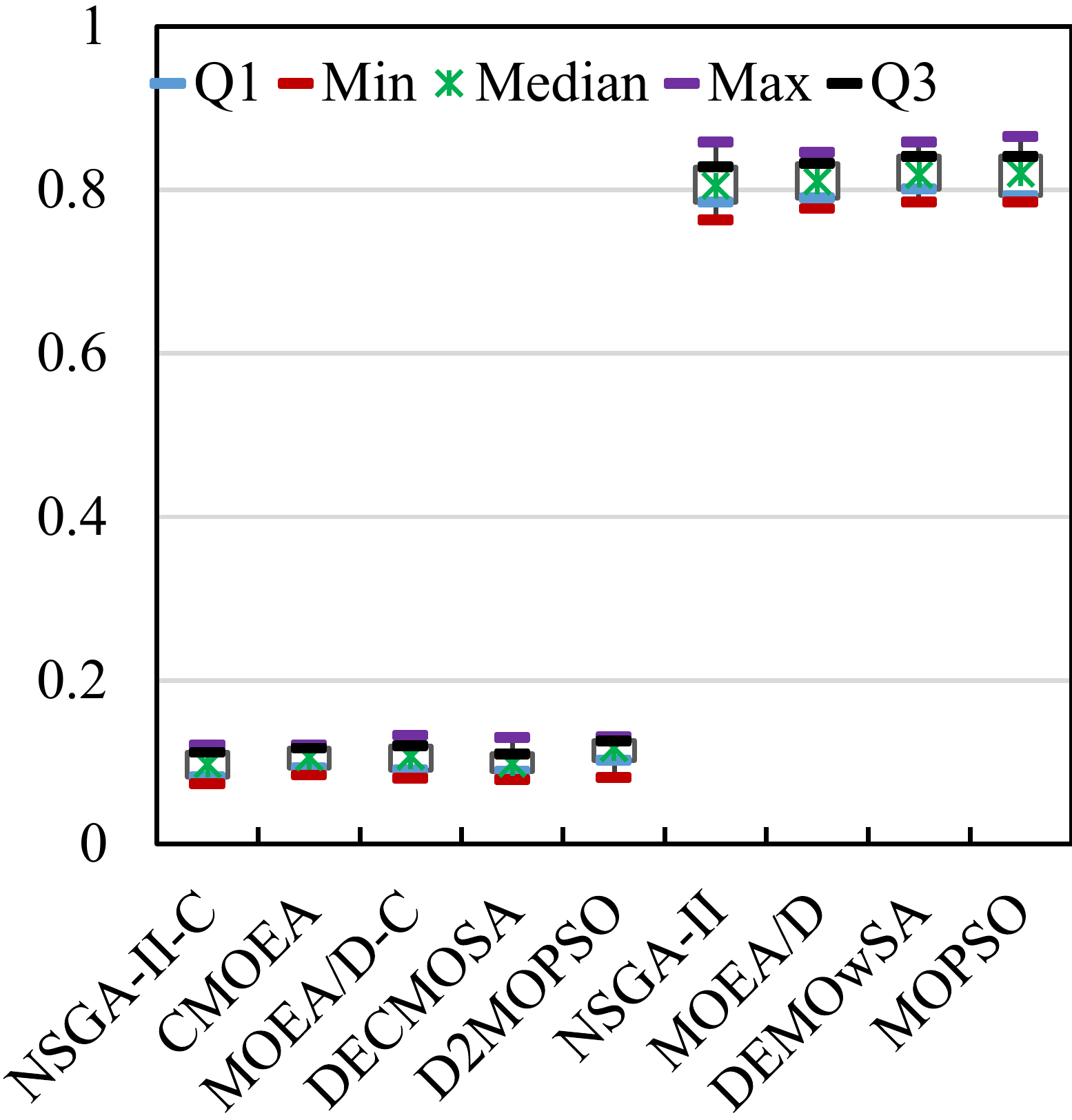}}
\subfigure[H1, 1$^\textnormal{st}$ half Apr]{\includegraphics[scale=0.281]{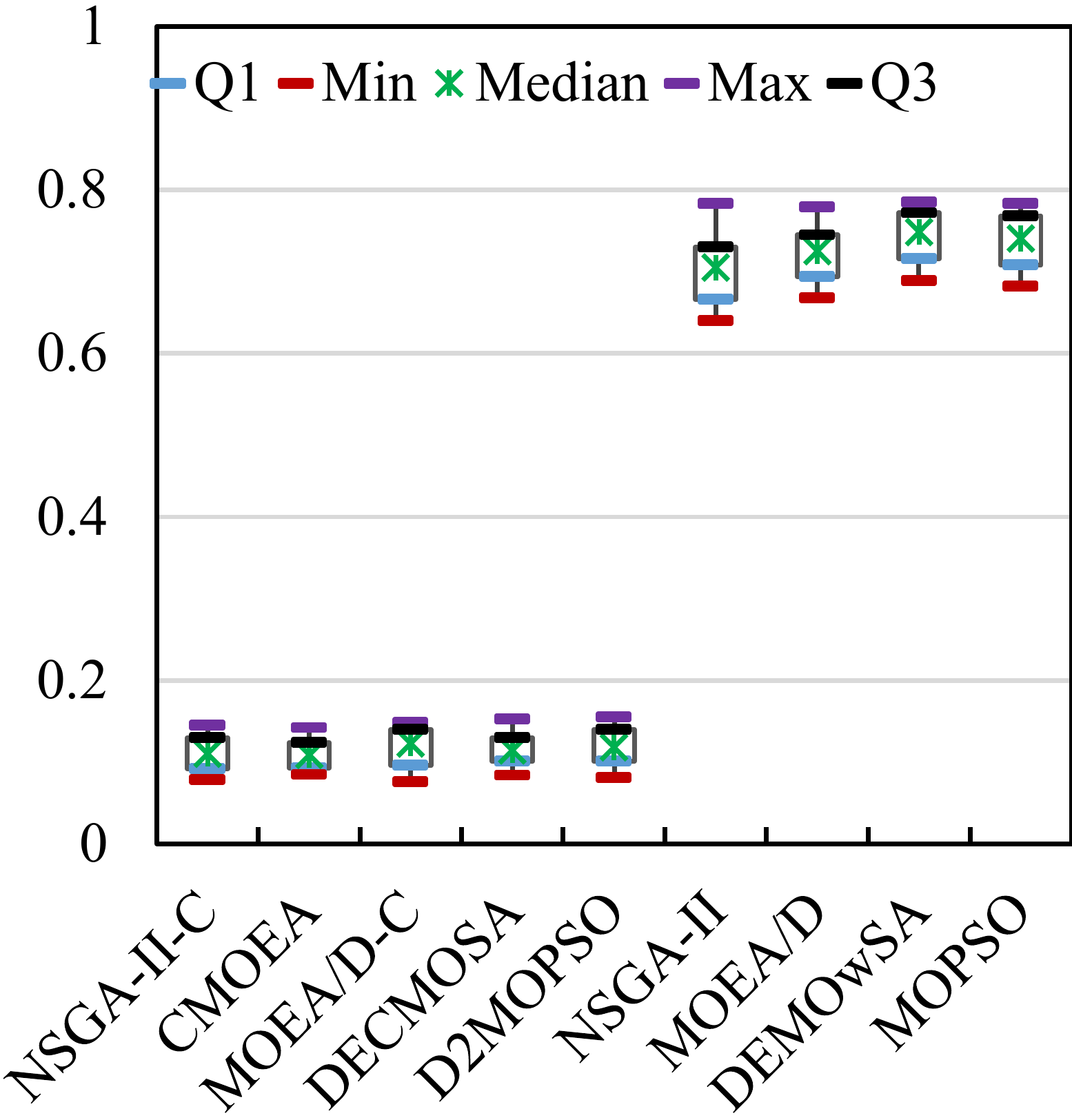}}
\subfigure[H2, 2$^\textnormal{nd}$ half Mar]{\includegraphics[scale=0.281]{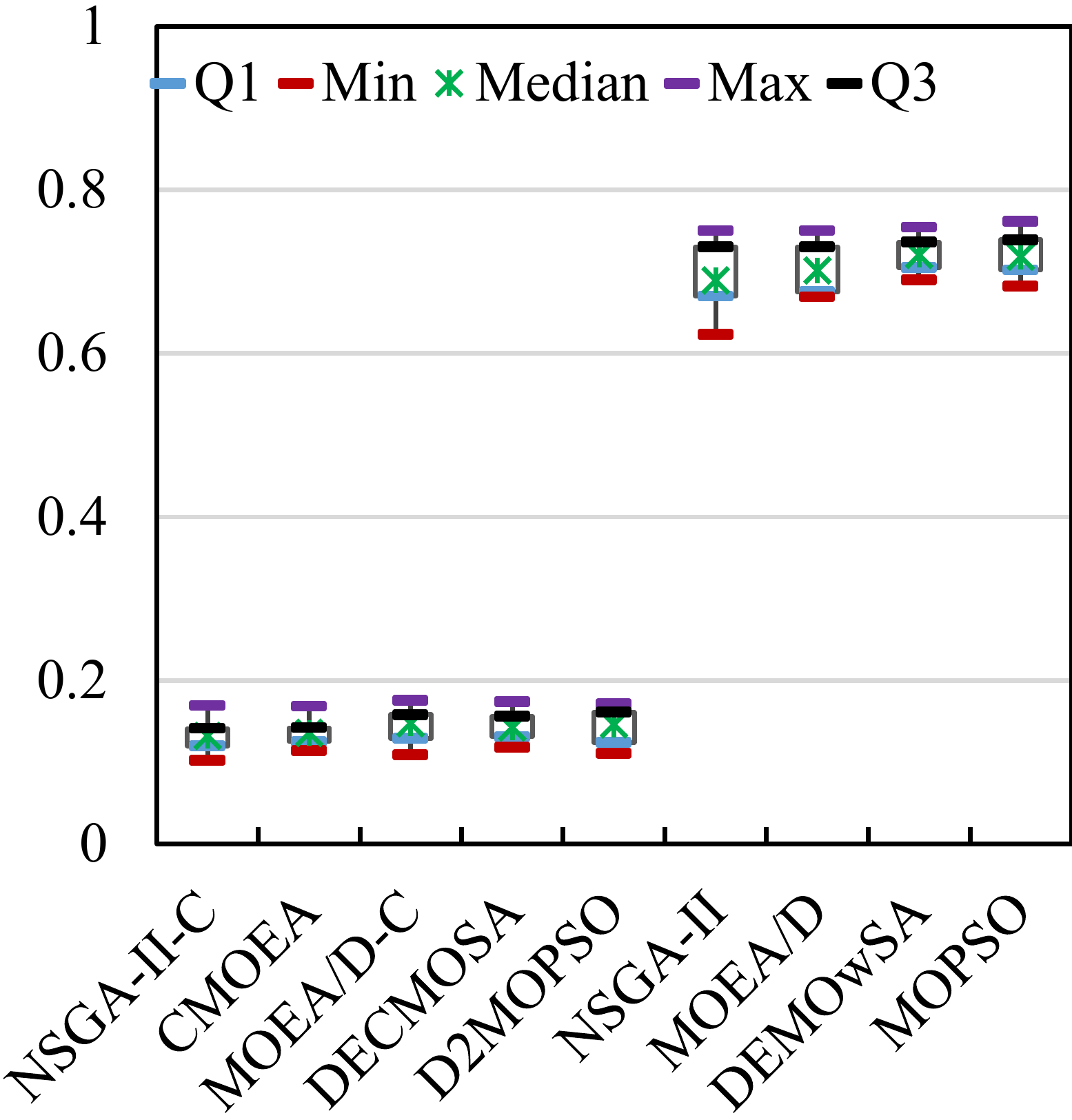}}
\subfigure[H2, 1$^\textnormal{st}$ half Apr]{\includegraphics[scale=0.281]{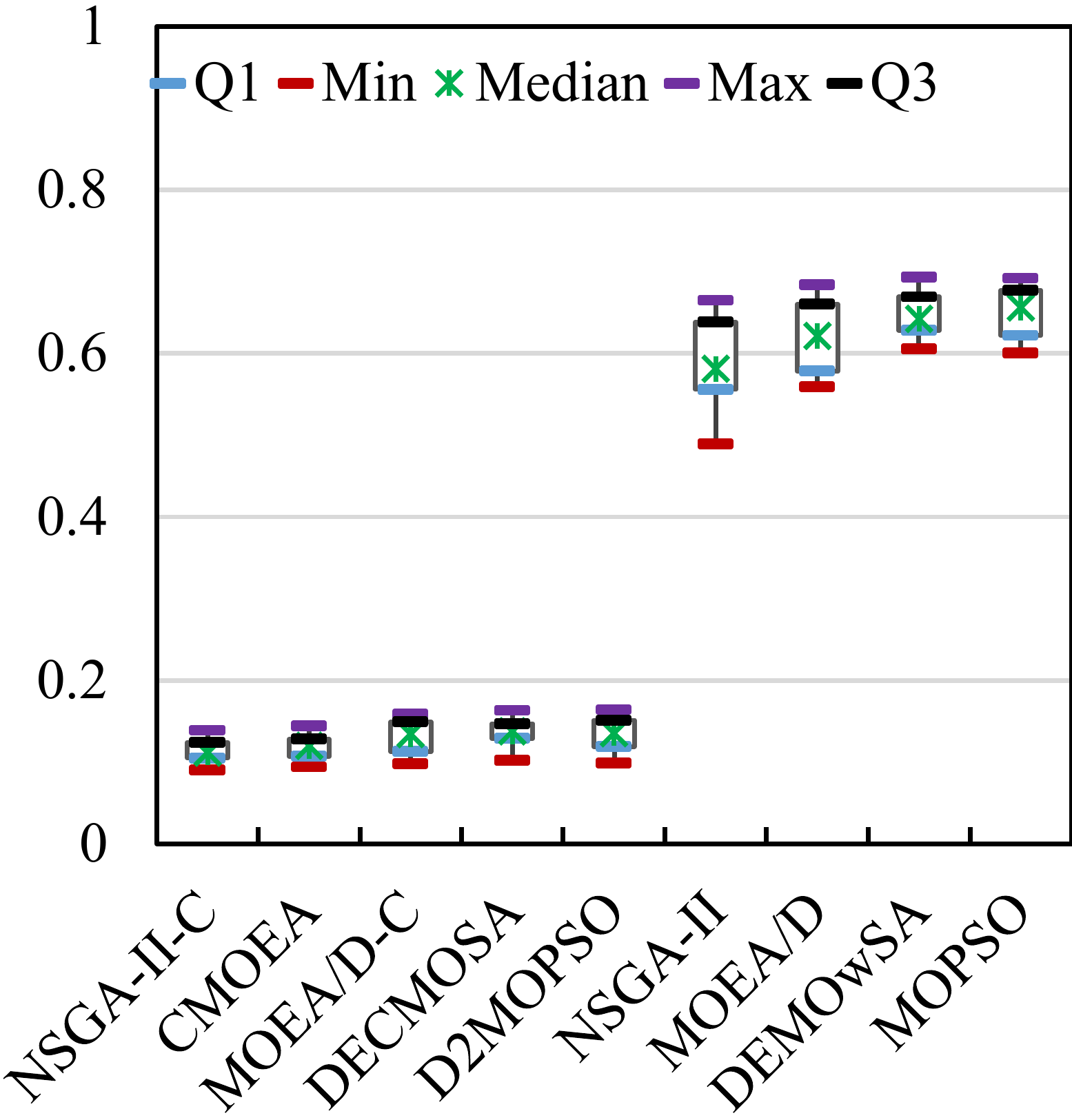}}
\hfil
\subfigure[H3, 2$^\textnormal{nd}$ half Mar]{\includegraphics[scale=0.281]{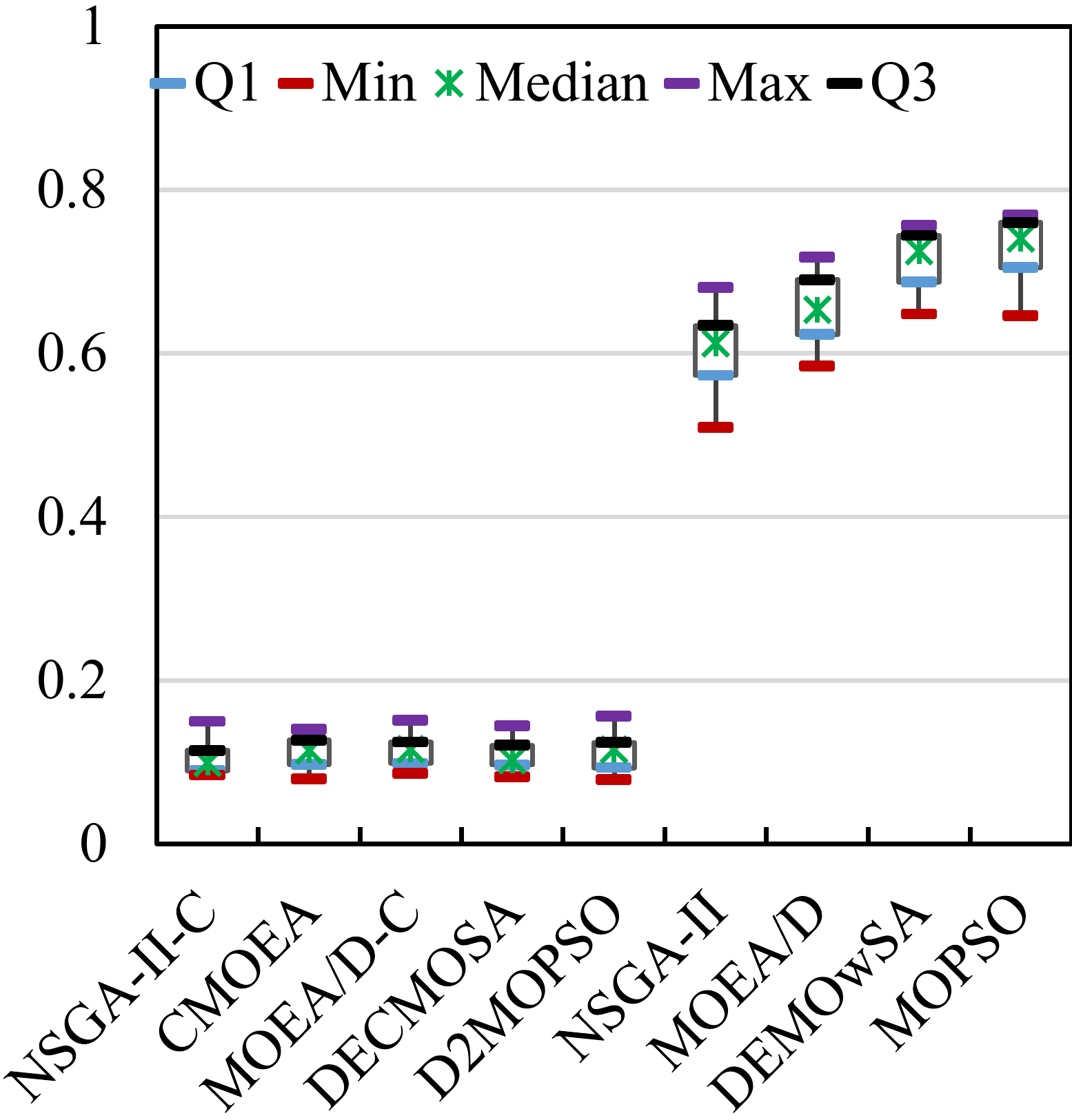}}
\subfigure[H3, 1$^\textnormal{st}$ half Apr]{\includegraphics[scale=0.281]{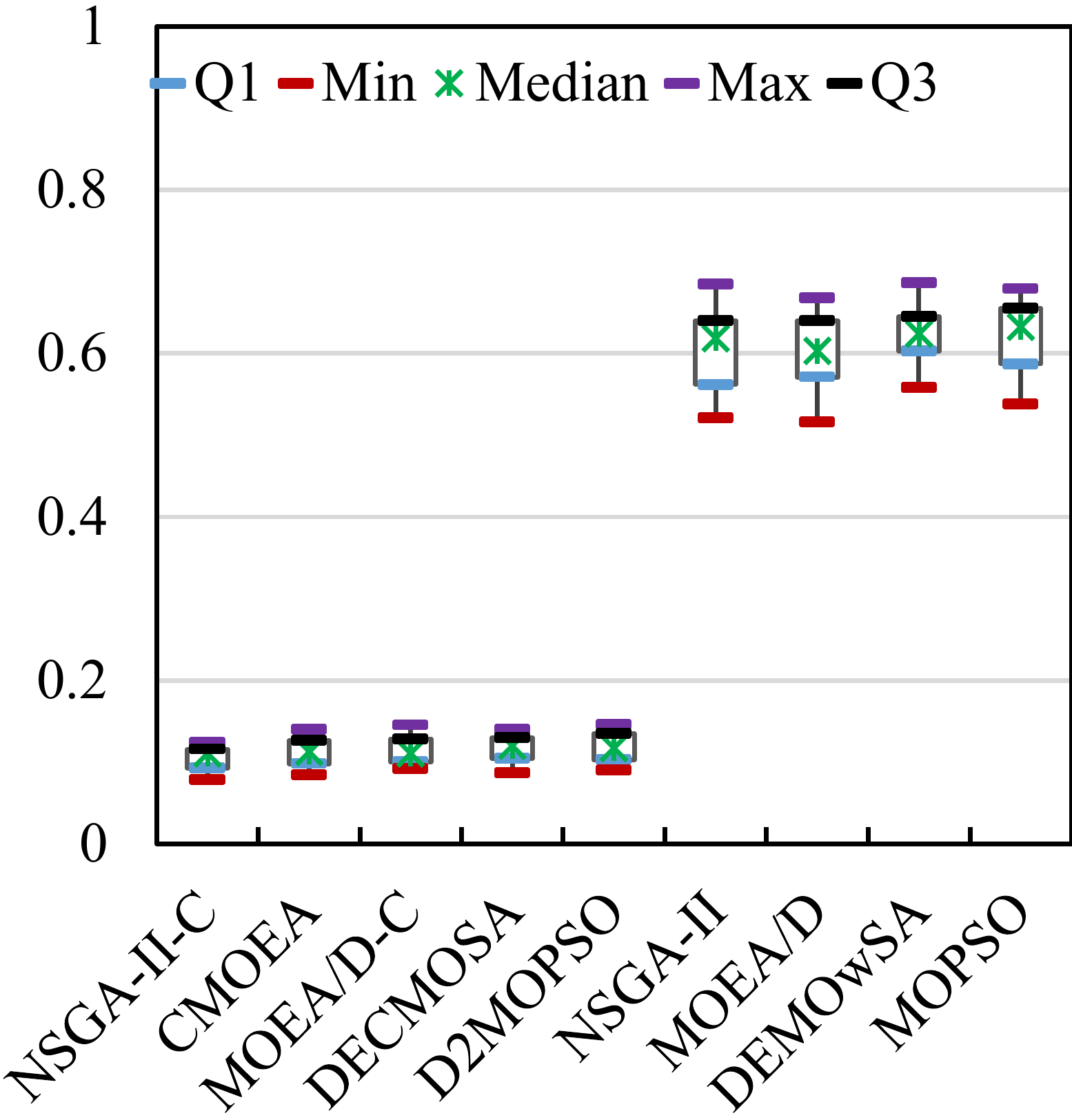}}
\subfigure[H4, 2$^\textnormal{nd}$ half Mar]{\includegraphics[scale=0.281]{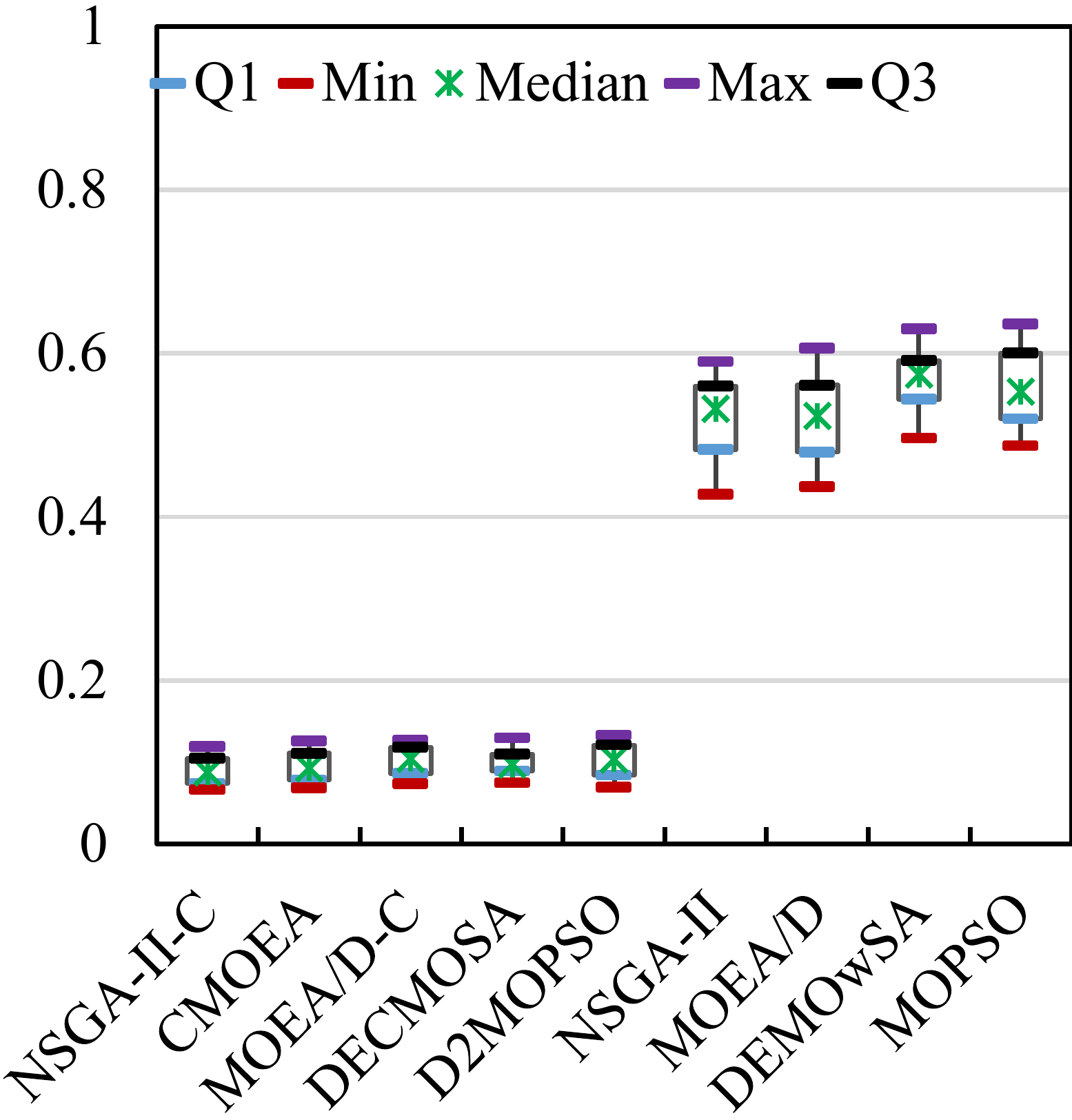}}
\subfigure[H4, 1$^\textnormal{st}$ half Apr]{\includegraphics[scale=0.281]{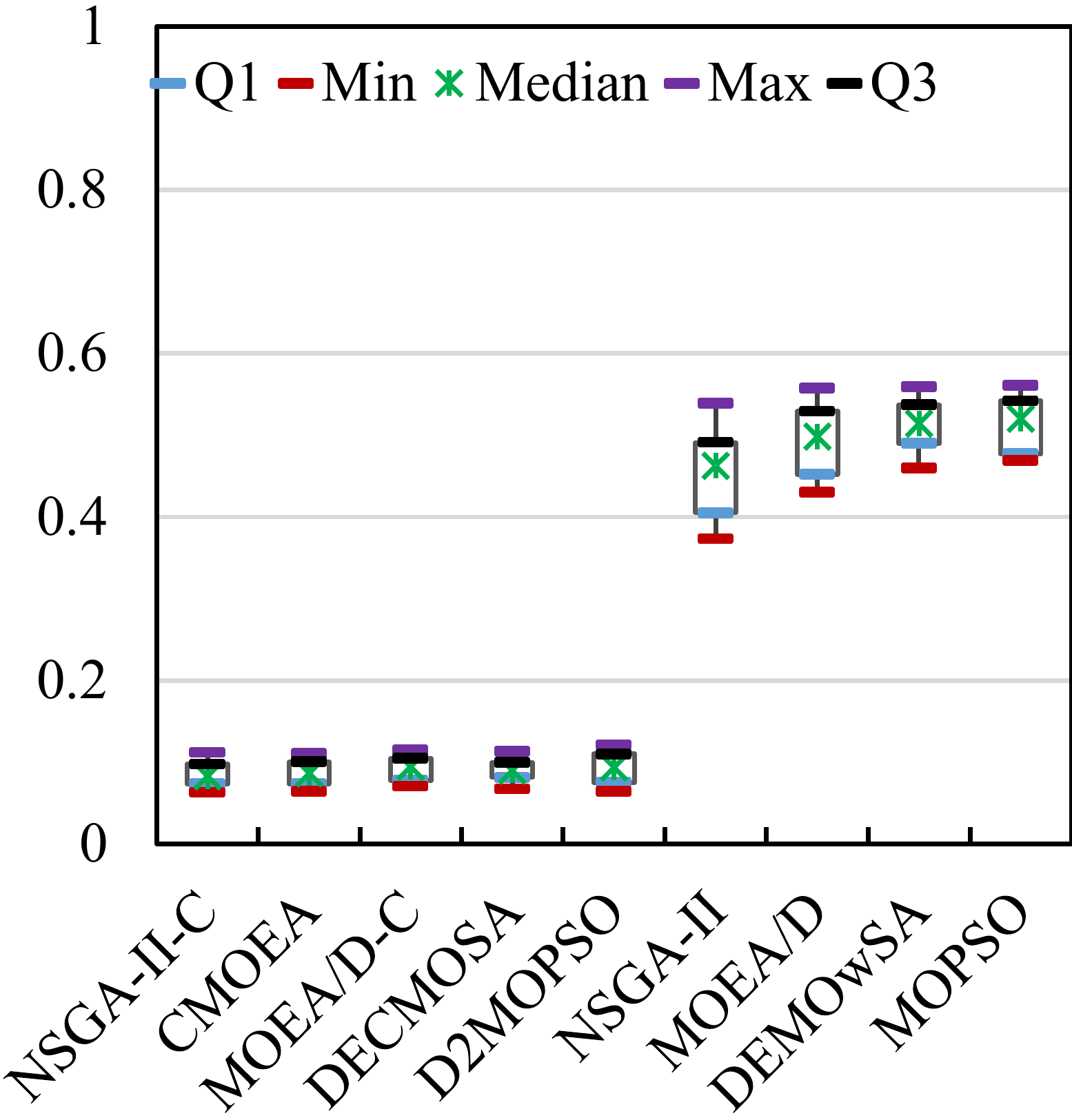}}
\hfil
\subfigure[H5, 2$^\textnormal{nd}$ half Mar]{\includegraphics[scale=0.281]{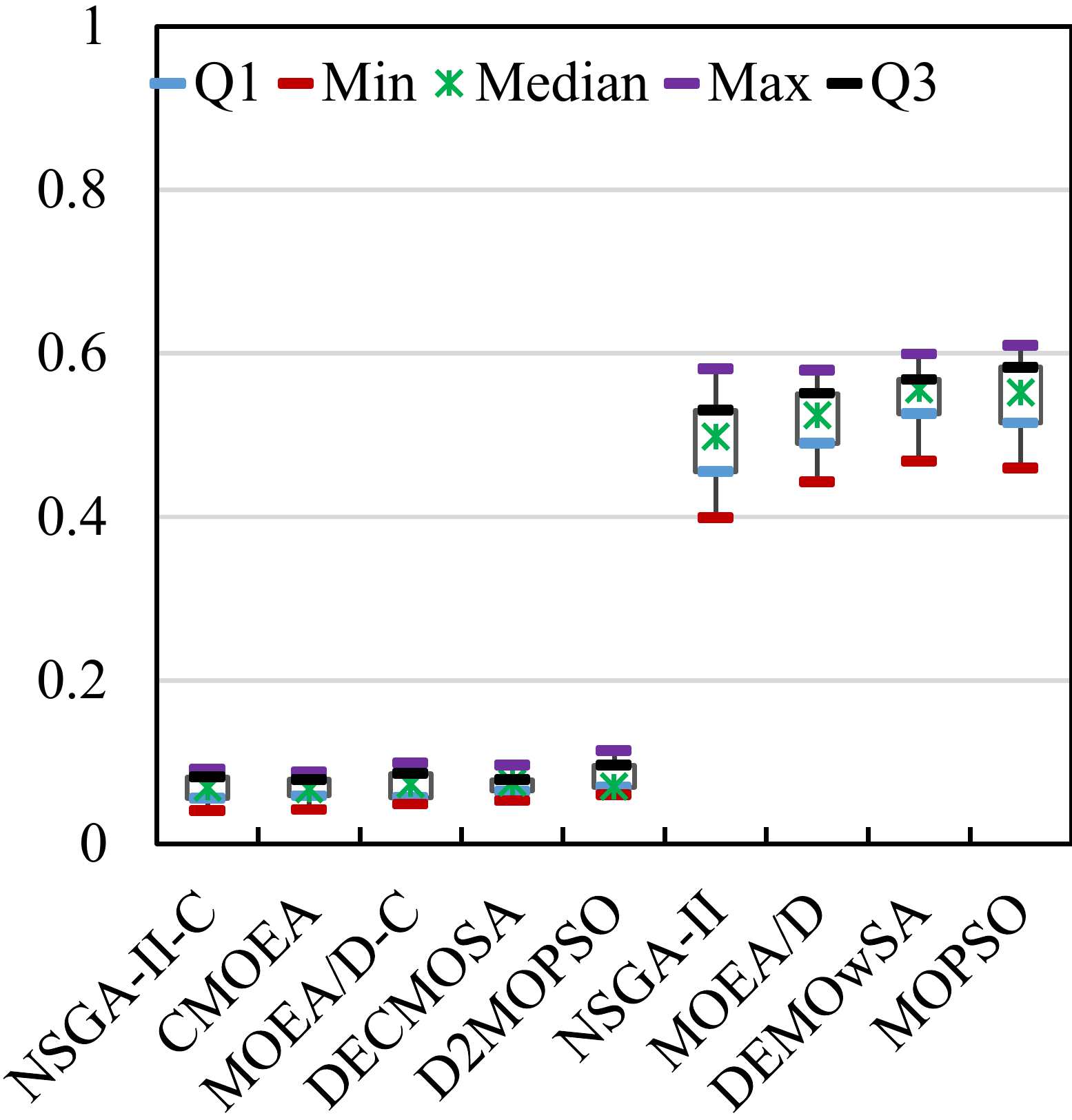}}
\subfigure[H5, 1$^\textnormal{st}$ half Apr]{\includegraphics[scale=0.281]{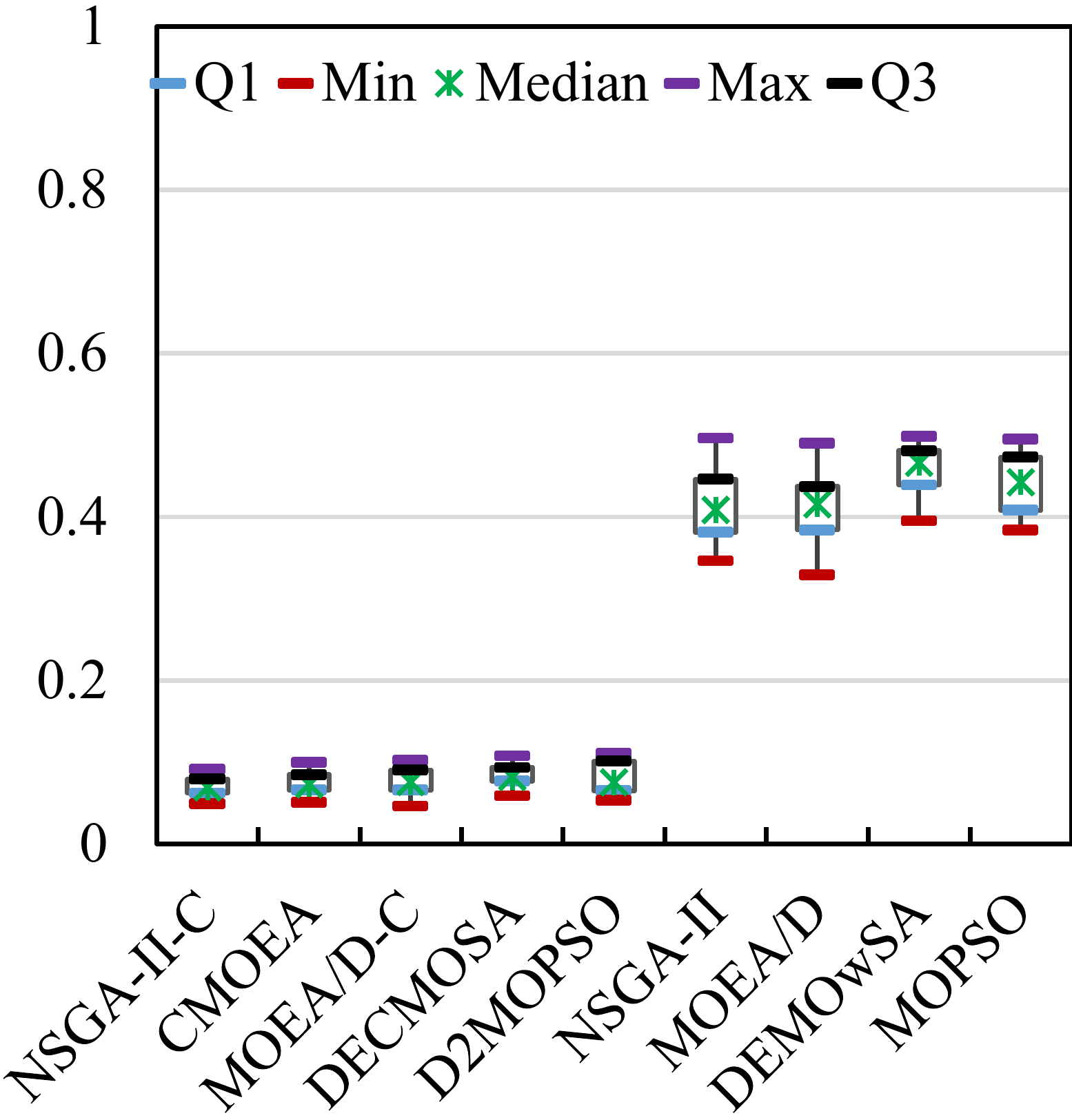}}
\hfil
\caption{Comparison of hyperareas obtained by the algorithms on main problem instances. Each box plot shows the maximum, minimum, median, first quartile (Q1), and third quartile (Q3) of hyperareas over the 30 runs of an algorithm.}
\label{fig:res}\end{figure*}

Fig. \ref{fig:res} compares the hyperarea (the area under the Pareto-approximated front in objective space, also known as the hypervolume) \cite{Zitzler99TEC,Coello2002MOEA} obtained by each algorithm on each main problem instance. It is clear that the last four algorithms using transform-and-divide exhibit significant performance advantages over the first five algorithms. On large-size instances of ZJHTCM and H5, the median hyperareas of the four transform-and-divide EAs are approximately eight to ten times of those of the basic EAs; on the other instances, the median hyperareas of the transform-and-divide EAs are approximately six to eight times of those of the basic EAs. In general, the performance of the transform-and-divide EAs is mainly affected by $m$ (the number of diseases) and $C$ (the total budget), while that of the basic EAs is mainly affected by $n\!+\!n'$ (the number of supplies). This is why all algorithms obtain relatively high hyperareas on the instances of H1 and H2, where $m$, $n\!+\!n'$, and $C$ are relatively small. Nevertheless, the performance advantages of the transform-and-divide EAs over the basic EAs are very significant on these relatively small-size instances. On the last two instances of H5, the values of these parameters are large, and the performance advantages of the transform-and-divide EAs over the basic EAs are not so significant. On each instance, the minimum hyperareas of the transform-and-divide EAs are still significantly larger (about four to five times) than the maximum hyperareas of the basic EAs. The results demonstrate that the proposed transform-and-divide method can greatly reduce the difficulty of solving the complex original problem.

\begin{figure*}[!t]
\centering
\subfigure[ZJHTCM, 1$^\textnormal{st}$ half Mar]{\includegraphics[scale=0.246]{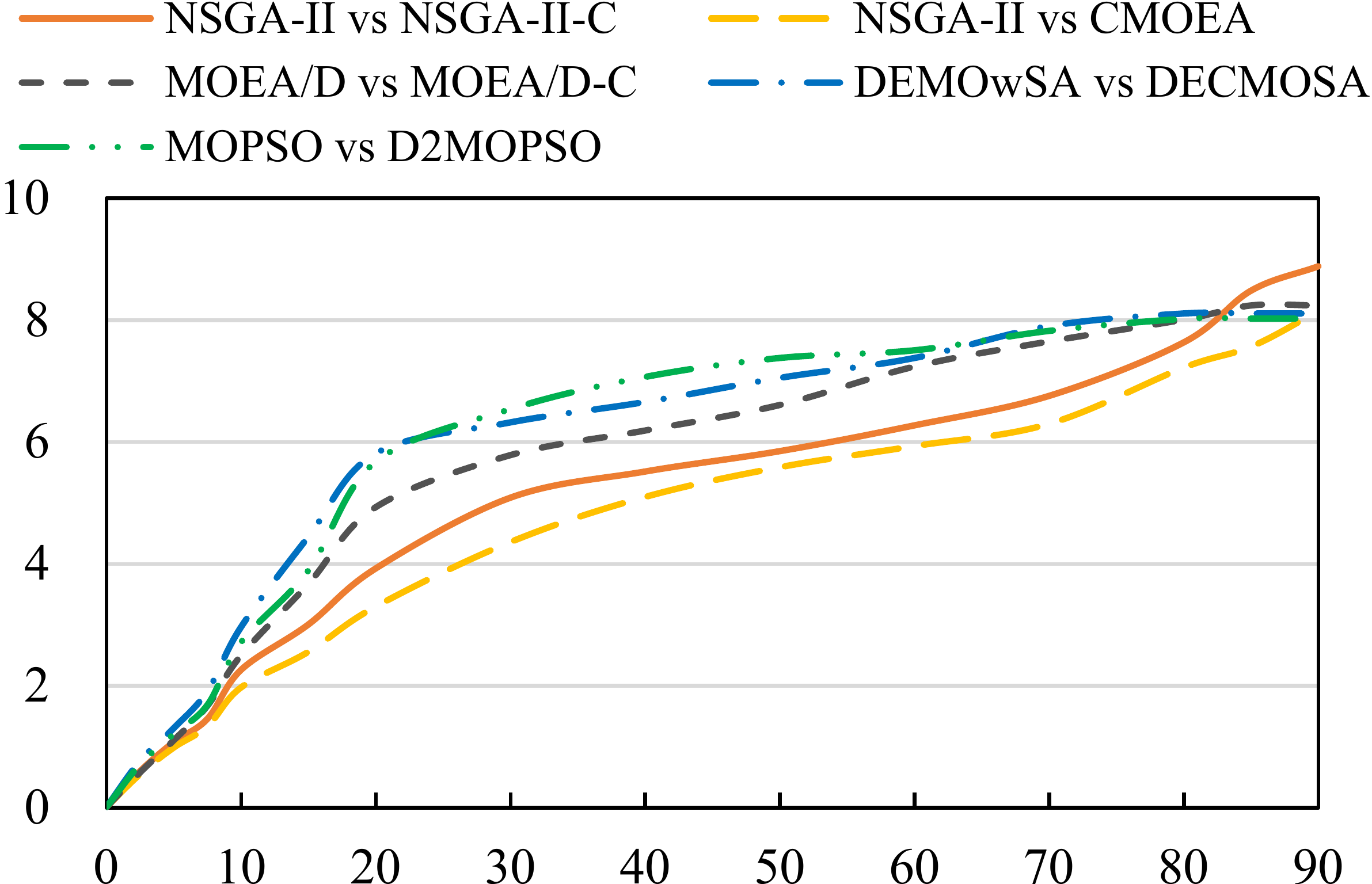}}
\subfigure[ZJHTCM, 2$^\textnormal{nd}$ half Mar]{\includegraphics[scale=0.246]{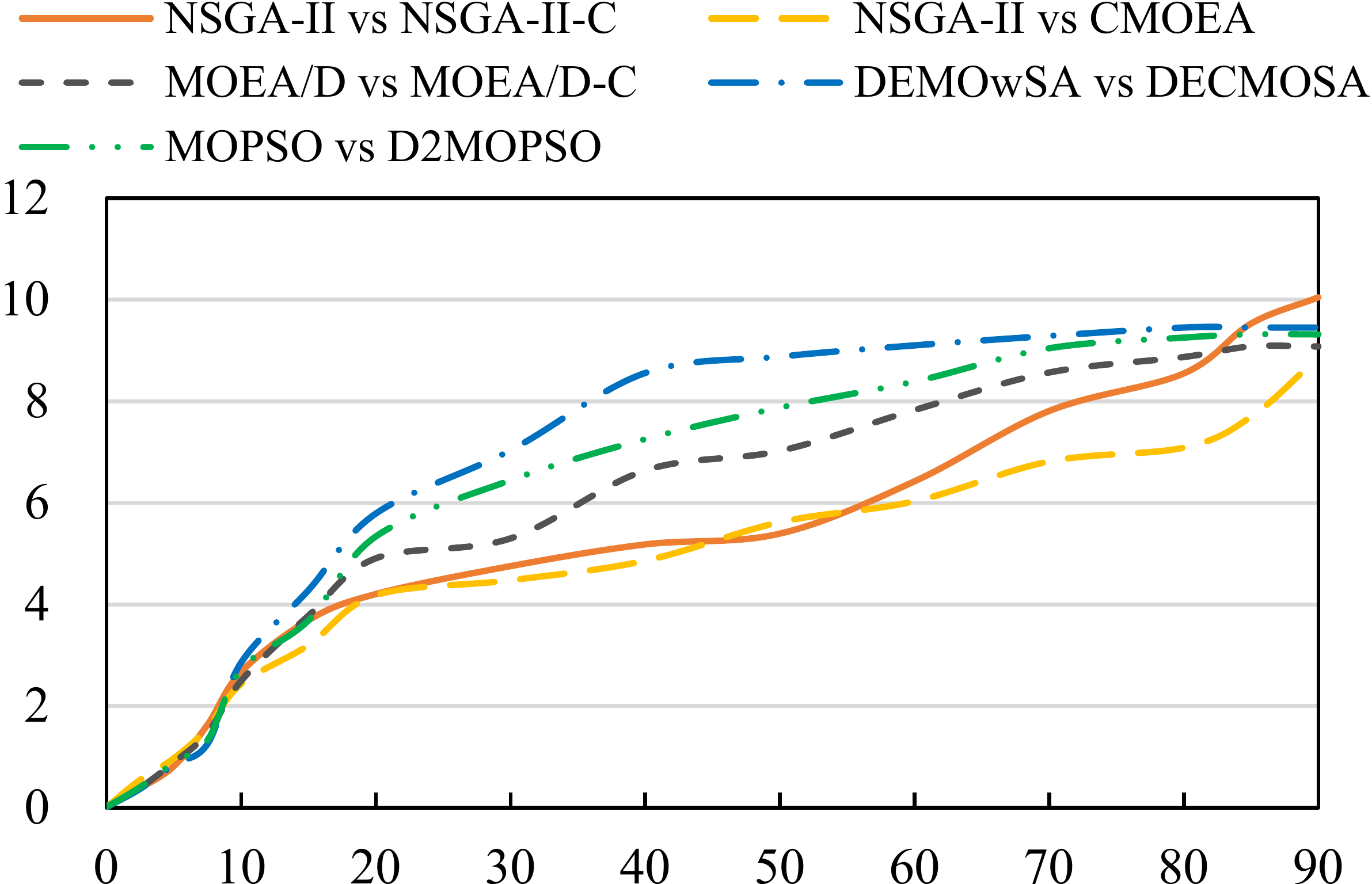}}
\subfigure[ZJHTCM, 1$^\textnormal{st}$ half Apr]{\includegraphics[scale=0.246]{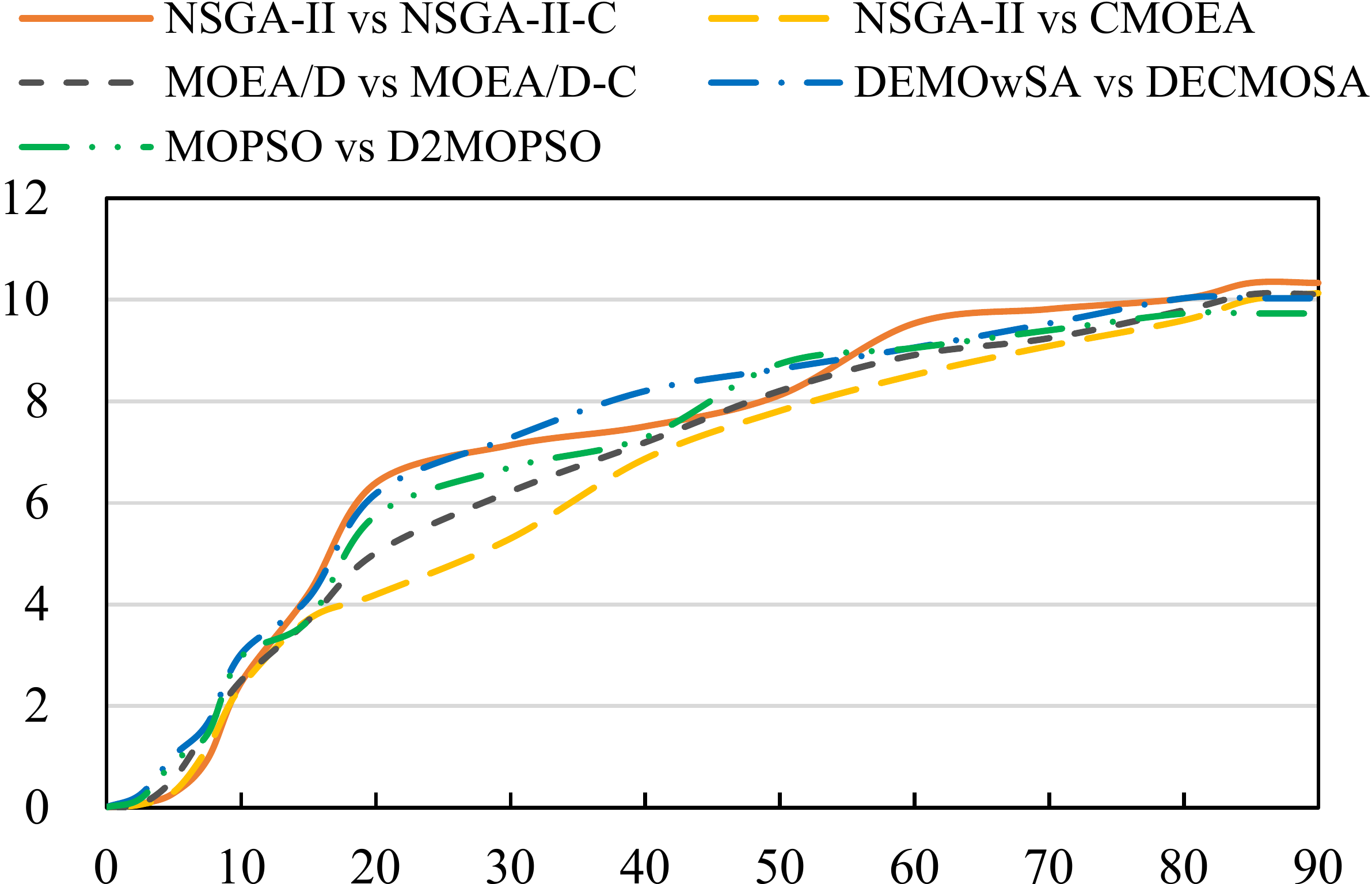}}
\hfil
\subfigure[ZJHTCM, 1$^\textnormal{st}$ half Apr]{\includegraphics[scale=0.246]{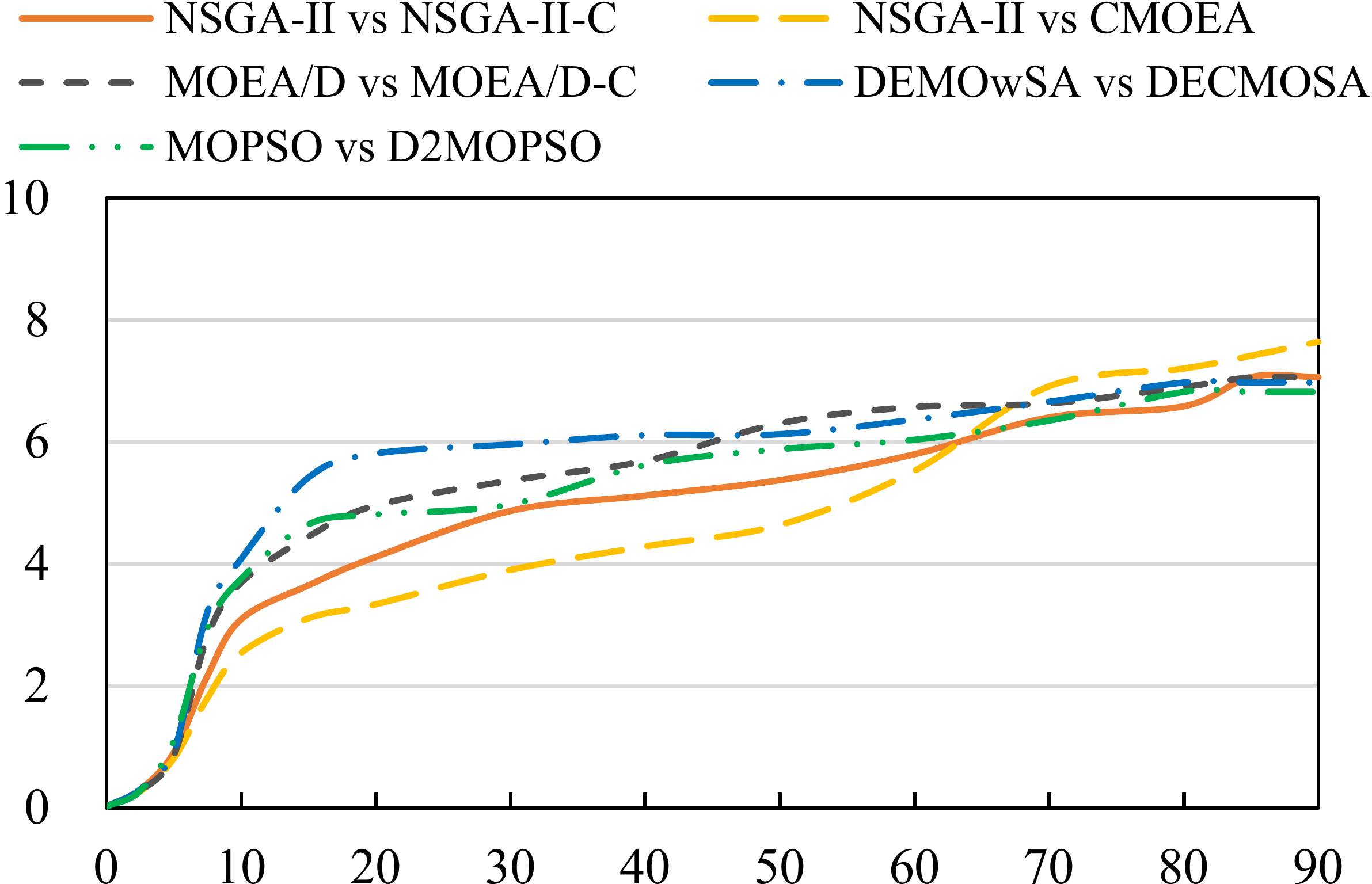}}
\subfigure[H1, 2$^\textnormal{nd}$ half Mar]{\includegraphics[scale=0.246]{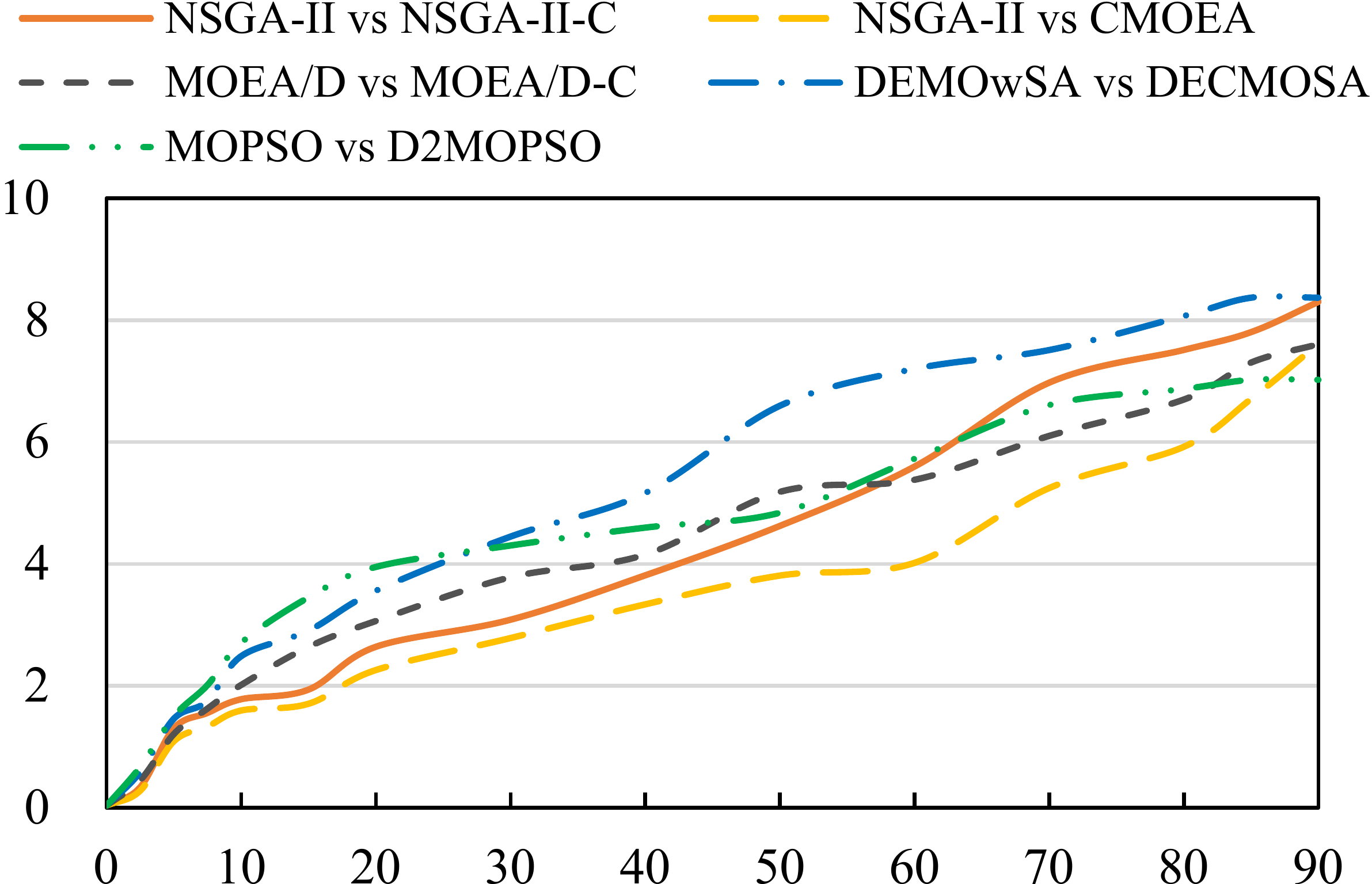}}
\subfigure[H1, 1$^\textnormal{st}$ half Apr]{\includegraphics[scale=0.246]{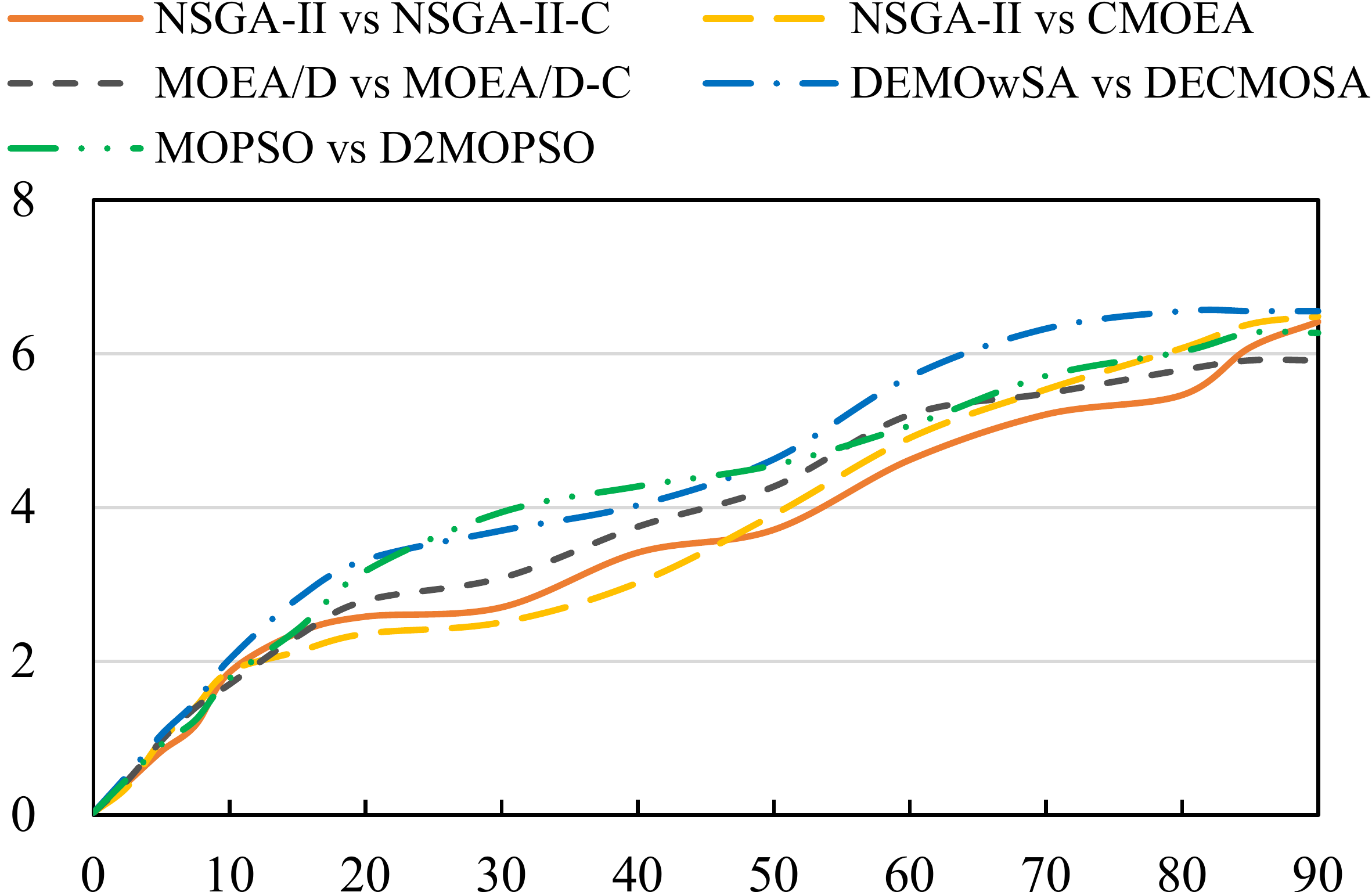}}
\hfil
\subfigure[H2, 2$^\textnormal{nd}$ half Mar]{\includegraphics[scale=0.246]{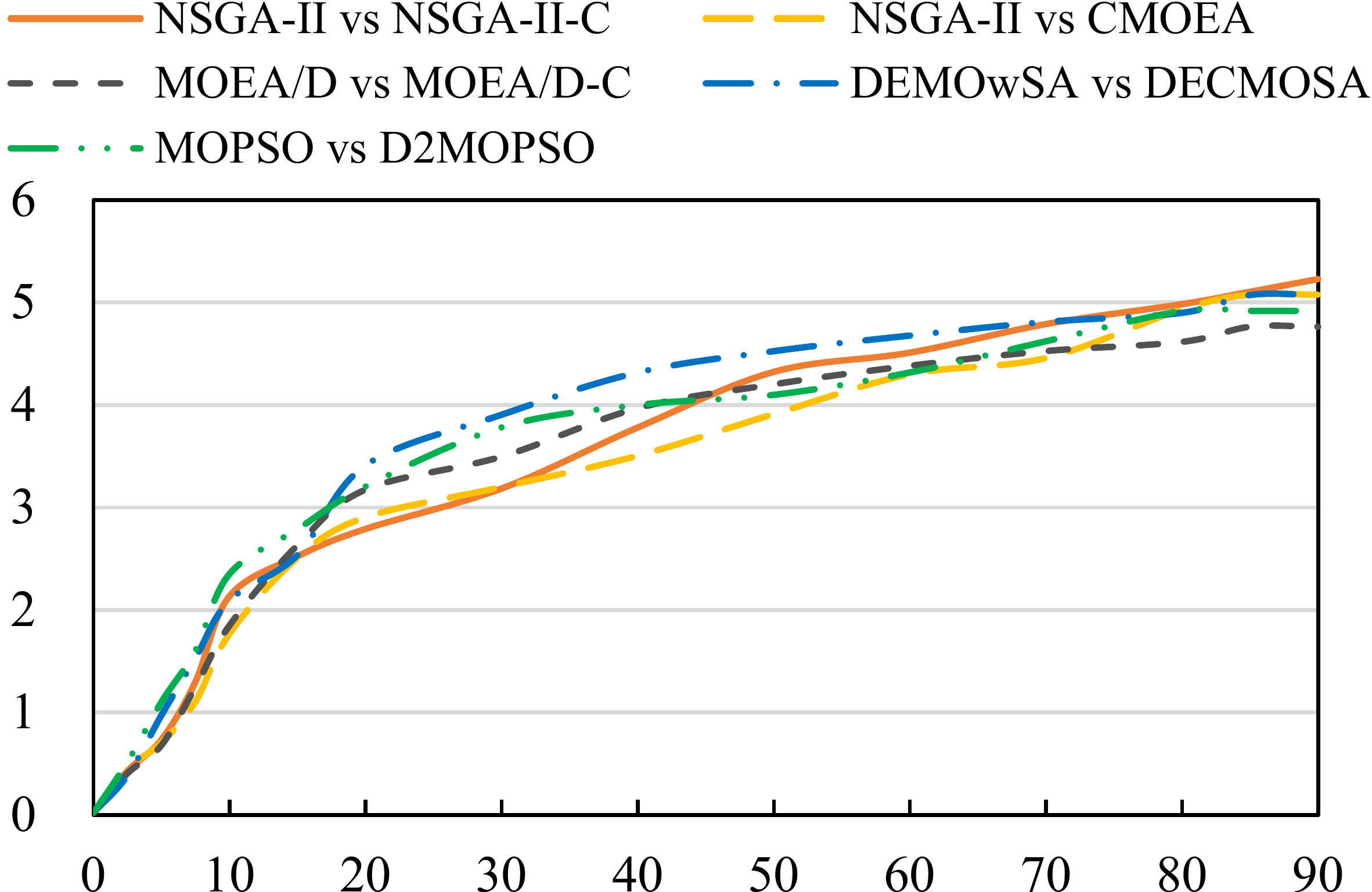}}
\subfigure[H2, 1$^\textnormal{st}$ half Apr]{\includegraphics[scale=0.246]{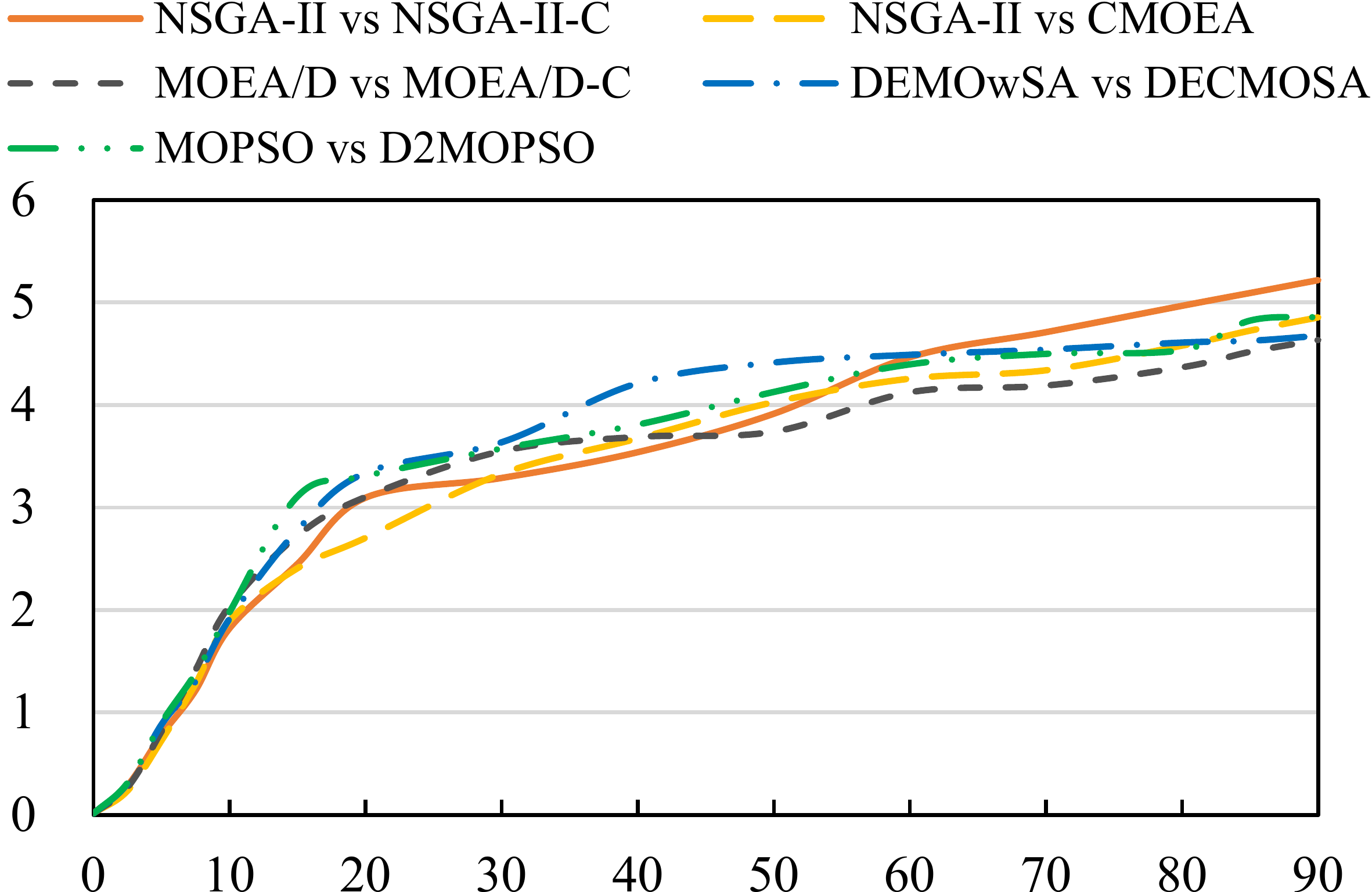}}
\subfigure[H3, 2$^\textnormal{nd}$ half Mar]{\includegraphics[scale=0.246]{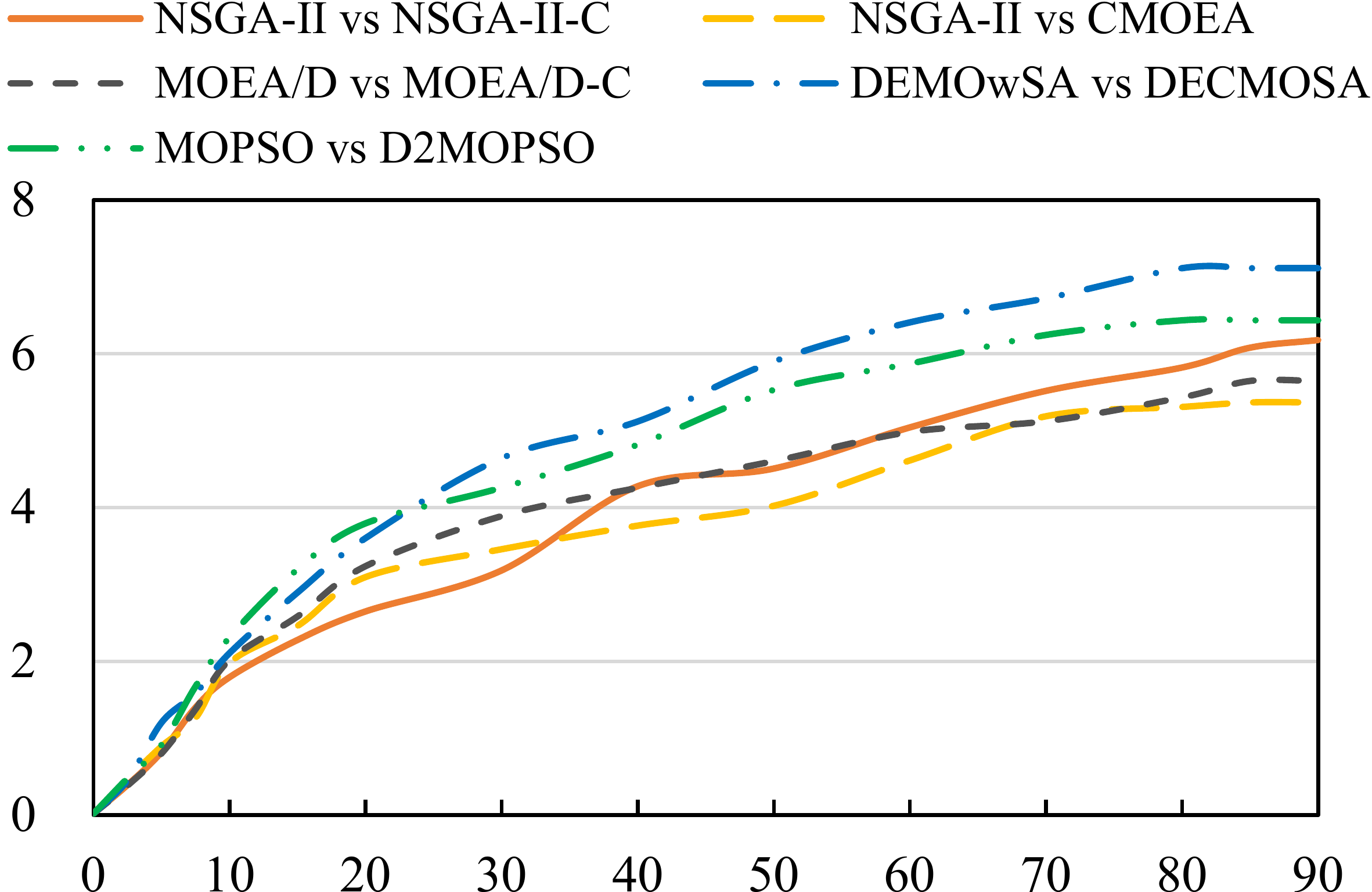}}
\hfil
\subfigure[H3, 1$^\textnormal{st}$ half Apr]{\includegraphics[scale=0.246]{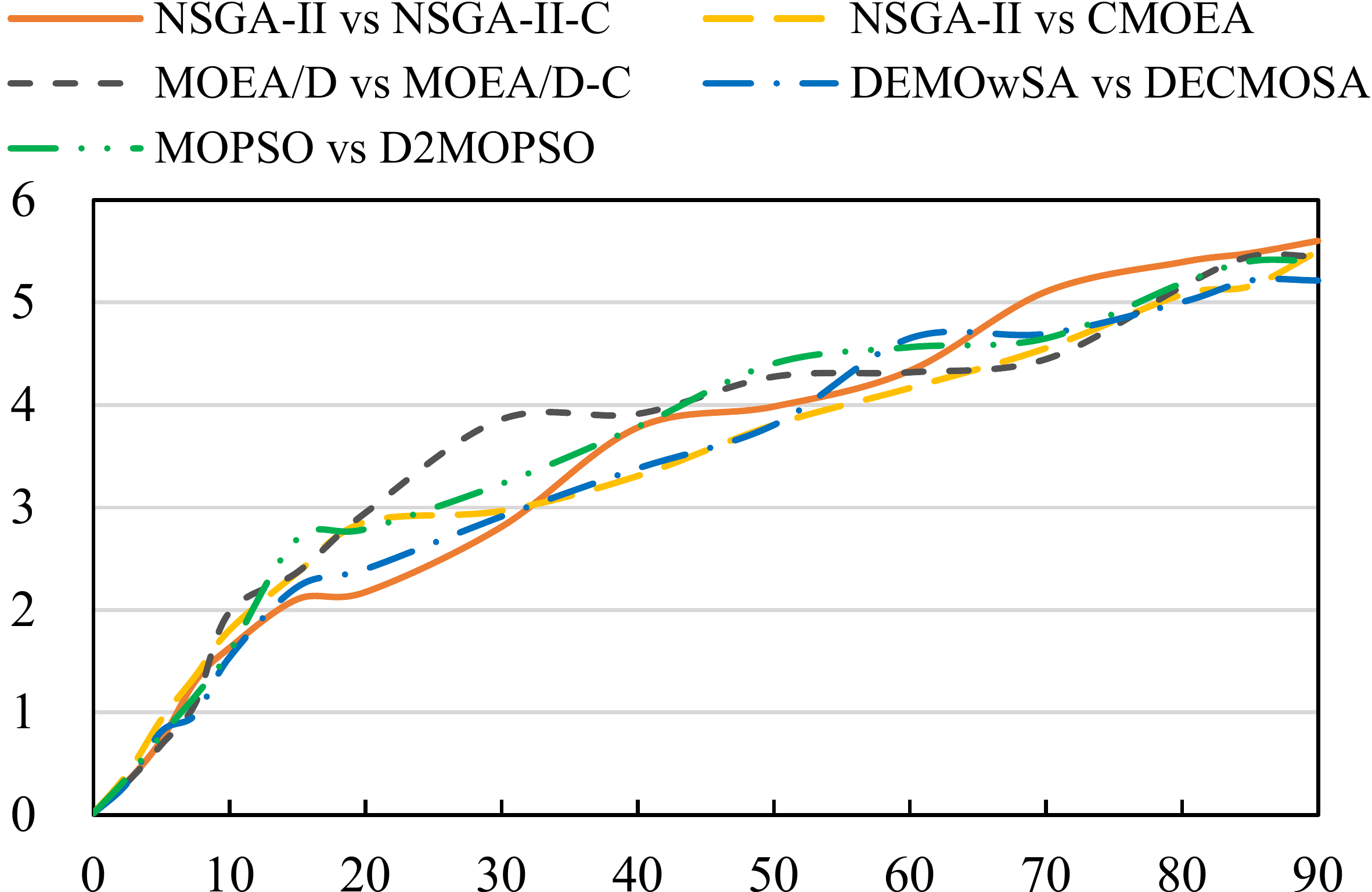}}
\subfigure[H4, 2$^\textnormal{nd}$ half Mar]{\includegraphics[scale=0.246]{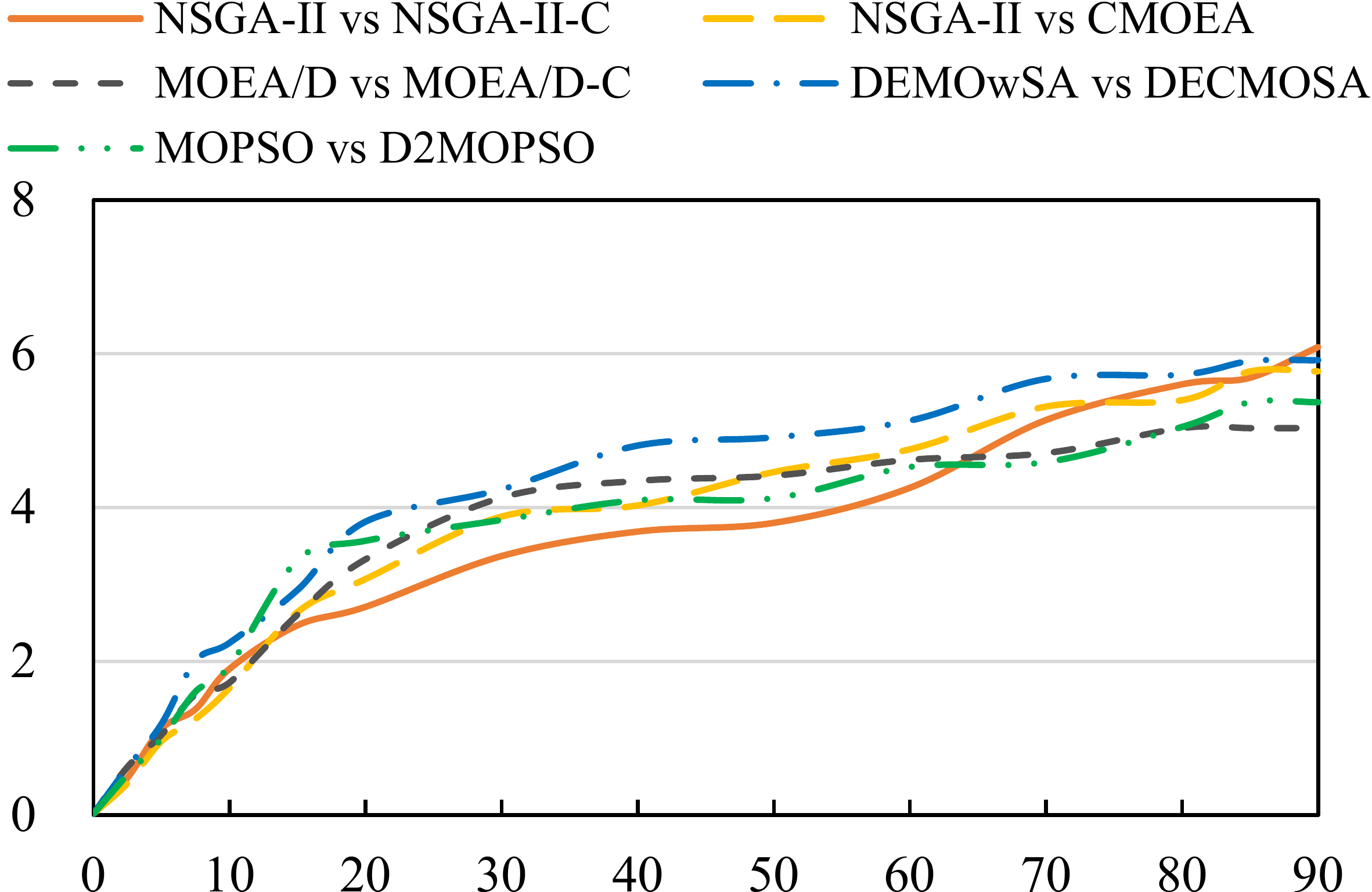}}
\subfigure[H4, 1$^\textnormal{st}$ half Apr]{\includegraphics[scale=0.246]{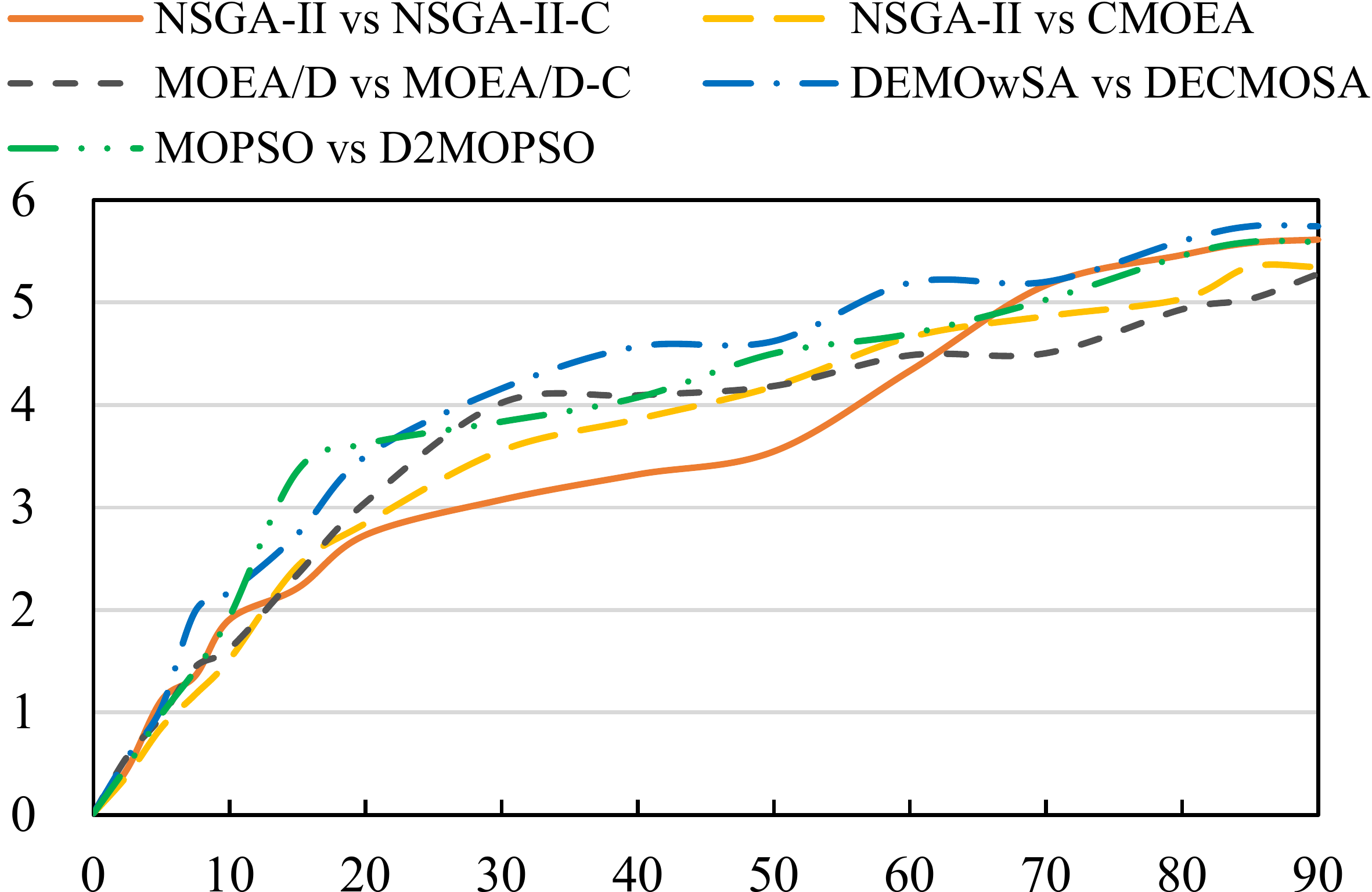}}
\hfil
\subfigure[H5, 2$^\textnormal{nd}$ half Mar]{\includegraphics[scale=0.246]{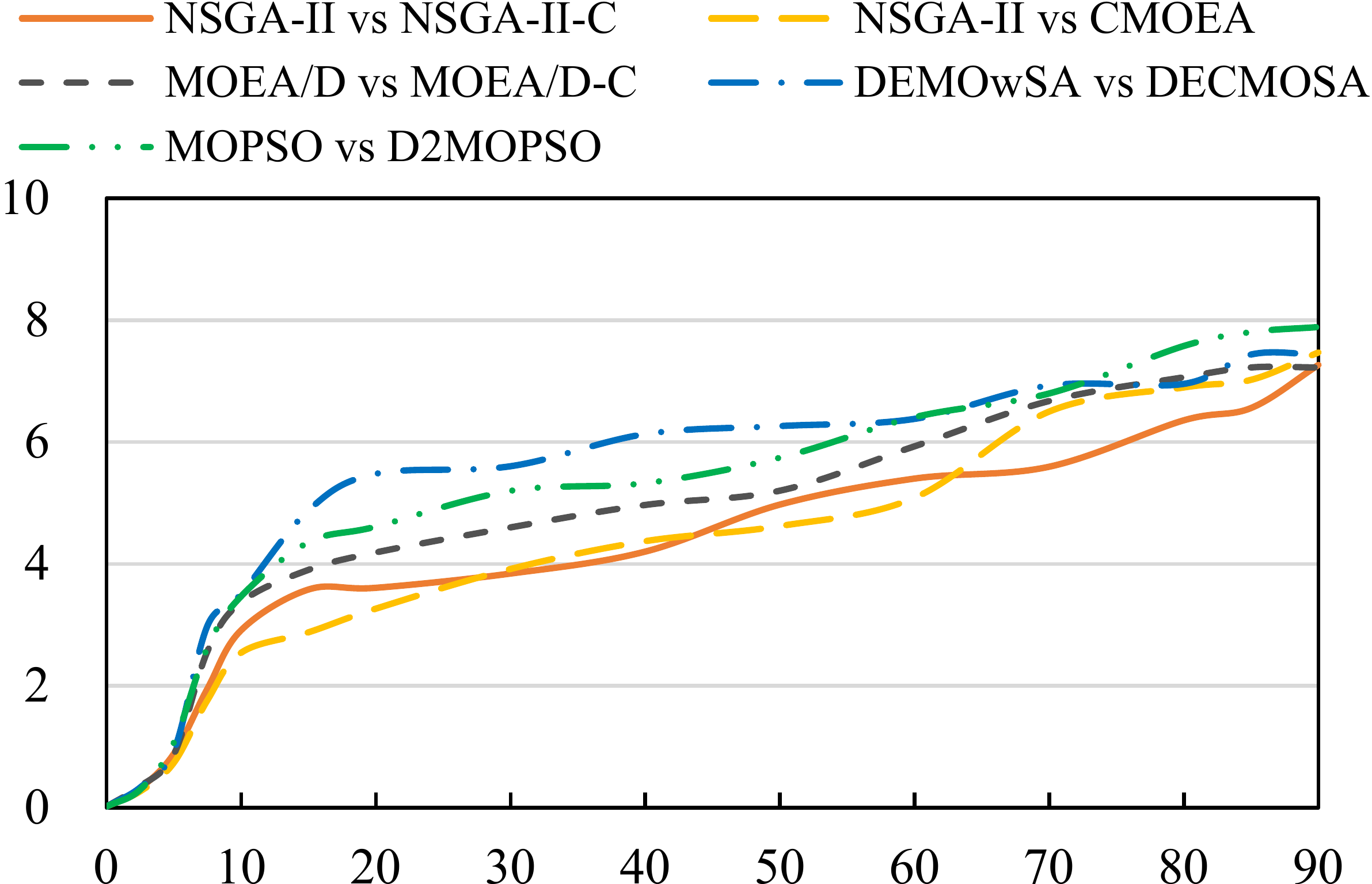}}
\subfigure[H5, 1$^\textnormal{st}$ half Apr]{\includegraphics[scale=0.246]{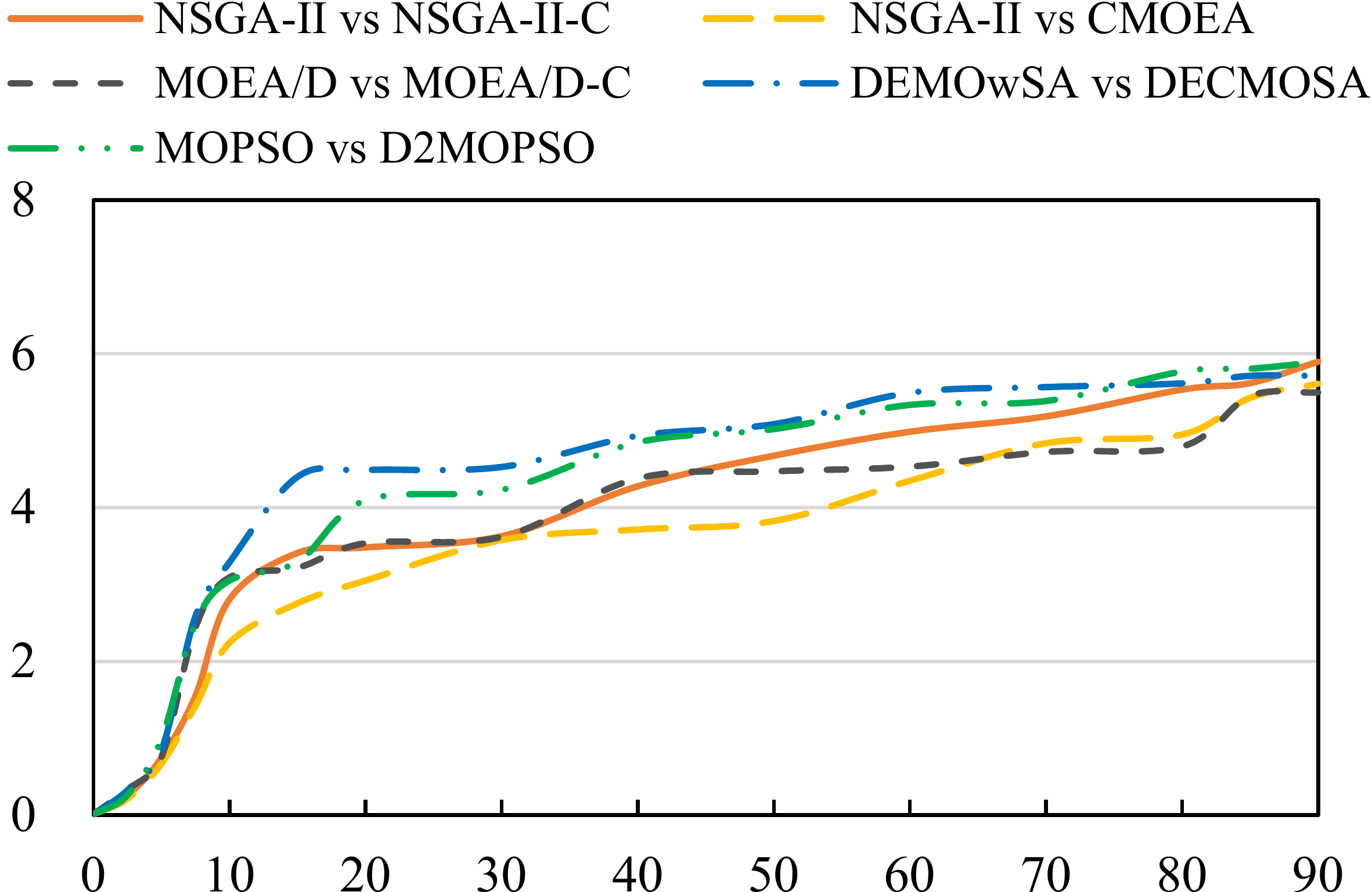}}
\caption{Pairwise comparison between each unconstrained multiobjective EA used in transform-and-divide and its constrained counterpart for the original problem. The vertical axis is the ratio of the hyperarea obtained by the first algorithm to the maximum hyperarea obtained by the second algorithm, and the horizontal axis is the CPU time (in minutes).}
\label{fig:cmp}\end{figure*}

We also make a pairwise comparison between each transform-and-divide EA and its constrained version for the original problem, e.g., NSGA-II vs. NSGA-II-C, NSGA-II vs. CMOEA, MOEA/D vs. MOEA/D-C, DEMOwSA vs DECMOSA, and MOPSO vs D$^2$MOPSO. That is, in each pair, the first is an unconstrained multiobjective EA used in transform-and-divide, and the second is a multiobjective EA adding constraint handling mechanisms (such as constrained-domination or penalty functions) to the first one. The comparison is made based on the ratio of the hyperarea obtained by the first algorithm to the maximum hyperarea obtained by the second algorithm. Fig. \ref{fig:cmp} presents the changes of hyperarea ratios over the CPU time on each problem instance. In general, it takes about five minutes for a transform-and-divide EA to reach the maximum hyperarea obtained by its counterpart after 90 minutes; on large-size instances of ZJHTCM and H5, the ratio is between 3 and 4 at ten minutes and exceeds 5 before twenty minutes; on the other instances, the ratio is approximately 2 at ten minutes and between 3 and 4 at twenty minutes. The high hyperarea ratios demonstrate that the transform-and-divide EAs approximate the Parato front of the problem significantly more efficiently than the basic EAs, mainly because the transform-and-divide strategy significantly reduces the solution space, and also because the unconstrained EAs for the transformed problem can explore the solution space more effectively than their constrained versions. Comparatively, the ratios of NSGA-II vs. its counterparts (NSGA-II-C using constrained-domination and CMOEA using a penalty) grow more slowly than those of DEMOwSA and MOPSO vs. their counterparts in early stages of evolution. As there is no significant difference among the five basic EAs for the original problem, the results indicate that DEMOwSA and MOPSO are more efficient in exploring the solution space to approximate the Parato front of the transformed problem.

Besides the hyperarea metric, we also compare the algorithms in terms of the coverage ($\textit{Cov}$) metric \cite{Zitzler99TEC}, i.e., $\textit{Cov}(X,X')$ is the fraction of solution set $X$ obtained by an algorithm that are strictly dominated by at least one solution of set $X'$ obtained by another algorithm. The results are clear that 100\% solutions obtained by the basic EA are dominated by at least a solution obtained by its transform-and-divide counterpart, while none of solutions obtained by the transform-and-divide EA is dominated by at least a solution obtained by the corresponding basic EA. Consequently, decision-makers always prefer to adopt solutions produced by transform-and-divide EAs, while solutions obtained by the basic EAs can hardly provide reference. In practice, decision-makers choose final solutions for implementation as follows:
\begin{itemize}
\item The best solutions of MOEA/D on the fourth instance of ZJHTCM and the second instance of H4.
\item The best solutions of DEMOwSA on the second instance of ZJHTCM, the second instance of H1, the second instance of H2, the second instance of H3, and the second instance of H5.
\item The best solutions of MOPSO on the remaining seven instances.
\end{itemize}

\section{Conclusion}\label{sec:conclu}
This paper presents a transform-and-divide evolutionary optimization approach to medical supplies procurement under the background of COVID-19. Our approach first transforms the original high-dimensional, constrained multiobjective optimization problem to a low-dimensional, unconstrained multiobjective optimization problem, and then evaluates each solution to the transformed problem by solving a set of simple single-objective optimization subproblems, such that the problem can be efficiently solved by existing evolutionary multiobjective algorithms. We applied the transform-and-divide evolutionary optimization approach to six hospitals in Zhejiang Province, China, during the peak of COVID-19. Results showed that our approach exhibits significantly better performance than that of directly solving the original problem. Decision-makers of the hospitals always choose the best solutions produced by the transform-and-divide method for implementation and achieve promising results in balancing epidemic control and common disease treatment in practice.

The proposed transform-and-divide evolutionary optimization based on problem-specific knowledge can be an efficient solution approach to many other complex problems. For example, considering a problem of personalizing healthcare solutions for a large number of residents. As the number of candidate healthcare medicines and treatment items are large, the dimension of the problem is high. However, by clustering the residents based on their health status and limit the medicines and treatment items to each cluster, the problem dimension can be significantly reduced, and we can solve the subproblem of personalizing healthcare solutions for each cluster much more efficiently. Another example is to distribute $m$ types of disaster relief supplies from $n_1$ suppliers to $n_2$ demanders, we need to determine the quantity of each supply from each supplier to each demander. The dimension of the problem is $mn_1n_2$. However, we can establish a virtual ``intermediary'' and transform the problem to a new problem of determining the quantity of each supply from each supplier to the intermediary and that from the intermediary to each demander. Therefore, the dimension of the transformed problem is $m(n_1\!+\!n_2)$, but we need to solve additional subproblems of determining at the intermediary which parts of supplies should be sent to each demander. In many cases, the transform-and-divide strategy can greatly reduce the difficulty of problem-solving, but it often requires effective discovery and utilization of problem-specific knowledge for problem transformation and division. Although EAs are regarded as robust \emph{problem-independent} search heuristics for a large variety of optimization problems \cite{Wege02EvoOpt,YuY15TEC}, we argue that proper exploitation of \emph{problem-dependent} knowledge can significantly improve the efficiency of EAs in solving highly complex problems and, therefore, enlarge the application field of EAs.

\bibliographystyle{IEEEtran}
\bibliography{D:/paper/EnvHealth,D:/paper/bim,D:/paper/opre}

\vfill
\end{document}